\documentclass{article}

\usepackage[final]{corl_2021} % Uncomment for the camera-ready ``final'' version.

\usepackage{amsmath}
\usepackage{amssymb}
\usepackage{amsthm}
\usepackage{amsfonts}
\usepackage[font=footnotesize]{caption}
\usepackage{subcaption}
\usepackage{booktabs}
\usepackage{graphics}
\usepackage{graphicx}
\usepackage{array,multirow}
\usepackage[ruled,vlined,noend]{algorithm2e}
\SetAlFnt{\footnotesize}
\usepackage{floatrow} % for figure within theorem

\usepackage{mdframed}
\newmdtheoremenv{lemma}{Lemma}
\newmdtheoremenv{theorem}{Theorem}

\newcommand{\phitarg}[1]{\phi_{\text{targ}#1}}
\newcommand{\adashtilde}{\Tilde{a}'}
\newcommand{\atilde}{\Tilde{a}}
\DeclareMathOperator{\argmax}{argmax}
\newcommand{\expertdataset}{\mathcal{D}^{\text{exp}}}
\usepackage{listings} % beautiful code
\usepackage{xcolor}
\definecolor{codegreen}{rgb}{0,0.6,0}
\definecolor{codegray}{rgb}{0.5,0.5,0.5}
\definecolor{codepurple}{rgb}{0.58,0,0.82}
\definecolor{backcolour}{rgb}{0.95,0.95,0.92}

\renewenvironment{table}
  {\setlength\abovecaptionskip{0\p@}%
   \setlength\belowcaptionskip{-10\p@}%
   \@float{table}}
  {\end@float}

\lstdefinestyle{mystyle}{
    backgroundcolor=\color{backcolour},   
    commentstyle=\color{codegreen},
    keywordstyle=\color{magenta},
    numberstyle=\tiny\color{codegray},
    stringstyle=\color{codepurple},
    basicstyle=\ttfamily\footnotesize,
    breakatwhitespace=false,         
    breaklines=true,                 
    captionpos=b,                    
    keepspaces=true,                 
    numbers=left,                    
    numbersep=5pt,                  
    showspaces=false,                
    showstringspaces=false,
    showtabs=false,                  
    tabsize=2
}
\title{ARC - Actor Residual Critic\\for Adversarial Imitation Learning}

\author{
  Ankur Deka$^{\dagger *}$, Changliu Liu$^{\dagger}$, Katia Sycara$^{\dagger}$\\
  $^{\dagger}$Robotics Institute, Carnegie Mellon University \quad $^{*}$Intel Labs\\ 
  \texttt{adeka@alumni.cmu.edu, \{cliu6,katia\}@cs.cmu.edu} \\
}

\begin{document}
\maketitle

% \newcommand\ks[1]{{\color{red}{#1}}}
% \newcommand\ad[1]{{\color{blue}{#1}}}
% \newcommand\cl[1]{{\color{orange}{#1}}}
%===============================================================================

%===============================================================================
    
    % \vspace{-1cm}
\begin{abstract}
   Adversarial Imitation Learning (AIL) is a class  of popular state-of-the-art Imitation Learning algorithms commonly used in robotics. In AIL, an artificial adversary's misclassification is used as a reward signal that is optimized by any standard Reinforcement Learning (RL) algorithm. Unlike most RL settings, the reward in AIL is $\emph{differentiable}$ but current model-free RL algorithms do not make use of this property to train a policy. The reward is AIL is also $\emph{shaped}$ since it comes from an adversary. We leverage the differentiability property of the shaped AIL reward function and formulate a class of Actor Residual Critic (ARC) RL algorithms. ARC algorithms draw a parallel to the standard Actor-Critic (AC) algorithms in RL literature and uses a residual critic, $C$ function (instead of the standard $Q$ function) to approximate only the discounted future return (excluding the immediate reward). ARC algorithms have similar convergence properties as the standard AC algorithms with the additional advantage that the gradient through the immediate reward is exact. For the discrete (tabular) case with finite states, actions, and known dynamics, we prove that policy iteration with $C$ function converges to an optimal policy. In the continuous case with function approximation and unknown dynamics, we experimentally show that ARC aided AIL outperforms standard AIL in simulated continuous-control and real robotic manipulation tasks. ARC algorithms are simple to implement and can be incorporated into any existing AIL implementation with an AC algorithm. Video and link to code are available at: \href{https://sites.google.com/view/actor-residual-critic}{\texttt{sites.google.com/view/actor-residual-critic}}.
\end{abstract}
    \vspace{-3mm}
\section{Introduction}
\vspace{-3mm}
    Although Reinforcement Learning (RL) allows us to train agents to perform complex tasks without manually designing controllers \cite{mnih2013playing, schulman2017proximal, haarnoja2018soft}, it is often tedious to hand-craft a dense reward function that captures the task objective in robotic tasks \cite{atkeson1997robot, schaal1997learning, argall2009survey}. Imitation Learning (IL) or Learning from Demonstration (LfD) is a popular choice in such situations \cite{atkeson1997robot, schaal1997learning,argall2009survey,abbeel2010autonomous}. Common approaches to IL are Behavior Cloning (BC) \cite{bain1995framework} and Inverse Reinforcement Learning (IRL) \cite{ng2000algorithms}.

    % This paper focuses on imitation learning for model free continuous control tasks as this is typically the case with real world robotics problems - the state and action spaces are continuous and we don't have an exact model of the environment.
    
    Within IRL, recent Adversarial Imitation Learning (AIL) algorithms have shown state-of-the-art performance, especially in continuous control tasks which make them relevant to real-world robotics problems.
    AIL methods cast the IL problem as an adversarial game between a policy and a learned adversary (discriminator). The adversary aims to classify between agent and expert trajectories and the policy is trained using the adversary's mis-classification as the reward function. This encourages the policy to imitate the expert.
    % AIL involves an agent and an adversary (discriminator) competing in a 2-player game. The adversary tries to distinguish agent trajectories from expert trajectories. The agent tries to fool the adversary by generating trajectories that are similar to expert trajectories. At convergence, the agent trajectories resemble the expert trajectories and the adversary cannot distinguish between them.
    Popular AIL algorithms include Generative Adversarial Imitation Learning (GAIL) \cite{ho2016generative}, Adversarial Inverse Reinforcement Learning (AIRL) \cite{fu2017learning} and $f$-MAX \cite{ghasemipour2020divergence}.
    
    The agent in AIL is trained with any standard RL algorithm. There are two popular categories of RL algorithms: (i) on-policy algorithms such as TRPO \cite{schulman2015trust}, PPO 
    \cite{schulman2017proximal},
    GAE 
    \cite{schulman2015high} based on the policy gradient theorem \cite{williams1992simple, sutton2018reinforcement};  and (ii) off-policy Actor-Critic (AC) algorithms such as DDPG \cite{lillicrap2015continuous}, TD3 \cite{fujimoto2018addressing}, SAC \cite{haarnoja2018soft} that compute the policy gradient through a critic ($Q$ function). 
    These standard RL algorithms were designed for arbitrary scalar reward functions; and they compute an \emph{approximate} gradient for updating the policy.
    Practical on-policy algorithms based on the policy gradient theorem use several approximations to the true gradient \cite{schulman2015trust,schulman2017proximal,schulman2015high} and off-policy AC algorithms first approximate policy return with a critic ($Q$ function) and subsequently compute the gradient through this critic \cite{lillicrap2015continuous, fujimoto2018addressing, haarnoja2018soft}. Even if the $Q$ function is approximated very accurately, the error in its gradient can be arbitrarily large, Appendix \ref{appendix:abritrary_large_error_in_gradient}. 
    
    Our insight is that the reward function in AIL has 2 special properties: (i) it is \emph{differentiable} which means we can compute the exact gradient through the reward function instead of approximating it and (ii) it is dense/shaped as it comes from an adversary. As we will see in section \ref{subsection:naive_diff}, naively computing the gradient through reward function would lead to a short-sighted sub-optimal policy. To address this issue, we formulate a class of Actor Residual Critic (ARC) RL algorithms that use a residual critic, $C$ function (instead of the standard $Q$ function) to approximate only the discounted future return (excluding immediate reward).
    
    The contribution of this paper is the introduction of ARC, which can be easily incorporated to replace the AC algorithm in any existing AIL algorithm for continuous-control and helps boost the asymptotic performance by computing the exact gradient through the shaped reward function.
    
\vspace{-4mm}
\section{Related Work}
\vspace{-6mm}
    \newcommand{\A}{\mathcal{A}}
    \newcommand{\rhoagent}{{\rho^\pi(s,a)}}
    \newcommand{\rhoexpert}{{\rho^\textnormal{exp}(s,a)}}
    \newcommand{\rhoagentshort}{{\rho^\pi}}
    \newcommand{\rhoexpertshort}{{\rho^\textnormal{exp}}}
    
    \newcommand{\D}{\mathcal{D}}
    \newcommand{\Ebb}{\mathbb{E}}
    \newcommand{\Q}{$Q$}
    \newcommand{\C}{$C$}
    \newcommand{\infnorm}[1]{||#1||_\infty}
    \newcommand{\abs}[1]{\left|#1\right|}
    \begin{table*}[h!]
    \centering
    \footnotesize
    \begin{tabular}{r||c|c||c}
        \toprule
        \multirow{2}{*}{\textbf{Algorithm}} & \multicolumn{2}{c||}{\textbf{Minimized $f$-Divergence}} & \multirow{2}{*}{$r(s,a)$}\\
         & \textbf{Name} & \textbf{Expression} & \\
        \toprule
        GAIL \cite{ho2016generative} & Jensen-Shannon & 
        $\frac{1}{2}
        \left\{
        \Ebb_\rhoexpertshort \log \frac{2\rhoexpertshort}{\rhoexpertshort+\rhoagentshort}
        +
        \Ebb_\rhoagentshort \log \frac{2\rhoagentshort}{\rhoexpertshort+\rhoagentshort}
        \right\}
        $
        & $\log D(s,a)$\\
        \midrule
        AIRL \cite{fu2017learning}, & \multirow{2}{*}{Reverse KL} & 
        \multirow{2}{*}{$\Ebb_\rhoagentshort \log \frac{\rhoagentshort}{\rhoexpertshort}$} &
        \multirow{2}{*}{$\log \frac{D(s,a)}{1-D(s,a)}$}
        \\
        $f$-MAX-RKL \cite{ghasemipour2020divergence} &   &
        & \\
        \bottomrule
    \end{tabular}
    \caption{Popular AIL algorithms, $f$-divergence metrics they minimize and their reward functions.}
    \label{tab:algos_divergence_reward}
\end{table*}
\vspace{-5mm}
    The simplest approach to imitation learning is Behavior Cloning \cite{bain1995framework} where an agent policy directly regresses on expert actions (but not states) using supervised learning. This leads to distribution shift and poor performance at test time \cite{ross2011reduction,ho2016generative}. Methods such as DAgger \cite{ross2011reduction} and Dart \cite{laskey2017iterative} eliminate this issue but assume an interactive access to an expert policy, which is often impractical.
    
    Inverse Reinforcement Learning (IRL) approaches recover a reward function which can be used to train an agent using RL \cite{ng2000algorithms,ziebart2008maximum} and have been more successful than BC. Within IRL, recent Adversarial Imitation Learning (AIL) methods inspired by Generative Adversarial Networks (GANs) \cite{goodfellowgenerative} have been extremely successful. GAIL \cite{ho2016generative} showed state-of-the-art results in imitation learning tasks following which several extensions have been proposed \cite{li2017infogail,jena2021augmenting}. AIRL \cite{fu2017learning} imitates an expert as well as recovers a robust reward function. \cite{ke2019imitation} and \cite{ghasemipour2020divergence} presented a unifying view on AIL methods by showing that they minimize different divergence metrics between expert and agent state-action distributions but are otherwise similar. \cite{ghasemipour2020divergence} also presented a generalized AIL method $f$-MAX which can minimize any specified $f$-divergence metric \cite{lin1991divergence} between expert and agent state-action distributions thereby imitating the expert. Choosing different divergence metrics leads to different AIL algorithms, e.g. choosing Jensen-Shannon divergence leads to GAIL \cite{ho2016generative}. \cite{zhang2020f} proposed a method that automatically learns a $f$-divergence metric to minimize. Our proposed Actor Residual Critic (ARC) can be augmented with any of these AIL algorithms to leverage the reward gradient.
    
    % Differentiable cost (negative reward) has been leveraged in control research in the form of LQR and its extensions but they assume access to a dynamics models.
    Some recent methods have leveraged the differentiable property of reward in certain scenarios but they have used this property in very different settings. \cite{ni2021f} used the gradient of the reward to improve the reward function but not to optimize the policy. We on the other hand explicitly use the gradient of the reward to optimize the policy. \cite{hafner2019dream} used the gradient through the reward to optimize the policy but operated in the model-based setting. If we have access to a differentiable dynamics model, we can directly obtain the gradient of the expected return (policy objective) w.r.t. the policy parameters, Appendix \ref{appendix:comparison_differentiable_il}. Since we can directly obtain the objective's gradient, we do not necessarily need to use either a critic ($Q$) as in standard Actor Critic (AC) algorithms or a residual critic ($C$) as in our proposed Actor Residual Critic (ARC) algorithms. Differentiable cost (negative reward) has also been leveraged in control literature for a long time to compute a policy, e.g. in LQR \cite{bemporad2002explicit} and its extensions; but they assume access to a known dynamics model. We on the other hand present a model-free method with unknown dynamics that uses the gradient of the reward to optimize the policy with the help of a new class of RL algorithms called Actor Residual Critic (ARC).
%auto-ignore

    \vspace{-3mm}
\section{Background}
\vspace{-3mm}
\paragraph{Objective}
Our goal is to imitate an expert from one or more demonstrated trajectories (state-action sequences) in a continuous-control task (state and action spaces are continuous). Given any Adversarial Imitation Learning (AIL) algorithm that uses an off-policy Actor-Critic algorithm RL algorithm, we wish to use our insight on the availability of a differentiable reward function to improve the imitation learning algorithm.
\vspace{-3mm}
\paragraph{Notation}
The environment is modeled as a Markov Decision Process (MDP) represented as a tuple $(\mathcal{S}, \mathcal{A}, \mathcal{P}, r, \rho_0, \gamma)$ with state space $\mathcal{S}$, action space $\mathcal{A}$, transition dynamics $\mathcal{P}:\mathcal{S} \times \mathcal{A} \times \mathcal{S} \rightarrow [0,1]$, reward function $r(s,a)$, initial state distribution $\rho_0(s)$, and discount factor $\gamma$. $\pi(.|s)$, $\pi^{\text{exp}}\left(.|s\right)$ denote policies and $\rhoagentshort, \rhoexpertshort: \mathcal{S} \times \mathcal{A} \rightarrow [0,1]$ denote state-action occupancy distributions for agent and expert respectively. $\mathcal{T}=\{s_1,a_1,s_2,a_2,\dots,s_T,a_T\}$ denotes a trajectory or episode and $(s,a,s',a')$ denotes a continuous segment in a trajectory.  A discriminator or adversary $D(s,a)$ tries to determine whether the particular $(s,a)$ pair belongs to an expert trajectory or agent trajectory, i.e. $D(s,a)=P(\text{expert}|s,a)$. The optimal discriminator is $D(s,a)=\frac{\rhoexpert}{\rhoexpert+\rhoagent}$ \cite{goodfellowgenerative}.
\vspace{-3mm}
\paragraph{Adversarial Imitation Learning (AIL)}
In AIL, the discriminator and agent are alternately trained. The discriminator is trained to maximize the likelihood of correctly classifying expert and agent data using supervised learning, \eqref{eq:discriminator_objective} and the agent is trained to maximize the expected discounted return, \eqref{eq:agent_objective}.
% \ks{I would be good to say in English  what these functions in eqs (1) and (2) are maximizing.  }
\vspace{-2mm}
\begin{align}
    &\max_D \Big\{ \Ebb_{s,a\sim\rhoexpertshort} [\log D(s,a)]
    +
    \Ebb_{s,a\sim\rhoagentshort} \left[ \log (1-D(s,a)) \right] \Big\}
    \label{eq:discriminator_objective} \\
    &\max_\pi \Big\{\Ebb_{s,a \sim \rho_0, \pi, \mathcal{P}} \sum_{t\geq 0} \gamma^t r(s_t,a_t)\Big\}
    % + \ad{\alpha \mathcal{H(\pi)}} \Big\}
    \label{eq:agent_objective}
\end{align}
\vspace{-5mm}

Here, reward $r_\psi(s,a)=h(D_\psi(s,a))$ is a function of the discriminator which varies between different AIL algorithms.
% $\mathcal{H(\pi)}$ is the entropy of agent policy and $\alpha$ is a hyperparameter.
Different AIL algorithms minimize different $f$-divergence metrics between expert and agent state-action distribution. Defining a $f$-divergence metric instantiates different reward functions \cite{ghasemipour2020divergence}. Some popular divergence choices are Jensen-Shannon in GAIL \cite{ho2016generative} and Reverse Kullback-Leibler in $f$-MAX-RKL \cite{ghasemipour2020divergence} and AIRL \cite{fu2017learning} as shown in Table \ref{tab:algos_divergence_reward}.

Any RL algorithm could be used to optimize \eqref{eq:agent_objective} and popular choices are off-policy Actor-Critic algorithms such as DDPG \cite{lillicrap2015continuous}, TD3 \cite{fujimoto2018addressing}, SAC \cite{haarnoja2018soft} and on-policy algorithms such as TRPO \cite{schulman2015trust}, PPO \cite{schulman2017proximal}, GAE \cite{schulman2015high} which are based on the policy gradient theorem \cite{williams1992simple,sutton2018reinforcement}. We focus on off-policy Actor-Critic algorithms as they are usually more sample efficient and stable than on-policy policy gradient algorithms \cite{fujimoto2018addressing, haarnoja2018soft}.
\vspace{-3mm}
\paragraph{Continuous-control using off-policy Actor-Critic}
The objective in off-policy RL algorithms is to maximize expected $Q$ function of the policy, $Q^\pi$ averaged over the state distribution of a dataset $\mathcal{D}$ (typically past states stored in buffer) and the action distribution of the policy $\pi$ \cite{silver2014deterministic}:
\begin{align}
    & \max_\pi \Ebb_{s\sim \mathcal{D}, a\sim \pi} Q^\pi(s,a)\\
    \text{where, } & Q^\pi(s,a) = \Ebb_{s,a\sim\rho_0,\pi,\mathcal{P}} \bigg[ \sum_{k\geq0} \gamma^{k} r_{t+k} \bigg\vert s_t=s, a_t=a \bigg] \label{eq:q_definition}
\end{align}
The critic and the policy denoted by $Q$, $\pi$ respectively are approximated by function approximators such as neural networks with parameters $\phi$ and $\theta$ respectively. There is an additional target $Q_{\phitarg{}}$ function parameterized by $\phitarg{}$. There are two alternating optimization steps:
\begin{enumerate}
    \item Policy evaluation: Fit critic ($Q_\phi$ function) by minimizing Bellman Backup error.
    \begin{align}
        & \min_\phi \Ebb_{s,a,s'\sim \mathcal{D}} \left\{ Q_\phi(s,a)-y(s,a) \right\} ^ 2 \label{eq:ac_bellman_error}\\
        \text{where, } &y(s,a) = r(s,a) + \gamma Q_{\phitarg{}}(s', a') \text{ and } a' \sim \pi_\theta(.|s') 
    \end{align}
    $Q_\phi$ is updated with gradient descent without passing gradient through the target $y(s,a)$.
    % \ks{I do not understand this about fixed supervised learning label--what is it and why is it relevant to mention here?}
    
    \item Policy improvement: Update policy with gradient ascent over RL objective.
    \begin{align}
         \Ebb_{s\sim \mathcal{D}} \big[ \nabla_\theta  Q_\phi(s,a\sim \pi_\theta(.|s)) \big] \label{eq:ac_policy_update}
    \end{align}
\end{enumerate}
All off-policy Actor Critic algorithms follow the core idea above (\eqref{eq:ac_bellman_error} and \eqref{eq:ac_policy_update}) along with additional details such as the use of a deterministic policy and target network in DDPG \cite{lillicrap2015continuous}, double Q networks and delayed updates in TD3 \cite{fujimoto2018addressing}, entropy regularization and reparameterization trick in SAC \cite{haarnoja2018soft}.

% \vspace{-5mm}
\paragraph{Naive-Diff and why it won't work}
\label{subsection:naive_diff}
Realizing that the reward in AIL is differentiable and shaped, we can formulate a Naive-Diff RL algorithm that updates the policy by differentiating the RL objective \eqref{eq:agent_objective} with respect to the policy parameters $\theta$.
% \ks{I do not know what you mean by  "simply differentiating through the reward".}
\begin{align}
    \Ebb_{\mathcal{T}\sim\mathcal{D}}
    \left[
    \nabla_\theta r(s_1,a_1) + \gamma \nabla_\theta r(s_2,a_2) + \gamma^2 \nabla_\theta r(s_3,a_3) + \dots
    \right] 
    \label{eq:naive_diff}
\end{align}
$\mathcal{T}=\{s_1,a_1,s_2,a_2\dots \}$ is a sampled trajectory in $\mathcal{D}$. Using standard autodiff packages such as Pytorch \cite{paszke2019pytorch} or Tensorflow \cite{abadi2016tensorflow} to naively compute the gradients in \eqref{eq:naive_diff} would produce incorrect gradients. Apart from the immediate reward $r(s_1,a_1)$, all the terms depend on the transition dynamics of the environment $\mathcal{P}(s_{t+1}|s_t,a_t)$, which is unknown and we cannot differentiate through it. So, autodiff will calculate the gradient of only immediate reward correctly and calculate the rest as $0$'s.
% The effective gradient calculated would be $\Ebb_{s\sim \mathcal{D}} \big[ \nabla_\theta  r(s_1,a_1) \big]$.
This will produce a short-sighted sub-optimal policy that maximizes only the immediate reward.
%auto-ignore

    \vspace{-3mm}
\section{Method}
\vspace{-5mm}
\begin{figure}[h!]
    \centering
    \includegraphics[width=0.45\textwidth]{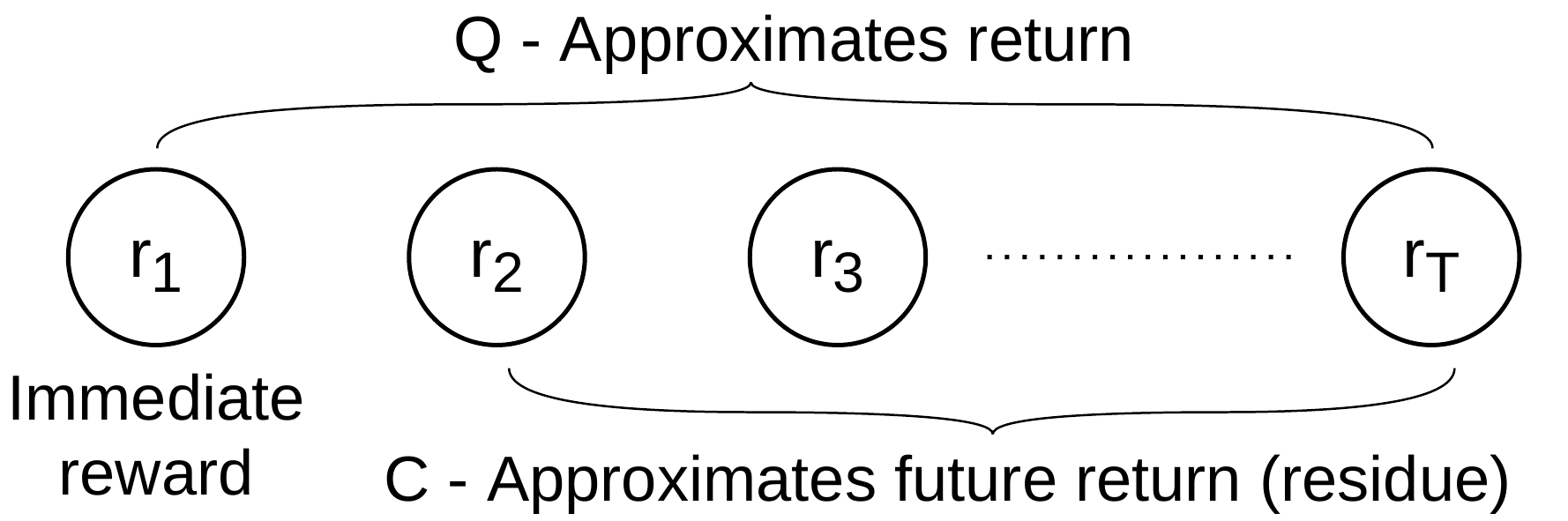}
    \vspace{-3mm}
    \caption{Visual illustration of approximating reward via Q function or C function.}
    \label{fig:q_vs_c}
\end{figure}
\vspace{-3mm}
The main lesson we learnt from Naive-Diff is that while we can obtain the gradient of immediate reward, we cannot directly obtain the gradient of future return due to unknown environment dynamics. This directly motivates our formulation of Actor Residual Critic (ARC). Standard Actor Critic algorithms use $Q$ function to approximate the return as described in Eq. \ref{eq:q_definition}. However, since we can directly obtain the gradient of the reward, we needn't approximate it with a $Q$ function. We, therefore, propose to use $C$ function to approximate only the future return, leaving out the immediate reward. This is the core idea behind Actor Residual Critic (ARC) and is highlighted in Fig. \ref{fig:q_vs_c}. The word ``Residual" refers to the amount of return that remains after subtracting the immediate reward from the return. As we will see in Section \ref{section:continuous_control_using_arc}, segregating the immediate reward from future return will allow ARC algorithms to leverage the exact gradient of the shaped reward. We now formally describe Residual Critic ($C$ function) and its relation to the standard critic ($Q$ function).
\vspace{-3mm}
\subsection{Definition of Residual Critic ($C$ function)}
\vspace{-3mm}
The Q function under a policy $\pi$, $Q^\pi(s,a)$, is defined as the expected discounted return from state $s$ taking action $a$, \eqref{eq:q_definition_repeat}.
% \begin{align}
%     Q^\pi(s,a) = \Ebb_{s,a\sim\rho_0,\pi,\mathcal{P}} \bigg[ \sum_{k\geq0} \gamma^{k} r_{t+k} \bigg\vert s_t=s, a_t=a \bigg]
%     \label{eq:q_definition_repeat}
%     % Q^\pi(s,a) = \Ebb \sum_{k\geq0} \gamma^{k} r_{t+k}
% \end{align}
The $C$ function under a policy $\pi$, $C^\pi(s,a)$, is defined as the expected discounted future return, excluding the immediate reward \eqref{eq:c_definition}. Note that the summation in \eqref{eq:c_definition} starts from $1$ instead of $0$. $Q$ function can be expressed in terms of $C$ function as shown in \eqref{eq:q_c_relation}.
\begin{align}
    Q^\pi(s,a) &= \Ebb_{s,a\sim\rho_0,\pi,\mathcal{P}} \bigg[ \sum_{k\geq0} \gamma^{k} r_{t+k} \bigg\vert s_t=s, a_t=a \bigg]
    \label{eq:q_definition_repeat}\\
    C^\pi(s,a) &= \Ebb_{s,a\sim\rho_0,\pi,\mathcal{P}} \bigg[ \sum_{k\geq1} \gamma^{k} r_{t+k} \bigg\vert s_t=s, a_t=a \bigg]
    % C^\pi(s_t,a_t) = \Ebb \sum_{k\geq1} \gamma^{k} r_{t+k}
    \label{eq:c_definition}\\
    Q^\pi(s,a) &= r(s,a) + C^\pi(s,a) \label{eq:q_c_relation}
\end{align}
% $Q$ function can be expressed in terms of $C$ function:
% \begin{align}
% Q^\pi(s,a) &= r(s,a) + C^\pi(s,a) \label{eq:q_c_relation}
% \end{align}

\vspace{-5mm}
\subsection{Policy Iteration using $C$ function}
\vspace{-2mm}
\begin{algorithm}[h!]
\SetAlgoLined
 Initialize $C^0(s,a) \forall s,a$\;
 \While{$\pi$ not converged}{
 // Policy evaluation\\
  \For{n=1,2,\dots until $C_k$ converges}
  {$C^{n+1}(s,a) \leftarrow \gamma \sum_{s'} P(s'|s,a) \sum_{a'} \pi(a'|s') \left(r(s',a') + C^{n}(s',a')\right) \quad \forall s,a $
  }
  // Policy improvement\\
  $\pi(s,a) \leftarrow$
  $\begin{cases}
  1, \text{ if } a = \argmax_{a'} \left(r(s,a') + C(s,a')\right)\\
  0, \text{ otherwise}
  \end{cases}
  \forall s,a$\\
 }
 \caption{Policy Iteration with $C$ function}
 \label{algo:c_policy_iteration}
% \vspace{-5mm}
\end{algorithm}

Using $C$ function, we can formulate a Policy Iteration algorithm as shown in Algorithm \ref{algo:c_policy_iteration}, which is guaranteed to converge to an optimal policy (Theorem \ref{theorem:c_policy_iteration}), similar to the case of Policy Iteration with $Q$ or $V$ function \cite{sutton2018reinforcement}. Other properties of $C$ function and proofs are presented in Appendix \ref{appendix:c_properties}.

\vspace{-3mm}
\subsection{Continuous-control using Actor Residual Critic}
\vspace{-3mm}
\label{section:continuous_control_using_arc}
% Now that we have a working policy iteration algorithm with $C$ function,
We can easily extend the policy iteration algorithm with $C$ function (Algorithm \ref{algo:c_policy_iteration}) for continuous-control tasks using function approximators instead of discrete $C$ values and a discrete policy (similar to the case of $Q$ function \cite{sutton2018reinforcement}). We call any RL algorithm that uses a policy, $\pi$ and a residual critic, $C$ function as an Actor Residual Critic (ARC) algorithm. Using the specific details of different existing Actor Critic algorithms, we can formulate analogous ARC algorithms. For example, using a deterministic policy and target network as in \cite{lillicrap2015continuous} we can get ARC-DDPG.  Using double C networks (instead of Q networks) and delayed updates as in \cite{fujimoto2018addressing} we can get  ARC-TD3. Using entropy regularization and reparameterization trick as in \cite{haarnoja2018soft} we can get ARC-SAC or SARC (Soft Actor Residual Critic).
% We show the SARC algorithm in Algorithm \ref{algo:sarc}. 

\vspace{-3mm}
\subsection{ARC aided Adversarial Imitation Learning}
\vspace{-3mm}
To incorporate ARC in any Adversarial Imitation Learning algorithm, we simply replace the Actor Critic RL algorithm with an ARC RL algorithm without altering anything else in the pipeline. For example,  we can replace SAC \cite{haarnoja2018soft} with SARC to get SARC-AIL as shown in Algorithm \ref{algo:sarc_ail}. Implementation-wise this is extremely simple and doesn't require any additional functional parts in the algorithm. The same neural network that approximated $Q$ function can be now be used to approximate $C$ function.
{
\begin{algorithm}[h]
\SetAlgoLined
 \textbf{Intialization}: Environment (env), Discriminator parameters $\psi$, Policy parameters $\theta$, $C$-function parameters $\phi_1$, $\phi_2$, dataset of expert demonstrations $\expertdataset$, replay buffer $\mathcal{D}$, 
 Target parameters $\phitarg{1}\leftarrow\phi_1$, $\phitarg{2} \leftarrow \phi_2$, Entropy regularization coefficient $\alpha$\;
 \While{Max no. of environment interactions is not reached}{
 $a\sim\pi_\theta(.|s)$\;
 $s',r,d = \text{env.step(a)}$; \quad $d=1$ if $s'$ is terminal state, $0$ otherwise\\
 Store $(s,a,s',d)$ in replay buffer $\mathcal{D}$\;
 \If{Update interval reached}{
    \For{no. of update steps}{
        Sample batch $B={(s,a,s',d)}\sim \mathcal{D}$\;
        Sample batch of expert demonstrations $B^\text{exp}={(s,a)}\sim \expertdataset$\;
        
        Update Discriminator parameters ($\psi$) with gradient ascent.\\
        \qquad
        $\nabla_\psi \Big\{
        \sum_{(s,a)\in B^{\text{exp}}} [\log D_\psi(s,a)]
        +
        \sum_{(s,a,s',d)\in B} \left[ \log (1-D_\psi(s,a)) \right] \Big\} $ \;
        Compute $C$ targets $\forall (s,a,s',d) \in B$\\
        \qquad
        $
        y(s,a,d) = \gamma \left( 
        r_\psi(s',\adashtilde) + \min_{i=1,2} C_{\phitarg{i}}(s', \adashtilde) - \alpha \log \pi_\theta
        (\adashtilde|s')\right), \quad \adashtilde \sim \pi_\theta(.|s'), r_\psi(s',\adashtilde) = h(D_\psi(s,',\adashtilde))
        $
        
        Update C-functions parameters ($\phi_1, \phi_2$) with gradient descent.\\
        \qquad
        $
        \nabla_{\phi_{i}} 
        \frac{1}{|B|} \sum_{(s,a,s',d) \in B}
        \left(
        C_{\phi_i}(s,a) - y(s,a,d)
        \right)^2, \quad \text{ for } i = 1,2
        $
            
        Update policy parameters ($\theta$) with gradient ascent.\\
        \qquad
        $
        \nabla_\theta \frac{1}{|B|}\sum_{s\in B}
        \bigg(
        r_\psi(s,\atilde)
        + \min_{i=1,2} C_{\phi_i}(s,\atilde) 
        -\alpha \log \pi_\theta(\atilde|s) 
        \bigg),
        \quad
        \atilde \sim \pi_\theta(.|s), r_\psi(s,\atilde) = h(D_\psi(s,\atilde))
        $
        
        Update target networks.\\
        \qquad
        $
        \phitarg{i} \leftarrow \zeta \phitarg{i} + (1-\zeta) \phi_{i}, \quad \text{ for }i=1,2; \quad \zeta$
        controls polyak averaging
    }
 }
 }
 \caption{\footnotesize SARC-AIL: Soft Actor Residual Critic Adversarial Imitation Learning}
\label{algo:sarc_ail}
% \vspace{-5mm}
\end{algorithm}
% \vspace{-5cm}
}

\vspace{-3mm}
\subsection{Why choose ARC over Actor-Critic in Adversarial Imitation Learning?}
\vspace{-3mm}
The advantage of using an ARC algorithm over an Actor-Critic (AC) algorithm is that we can leverage the exact gradient of the reward. Standard AC algorithms use $Q_\phi$ to approximate the immediate reward + future return and then compute the gradient of the policy parameters through the $Q_\phi$ function \eqref{eq:ac_gradient}. This is an approximate gradient with no bound on the error in gradient, since the $Q_\phi$ function is an estimated value, Appendix \ref{appendix:abritrary_large_error_in_gradient}. On the other hand, ARC algorithms segregate the immediate reward (which is known in Adversarial Imitation Learning) from the future return (which needs to be estimated). ARC algorithms then compute the gradient of policy parameters through the immediate reward (which is exact) and the $C$ function (which is approximate) separately \eqref{eq:arc_gradient}.
\begin{align}
    \text{Standard AC} \quad &\Ebb_{s\sim \mathcal{D}} \big[ \nabla_\theta  Q_\phi(s,a) \big], &a\sim \pi_\theta(.|s)
    \label{eq:ac_gradient}\\
    \text{ARC (Our)} \quad &\Ebb_{s\sim \mathcal{D}} \big[ \nabla_\theta  r(s,a) 
    + \nabla_\theta  C_\phi(s,a) \big], &a\sim \pi_\theta(.|s) 
    \label{eq:arc_gradient}
\end{align}
In Appendix \ref{appendix:arc_snr}, we derive the conditions under which ARC is likely to outperform AC by performing a (Signal to Noise Ratio) SNR analysis similar to \cite{robertssignal}. Intuitively, favourable conditions for ARC are (i) Error in gradient due to function approximation being similar or smaller for $C$ as compared to $Q$ (ii) the gradient of the immediate reward not having a high negative correlation with the gradient of $C$ ($\mathbb{E}\left[\nabla_a r(s,a) \nabla_a C(s,a)\right]$ is not highly negative). Under these conditions, ARC would produce a higher $SNR$ estimate of the gradient to train the policy. We believe that AIL is likely to present favourable conditions for ARC since the reward is shaped.

ARC would under-perform AC if the error in gradient due to function approximation of $C$ network is significantly higher than that of $Q$ network. In the general RL setting, immediate reward might be misleading (i.e. $\mathbb{E}\left[\nabla_a r(s,a) \nabla_a C(s,a)\right]$ might be negative) which might hurt the performance of ARC. However, we propose using ARC for AIL where the adversary reward measures how closely the agent imitates the expert. In AIL, the adversary reward is dense/shaped making ARC likely to be useful in this scenario, as experimentally verified in the following section.

    \vspace{-5mm}
\section{Results}
\vspace{-3mm}
\label{section:results}
In Theorem \ref{theorem:c_policy_iteration}, we proved that Policy Iteration with $C$ function converges to an optimal policy. In Fig. \ref{fig:policy_iteration_grid}, we experimentally validate this on an example grid world. The complete details are presented in Appendix \ref{appendix:grid_world}. In the following sections (\ref{section:results_continuous_control}, \ref{section:results_robotic_sim} and \ref{section:results_sim_to_real}) we show the effectiveness of ARC aided AIL in Mujoco continuous-control tasks, and simulated and real robotic manipulation tasks. In Appendix \ref{appendix:toy_example_gradient_accuracy}, we experimentally illustrate that ARC produces more accurate gradients than AC using a simple 1D driving environment. The results are discussed in more detail in Appendix \ref{apendix:result_discussion}.
% In the following section (Section \ref{section:results_continuous_control}), we show the effectiveness of ARC aided AIL in Mujoco continuous-control tasks through experiments in 4 different environments. 

\vspace{-3mm}
\subsection{Policy Iteration on a Grid World}
\label{section:results_grid_world}
\newcommand{\subfigwidth}{0.2\textwidth}
\begin{figure}[H]
     \centering
     \begin{subfigure}[b]{\subfigwidth}
         \centering
         \includegraphics[width=\textwidth]{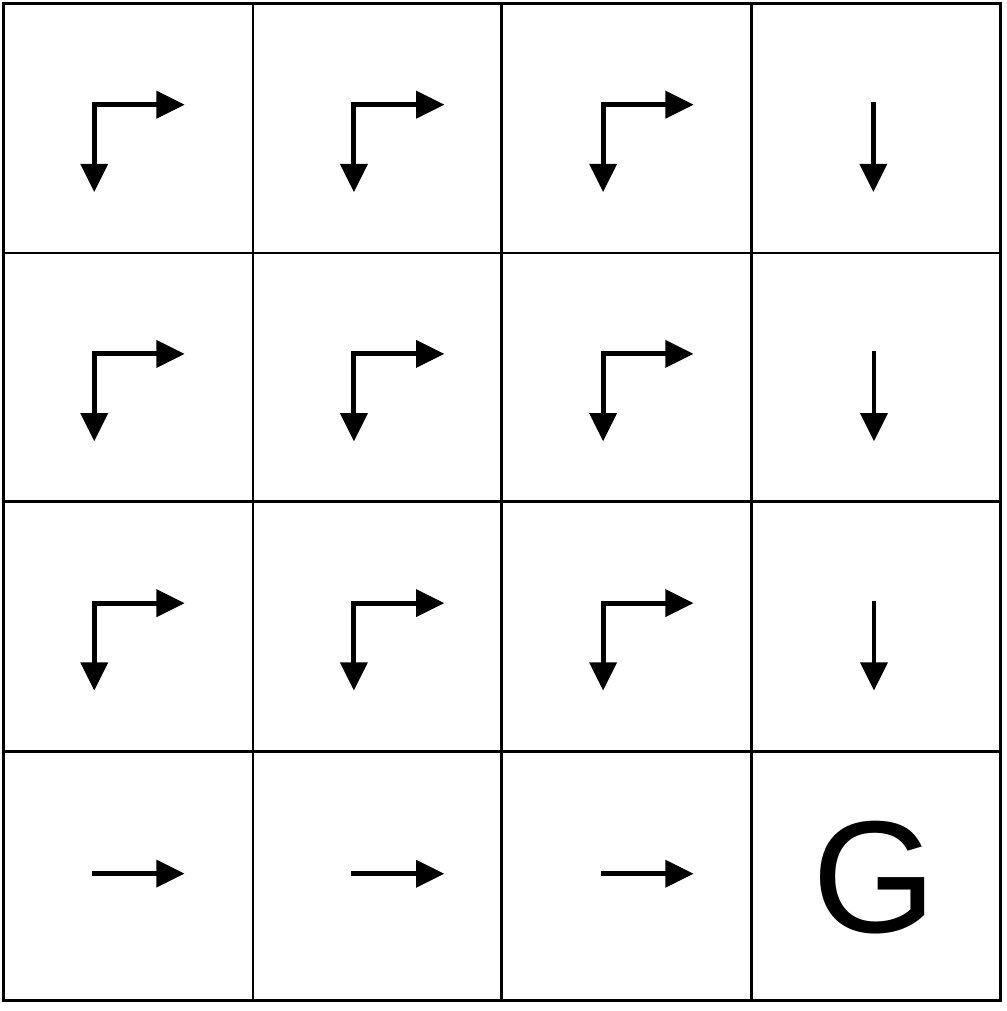}
         \vspace{-5mm}
         \caption{$\pi^*$}
         \label{fig:policy_iteration_grid_policy}
     \end{subfigure}
    %  \hfill
    \quad
     \begin{subfigure}[b]{\subfigwidth}
         \centering
         \includegraphics[width=\textwidth]{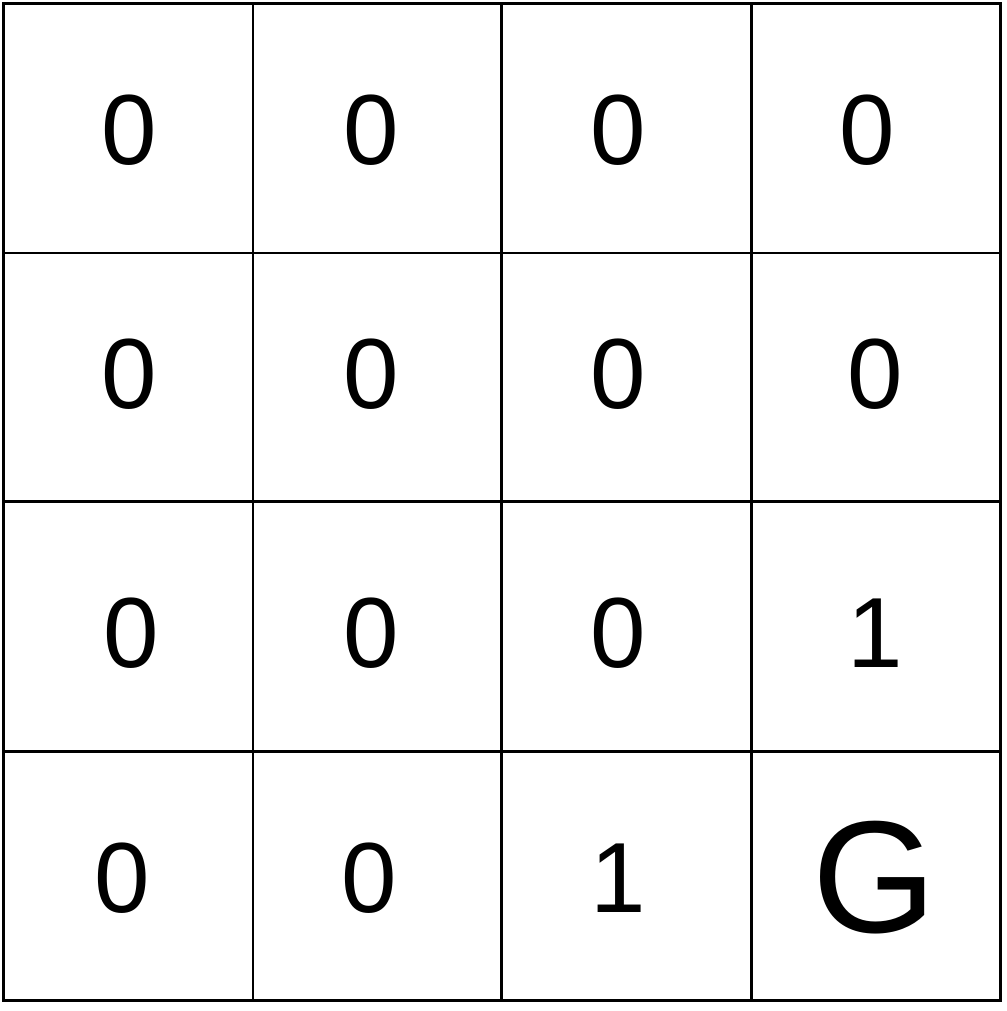}
         \vspace{-5mm}
         \caption{$r^*$}
         \label{fig:policy_iteration_grid_r}
     \end{subfigure}
    %  \hfill
    \quad
     \begin{subfigure}[b]{\subfigwidth}
         \centering
         \includegraphics[width=\textwidth]{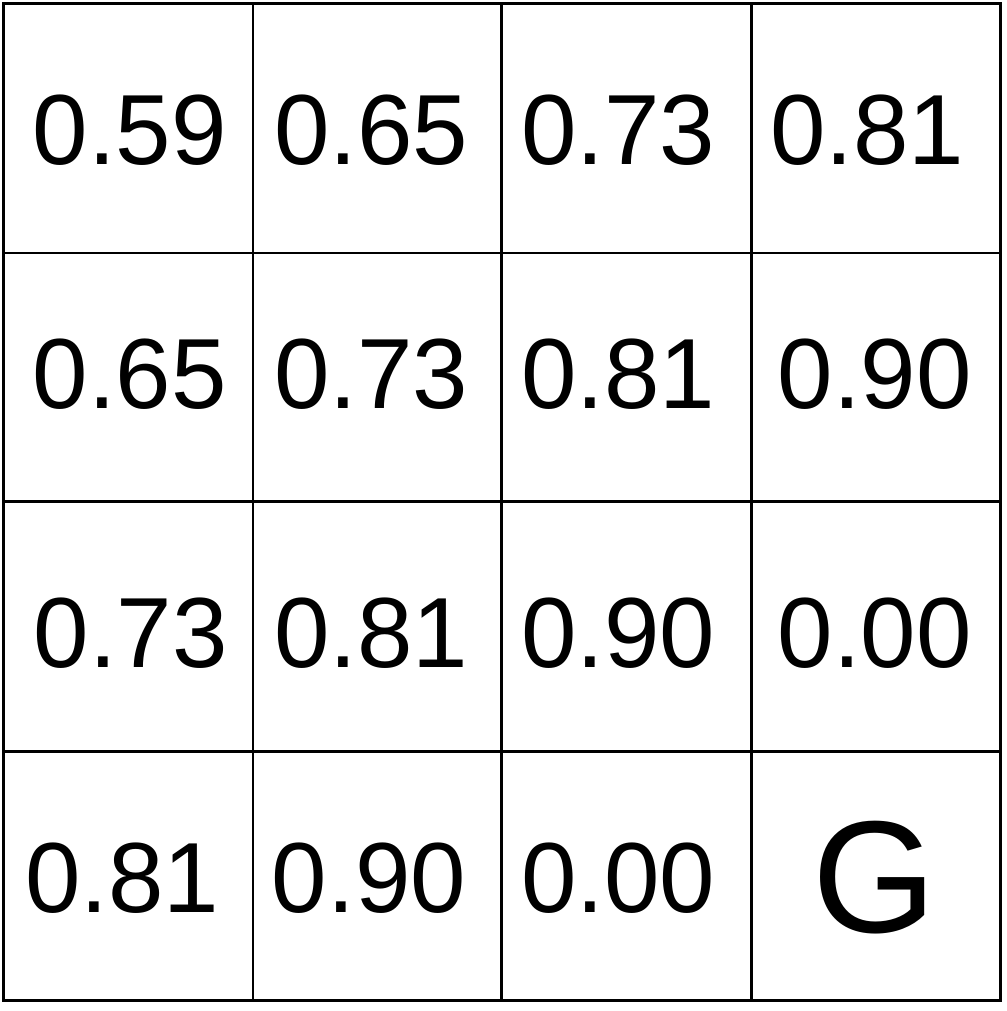}
         \vspace{-5mm}
         \caption{$C^*$}
         \label{fig:policy_iteration_grid_c}
     \end{subfigure}
    %  \hfill
    \quad
     \begin{subfigure}[b]{\subfigwidth}
         \centering
         \includegraphics[width=\textwidth]{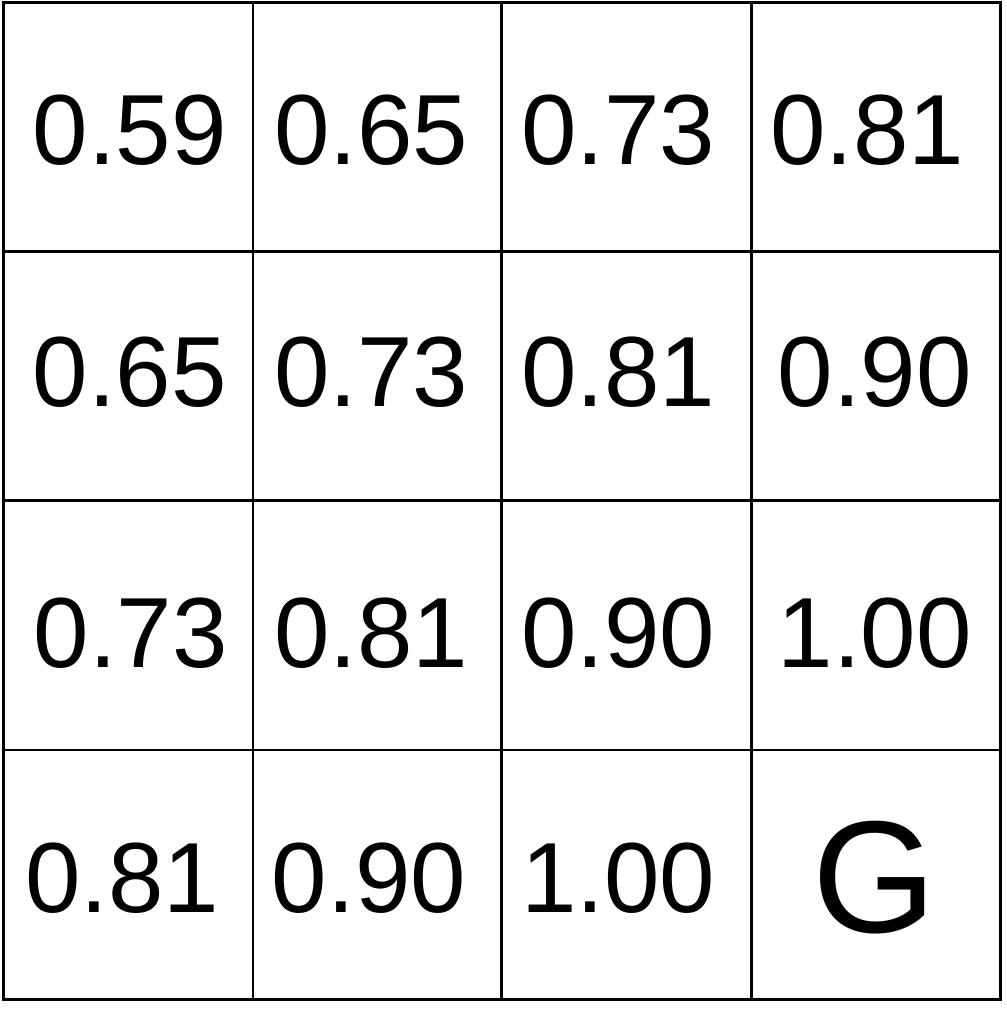}
         \vspace{-5mm}
         \caption{$Q^* = r^* + C^*$}
         \label{fig:policy_iteration_grid_q}
     \end{subfigure}
     \vspace{-2mm}
     \caption{On a Grid World, the results of running two Policy Iteration (PI) algorithms - PI with $C$ function (Algorithm \ref{algo:c_policy_iteration}) and the standard PI with $Q$ function (Appendix \ref{appendix:q_policy_iteration} Algorithm \ref{algo:q_policy_iteration}). Both algorithms converge in 7 policy improvement steps to the same optimal policy $\pi^*$ as shown in \subref{fig:policy_iteration_grid_policy}. The optimal policy gets the immediate reward shown shown \subref{fig:policy_iteration_grid_r}. The $C$ values \subref{fig:policy_iteration_grid_c} at the convergence of PI with $C$ function and the $Q$ values \subref{fig:policy_iteration_grid_q} at the convergence of PI with $Q$ function are consistent with their relation $Q^*=r^*+C^*$ \eqref{eq:q_c_relation}. Details are in \ref{appendix:grid_world}.}
     \label{fig:policy_iteration_grid}
% \vspace{-5mm}
\end{figure}

\vspace{-5mm}
\subsection{Imitation Learning in Mujoco continuous-control tasks}
\vspace{-3mm}
\label{section:results_continuous_control}

\begin{figure}[ht]
    \centering
    \begin{subfigure}[b]{\subfigwidth}
         \centering
         \includegraphics[width=\textwidth]{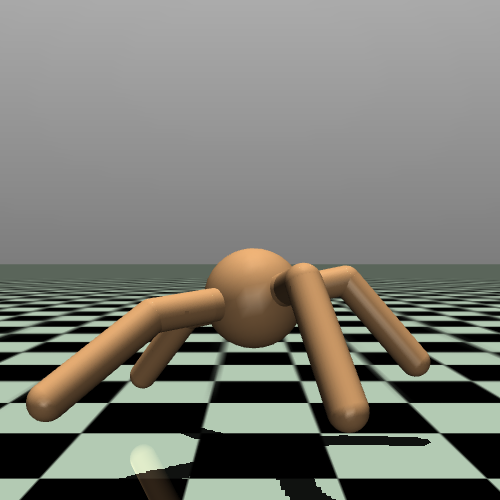}
         \vspace{-5mm}
         \caption{Ant-v2}
     \end{subfigure}
     \begin{subfigure}[b]{\subfigwidth}
         \centering
         \includegraphics[width=\textwidth]{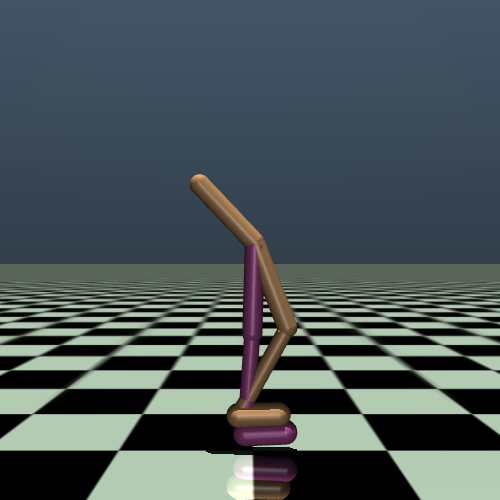}
         \vspace{-5mm}
         \caption{Walker-v2}
     \end{subfigure}
     \begin{subfigure}[b]{\subfigwidth}
         \centering
         \includegraphics[width=\textwidth]{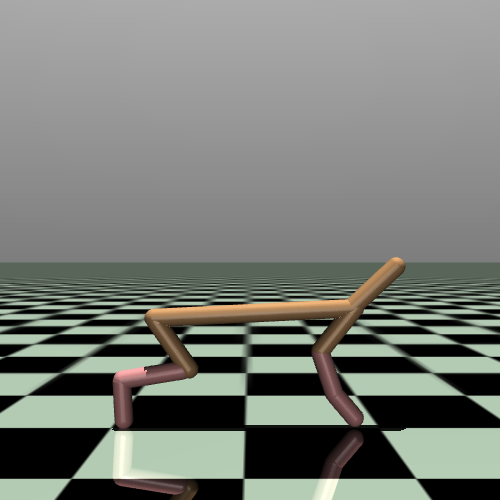}
         \vspace{-5mm}
         \caption{HalfCheetah-v2}
     \end{subfigure}
     \begin{subfigure}[b]{\subfigwidth}
         \centering
         \includegraphics[width=\textwidth]{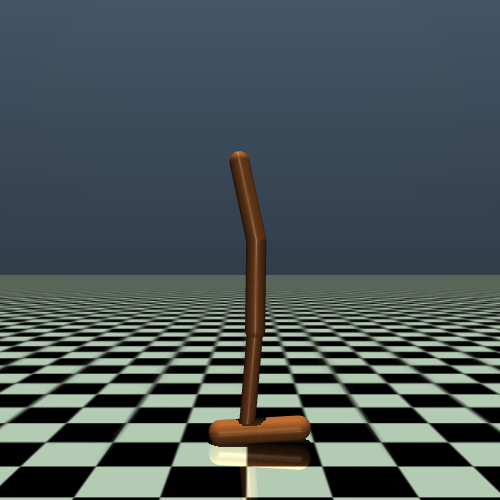}
         \vspace{-5mm}
         \caption{Hopper-v2}
     \end{subfigure}
     \vspace{-4mm}
    \caption{OpenAI Gym's \cite{brockman2016openai} Mujoco continuous-control environments used for evaluation.}
    \label{fig:mujoco_envs}
\end{figure}

We used 4 Mujoco continuous-control environments from OpenAI Gym \cite{brockman2016openai}, as shown in Fig. \ref{fig:mujoco_envs}. Expert trajectories were obtained by training a policy with SAC \cite{haarnoja2018soft}. We evaluated the benefit of using ARC with two popular Adversarial Imitation Learning (AIL) algorithms, $f$-MAX-RKL \cite{ghasemipour2020divergence} and GAIL \cite{ho2016generative}. For each of these algorithms, we evaluated the performance of standard AIL algorithms ($f$-MAX-RKL, GAIL), ARC aided AIL algorithms (ARC-$f$-MAX-RKL, ARC-GAIL) and Naive-Diff algorithm described in Section \ref{subsection:naive_diff} (Naive-Diff-$f$-MAX-RKL, Naive-Diff-GAIL). We also evaluated the performance of Behavior Cloning (BC). For standard AIL algorithms (GAIL and $f$-MAX-RKL) and BC, we used the implementation of \cite{ni2021f}. Further experimental details are presented in Appendix \ref{appendix:experimental_details}.

% Amongst all the algorithms we evaluated, ARC-$f$-MAX-RKL performed the best.

\vspace{-3mm}
\subsection{Imitation Learning in robotic manipulation tasks}
\vspace{-3mm}
\label{section:results_robotic_sim}
We used simplified 2D versions of FetchReach (Fig. \ref{fig:fetch_reach}) and FetchPush (Fig. \ref{fig:fetch_push}) robotic manipulation tasks from OpenAI Gym \cite{brockman2016openai} which have a simulated Fetch robot, \cite{wise2016fetch}. In the FetchReach task, the robot needs to take it's end-effector to the goal (virtual red sphere) as quickly as possible. In the FetchPush task, the robot's needs to push the block to the goal as quickly as possible. We used hand-coded proportional controller to generate expert trajectories for these tasks. Further details are presented in Appendix \ref{appendix:experimental_details_robotic_sim}.

% \vspace{-5mm}
\newcommand{\plotwidth}{0.23\textwidth}
\begin{figure}[t]
     \vspace{-2mm}
     \centering
     \begin{subfigure}[b]{\textwidth}
         \centering
         \includegraphics[width=\textwidth]{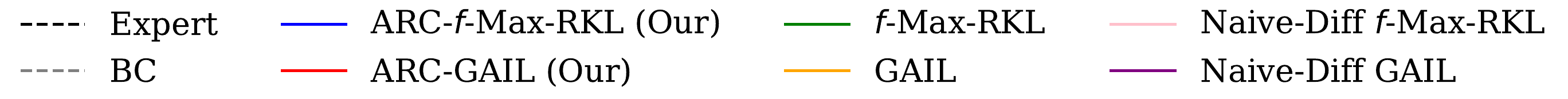}
         \vspace{-4mm}
     \end{subfigure}
     \begin{subfigure}[b]{\plotwidth}
         \centering
         \includegraphics[width=\textwidth]{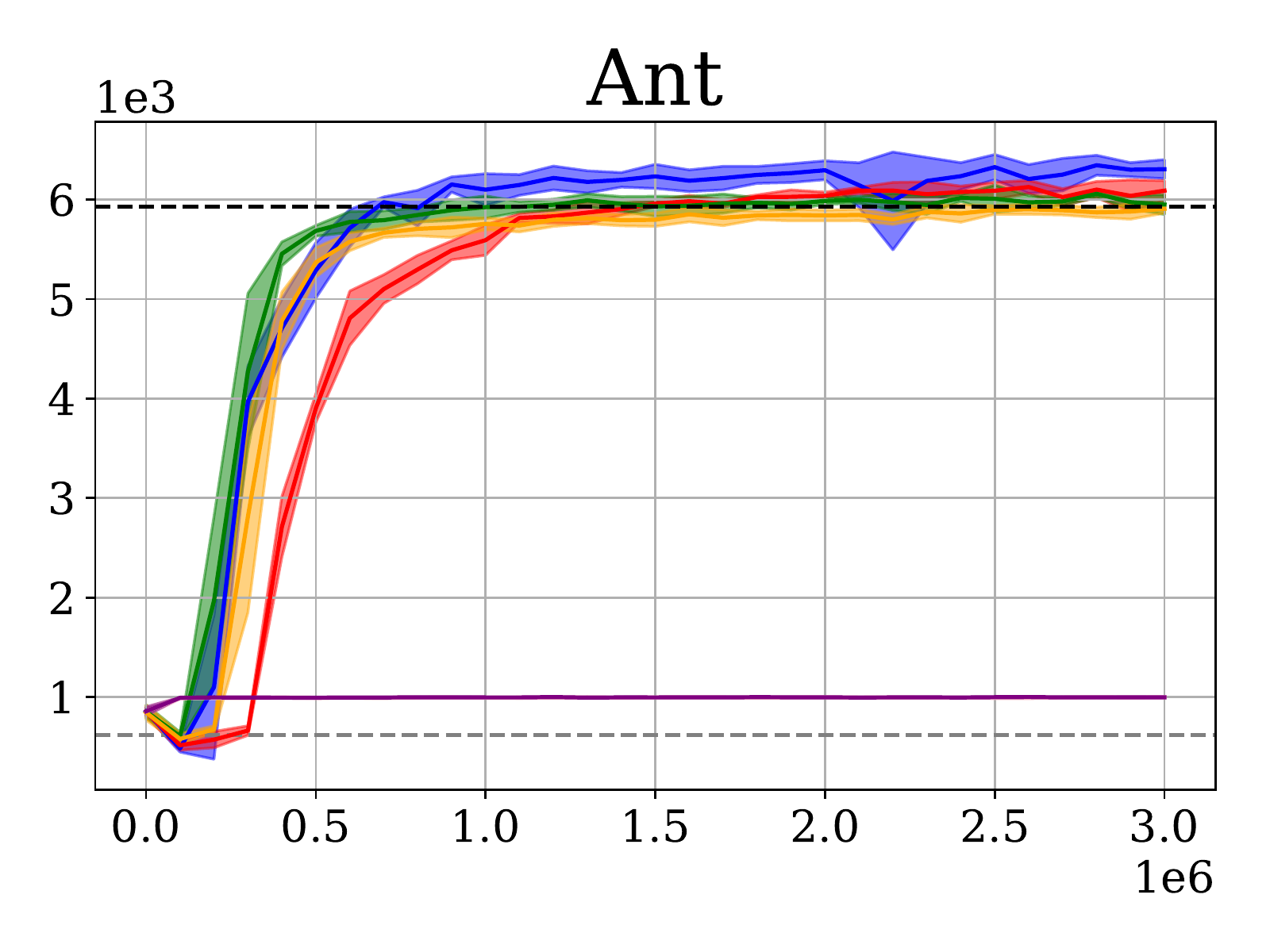}
     \end{subfigure}
     \hfill
     \begin{subfigure}[b]{\plotwidth}
         \centering
         \includegraphics[width=\textwidth]{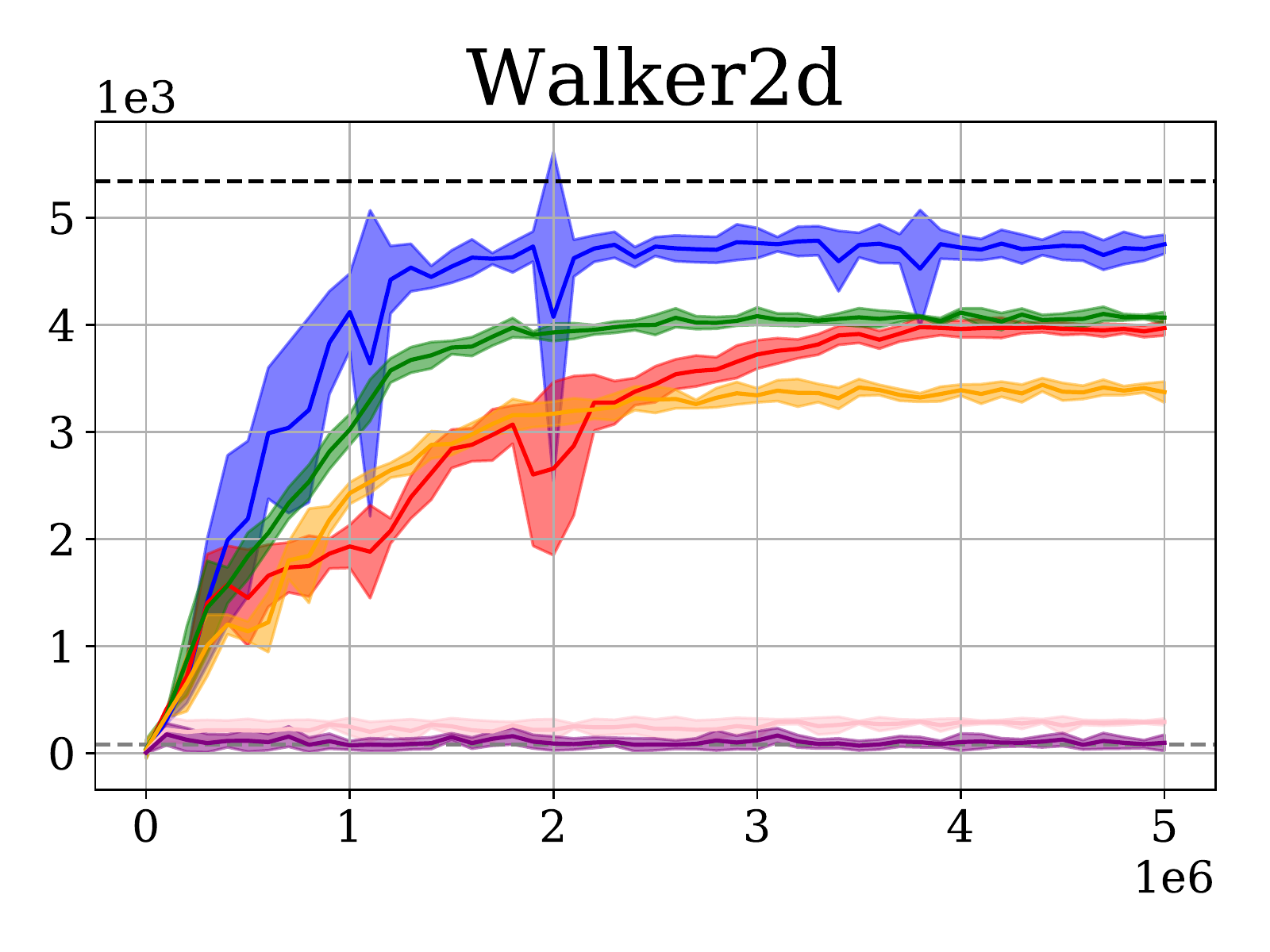}
     \end{subfigure}
     \hfill
     \begin{subfigure}[b]{\plotwidth}
         \centering
         \includegraphics[width=\textwidth]{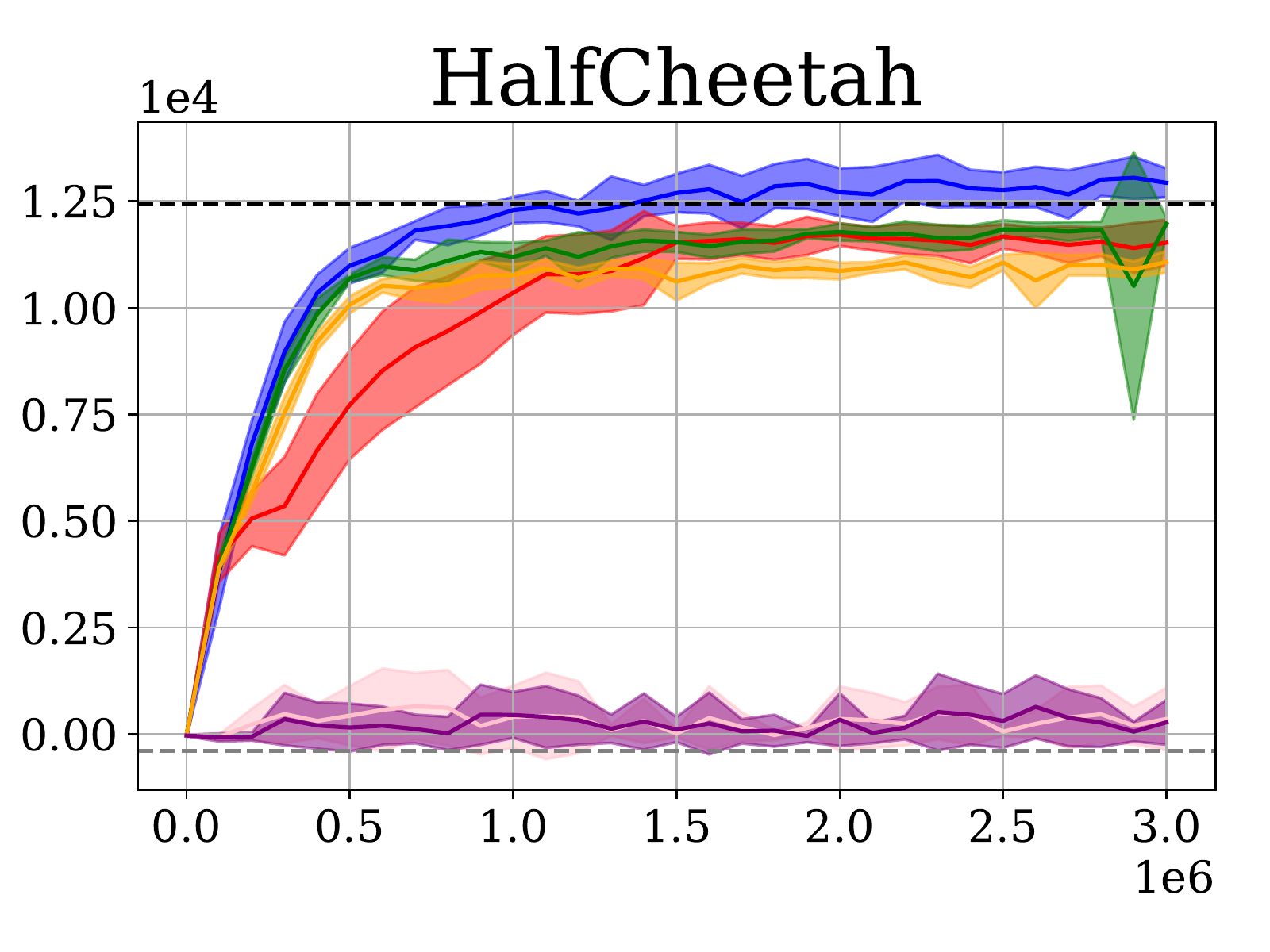}
     \end{subfigure}
     \hfill
     \begin{subfigure}[b]{\plotwidth}
         \centering
         \includegraphics[width=\textwidth]{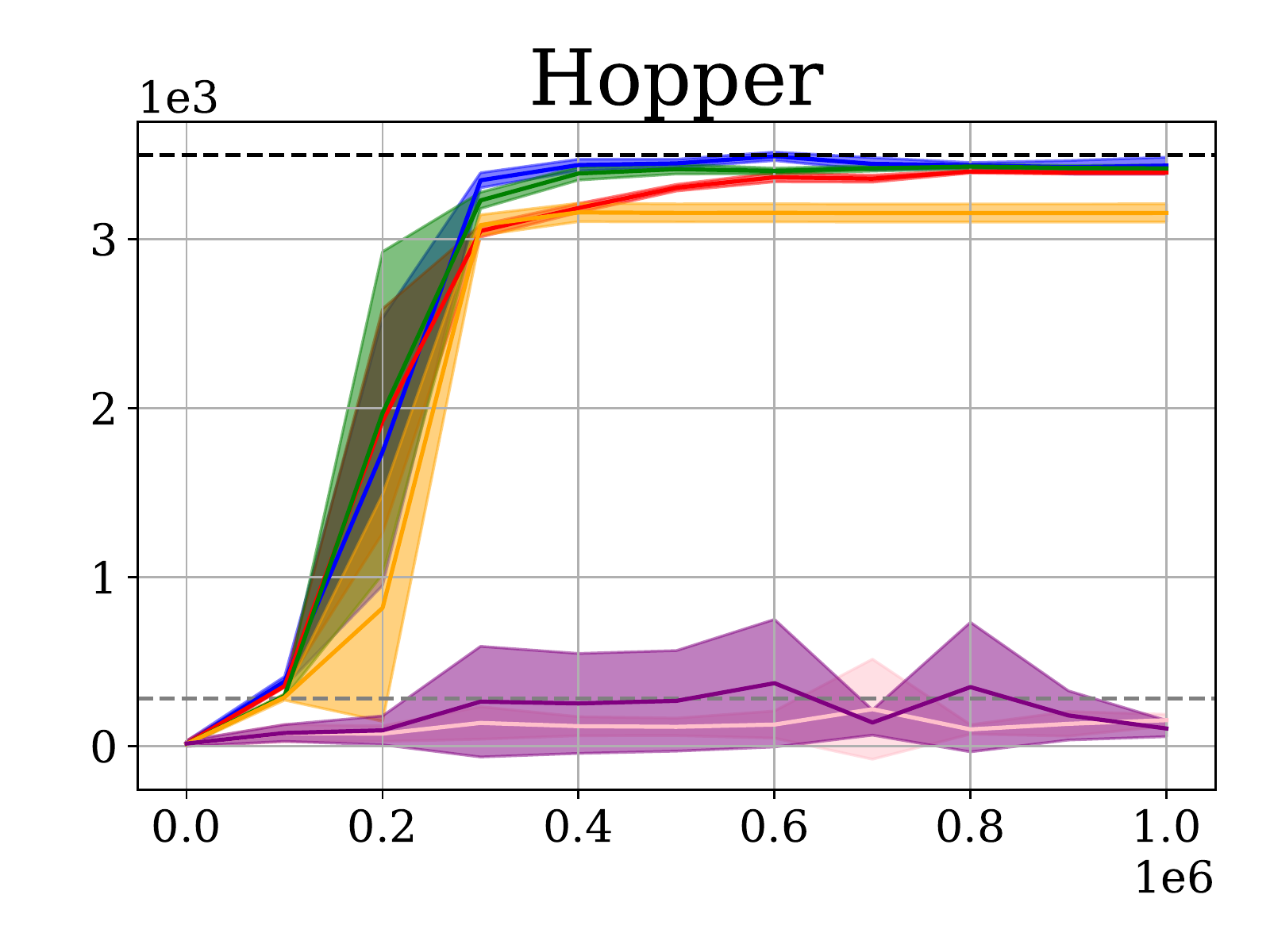}
        %  \caption{$Q^* = r^* + C^*$}
        %  \label{fig:policy_iteration_grid_q}
     \end{subfigure}
     \vspace{-5mm}
     \caption{Episode return versus number of environment interaction steps for different Imitation Learning algorithms on Mujoco continuous-control environments.}
     \label{fig:mujoco_training_plots}
    %  \vspace{-0.5cm}
\end{figure}
% \vspace{-3mm}
\begin{table}[h!]
    % \vspace{2mm}
    % \vspace{-2cm}
    \centering
    \scriptsize
    \begin{tabular}{c|c|c|c|c}
        \toprule
         Method &
         \multicolumn{1}{c|}{Ant} & \multicolumn{1}{c|}{Walker2d} & \multicolumn{1}{c|}{HalfCheetah} &
         \multicolumn{1}{c}{Hopper} \\
        \midrule
        Expert return &
  \multicolumn{1}{c|}{5926.18 $\pm$ 124.56} & \multicolumn{1}{c|}{5344.21 $\pm$ 84.45} & \multicolumn{1}{c|}{12427.49 $\pm$ 486.38} &        
  \multicolumn{1}{c}{3592.63 $\pm$ 19.21} \\
        %  \# Expert traj
        %   & 1 & 4 & 16
        %   & 1 & 4 & 16
        %   & 1 & 4 & 16
        %   & 1 & 4 & 16 \\
       \midrule
    ARC-$f$-Max-RKL (Our) & \textbf{6306.25} $\pm$ 95.91 & \textbf{4753.63} $\pm$ 88.89 & \textbf{12930.51} $\pm$ 340.02 & \textbf{3433.45} $\pm$ 49.48 \\  
    $f$-Max-RKL & 5949.81 $\pm$ 98.75 & 4069.14 $\pm$ 52.14 & 11970.47 $\pm$ 145.65 & 3417.29 $\pm$ 19.8 \\ 
    Naive-Diff $f$-Max-RKL & 998.27 $\pm$ 3.63 & 294.36 $\pm$ 31.38 & 357.05 $\pm$ 732.39 & 154.57 $\pm$ 34.7 \\ 
    \midrule
    ARC-GAIL (Our) & \textbf{6090.19} $\pm$ 99.72 & \textbf{3971.25} $\pm$ 70.11 & \textbf{11527.76} $\pm$ 537.13 & \textbf{3392.45} $\pm$ 10.32 \\
    GAIL & 5907.98 $\pm$ 44.12 & 3373.26 $\pm$ 98.18 & 11075.31 $\pm$ 255.69 & 3153.84 $\pm$ 53.61 \\ 
    Naive-Diff GAIL & 998.17 $\pm$ 2.22 & 99.26 $\pm$ 76.11 & 277.12 $\pm$ 523.77 & 105.3 $\pm$ 48.01 \\
    \midrule
    BC & 615.71 $\pm$ 109.9 & 81.04 $\pm$ 119.68 & -392.78 $\pm$ 74.12 & 282.44 $\pm$ 110.7 \\ 
        \bottomrule
    \end{tabular}
    \vspace{3mm}
    \caption{Policy return on Mujoco environments using different Imitation Learning algorithms. Each algorithm is run with 10 random seeds. Each seed is evaluated for 20 episodes.}
    \label{tab:mujoco_performance}
    \vspace{-5mm}
\end{table}
Fig. \ref{fig:mujoco_training_plots} shows the training plots and Table \ref{tab:mujoco_performance} shows the final performance of the different algorithms. Across all environments and across both the AIL algorithms, incorporating ARC shows consistent improvement over standard AIL algorithms (Table \ref{tab:mujoco_performance}). BC suffers from distribution shift at test time \cite{ross2011reduction,ho2016generative} and performs very poorly.
As we predicted in Section \ref{subsection:naive_diff}, Naive-Diff algorithms don't perform well as naively using autodiff doesn't compute the gradients correctly.

\vspace{-1mm} 
Fig. \ref{fig:sim_return_training_plot} shows the training plots and Table \ref{tab:robotic_task_performance} under the heading `Simulation' shows the final performance of the different algorithms. In both the FetchReach and FetchPush tasks, ARC aided AIL algorithms consistently outperformed the standard AIL algorithms.
% Amongst all the evaluated algorithms, ARC-$f$-Max-RKL performed the best in the FetchReach task and ARC-GAIL performed the best in the FetchPush task.
Fig. \ref{fig:action_vs_timestep} shows the magnitude of the $2^{nd}$ action dimension vs. time-step in one episode for different algorithms. The expert initially executed large actions when the end-effector/block was far away from the goal. As the end-effector/block approached the goal, the expert executed small actions. ARC aided AIL algorithms (ARC-$f$-Max-RKL and ARC-GAIL) showed a similar trend while standard AIL algorithms ($f$-Max-RKL and GAIL) learnt a nearly constant action. Thus, ARC aided AIL algorithms were able to better imitate the expert than standard AIL algorithms.

\vspace{-4mm}
\subsection{Sim-to-real transfer of robotic manipulation policies}
\begin{figure}[h!]
    \centering
    \begin{subfigure}[b]{\subfigwidth}
         \centering
         \includegraphics[width=\textwidth]{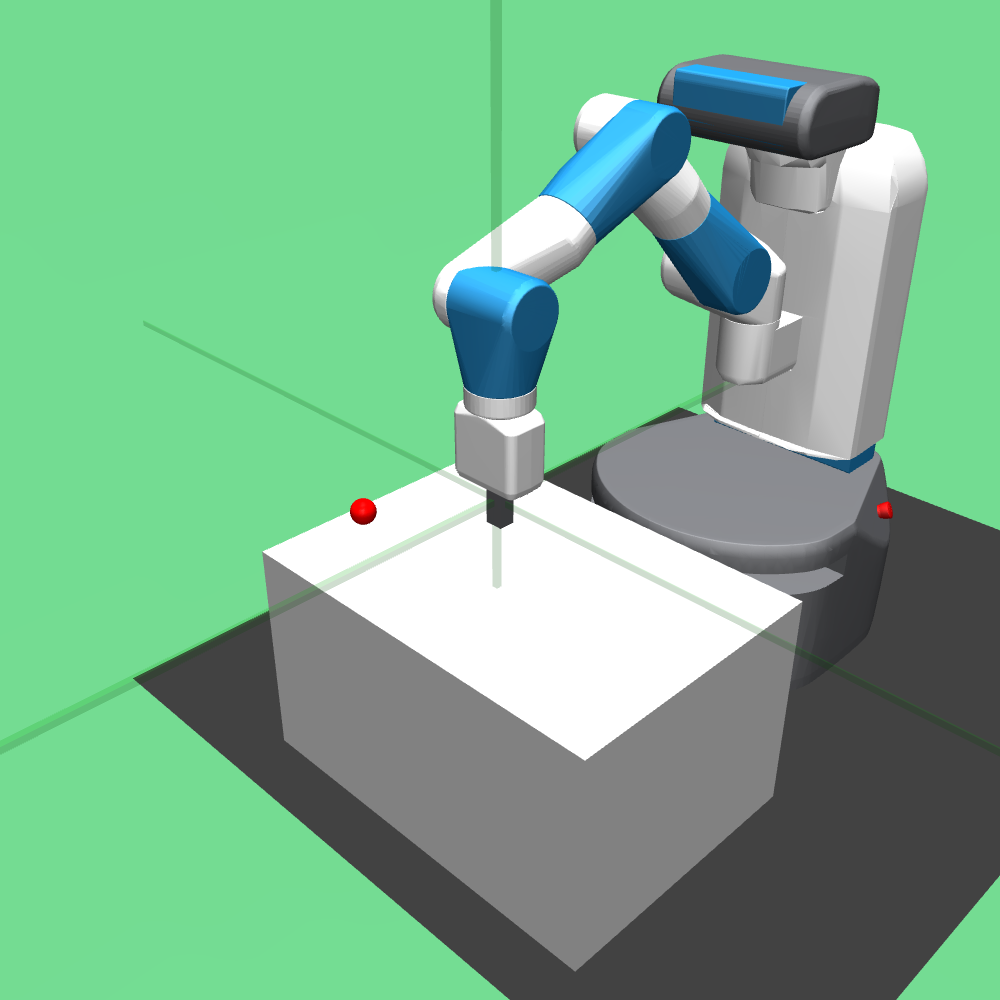}
         \vspace{-4mm}
         \caption{FetchReach}
         \label{fig:fetch_reach}
     \end{subfigure}
     \begin{subfigure}[b]{\subfigwidth}
         \centering
         \includegraphics[width=\textwidth]{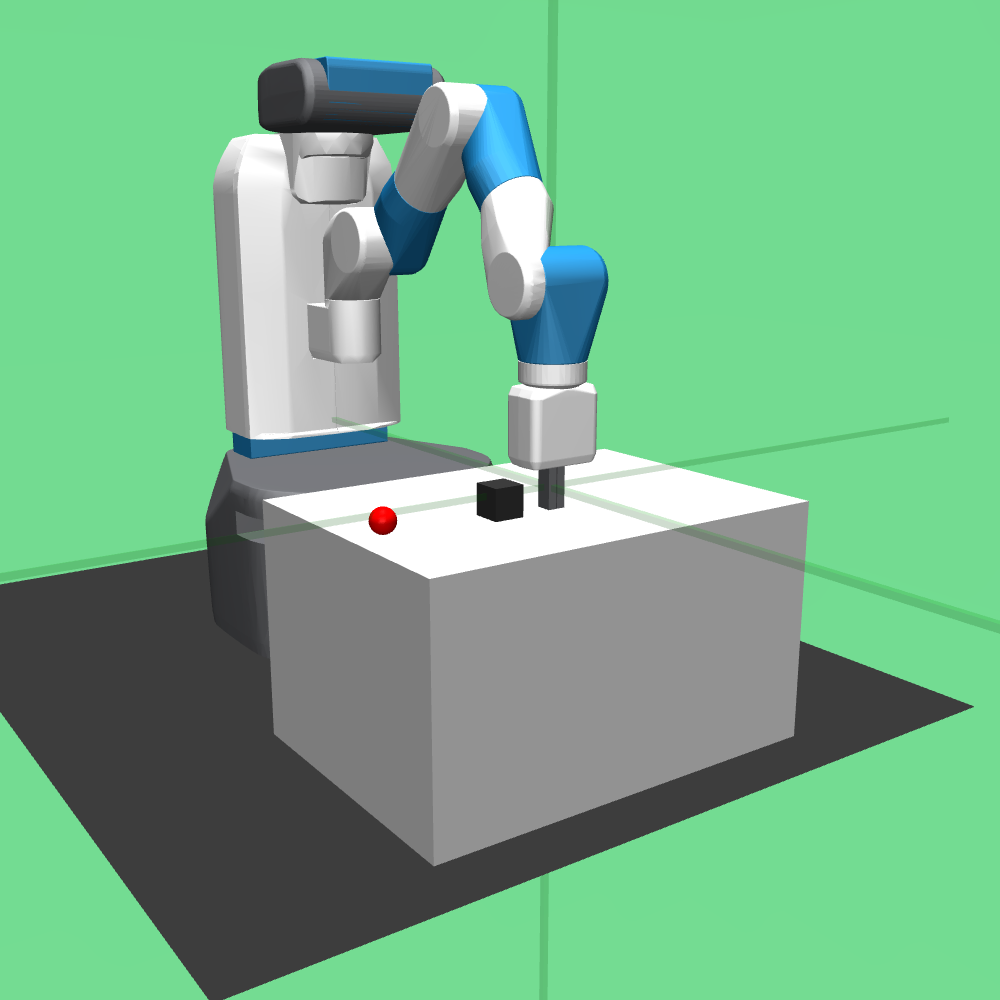}
         \vspace{-4mm}
         \caption{FetchPush}
         \label{fig:fetch_push}
     \end{subfigure}
     \begin{subfigure}[b]{\subfigwidth}
         \centering
         \includegraphics[width=\textwidth]{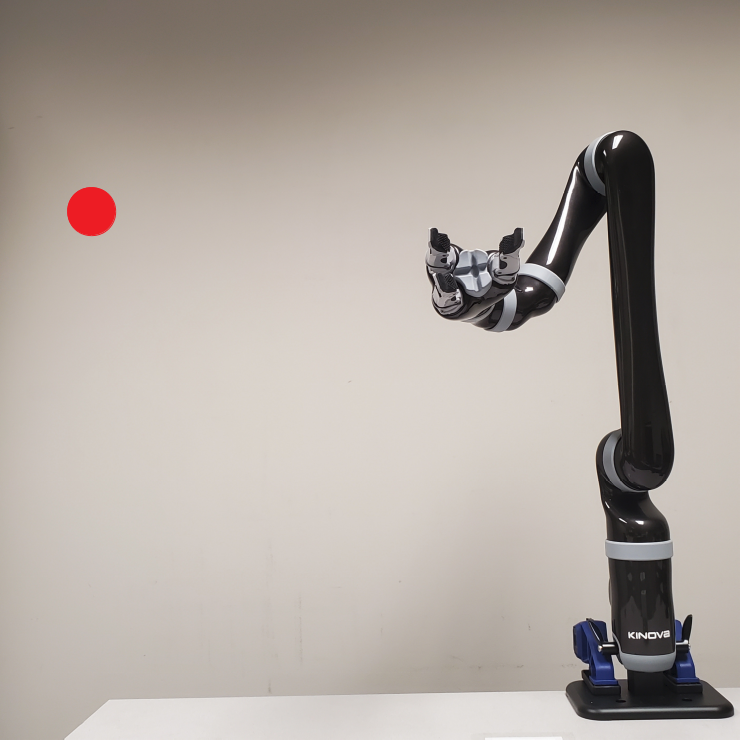}
         \vspace{-4mm}
         \caption{JacoReach}
         \label{fig:jaco_reach}
     \end{subfigure}
     \begin{subfigure}[b]{\subfigwidth}
         \centering
         \includegraphics[width=\textwidth]{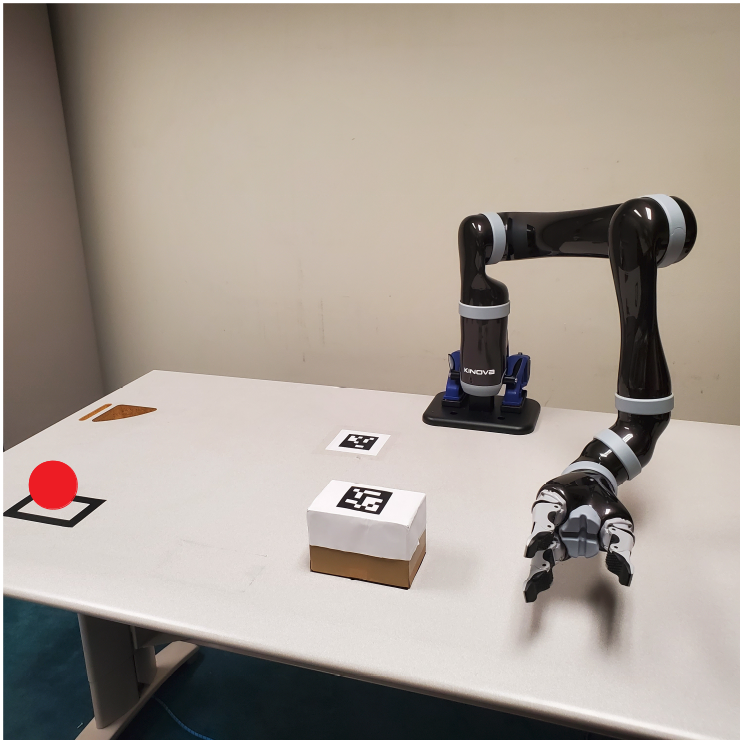}
         \vspace{-4mm}
         \caption{JacoPush}
         \label{fig:jaco_push}
     \end{subfigure}
     \vspace{-4mm}
    \caption{Simulated and real robotic manipulation tasks used for evaluation. Simplified 2D versions of the FetchReach \subref{fig:fetch_reach} and FetchPush \subref{fig:fetch_push} tasks from OpenAI Gym, \cite{brockman2016openai} with a Fetch robot, \cite{wise2016fetch}. Corresponding JacoReach \subref{fig:jaco_reach} and JacoPush \subref{fig:jaco_push} tasks with a real Kinova Jaco Gen 2 arm, \cite{campeau2019kinova}.} 
    \label{fig:robot_envs}
\end{figure}

\vspace{-3mm}
\label{section:results_sim_to_real}
For testing the sim-to-real transfer of the different trained AIL manipulation policies, we setup JacoReach (Fig. \ref{fig:jaco_reach}) and JacoPush (Fig. \ref{fig:jaco_push}) tasks with a Kinova Jaco Gen 2 arm, similar to the FetchReach and FetchPush tasks in the previous section. The details are presented in Appendix \ref{appendix:sim_to_real_environment}.

Table \ref{tab:robotic_task_performance} under the heading `Real Robot' shows the performance of the different AIL algorithms in the real robotic manipulation tasks. The real robot evaluations showed a similar trend as in the simulated tasks. ARC aided AIL consistently outperformed the standard AIL algorithms.
% Amongst all the evaluated algorithms, ARC-$f$-Max-RKL performed the best in both the JacoReach and JacoPush tasks.
 Appendix \ref{appendix:additional_results} Fig. \ref{fig:jaco_final_position} visualizes the policies in the JacoPush task showing that ARC aided AIL algorithms were able to push the block closer to the goal as compared to the standard AIL algorithms. Project website contains videos of the same. Since we didn't tune hyper-parameters for these tasks (both our methods and the baselines, details in Appendix \ref{appendix:experimental_details_robotic_sim}), it is likely that the performances would improve with further parameter tuning. Without fine-tuning hyper-parameters for these tasks, ARC algorithms showed higher performance than the baselines. This shows that ARC algorithms are parameter robust and applicable to real robot tasks without much fine tuning.

\newcommand{\doubleplotwidth}{0.46\textwidth}
\begin{figure}[h!]
     \vspace{-3mm}
     \centering
     \begin{subfigure}[b]{0.95\textwidth}
         \centering
         \includegraphics[width=\textwidth]{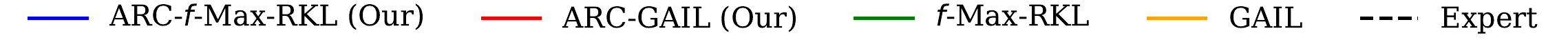}
     \end{subfigure}
     \begin{subfigure}[b]{\doubleplotwidth}
         \centering
         \includegraphics[width=0.49\textwidth]{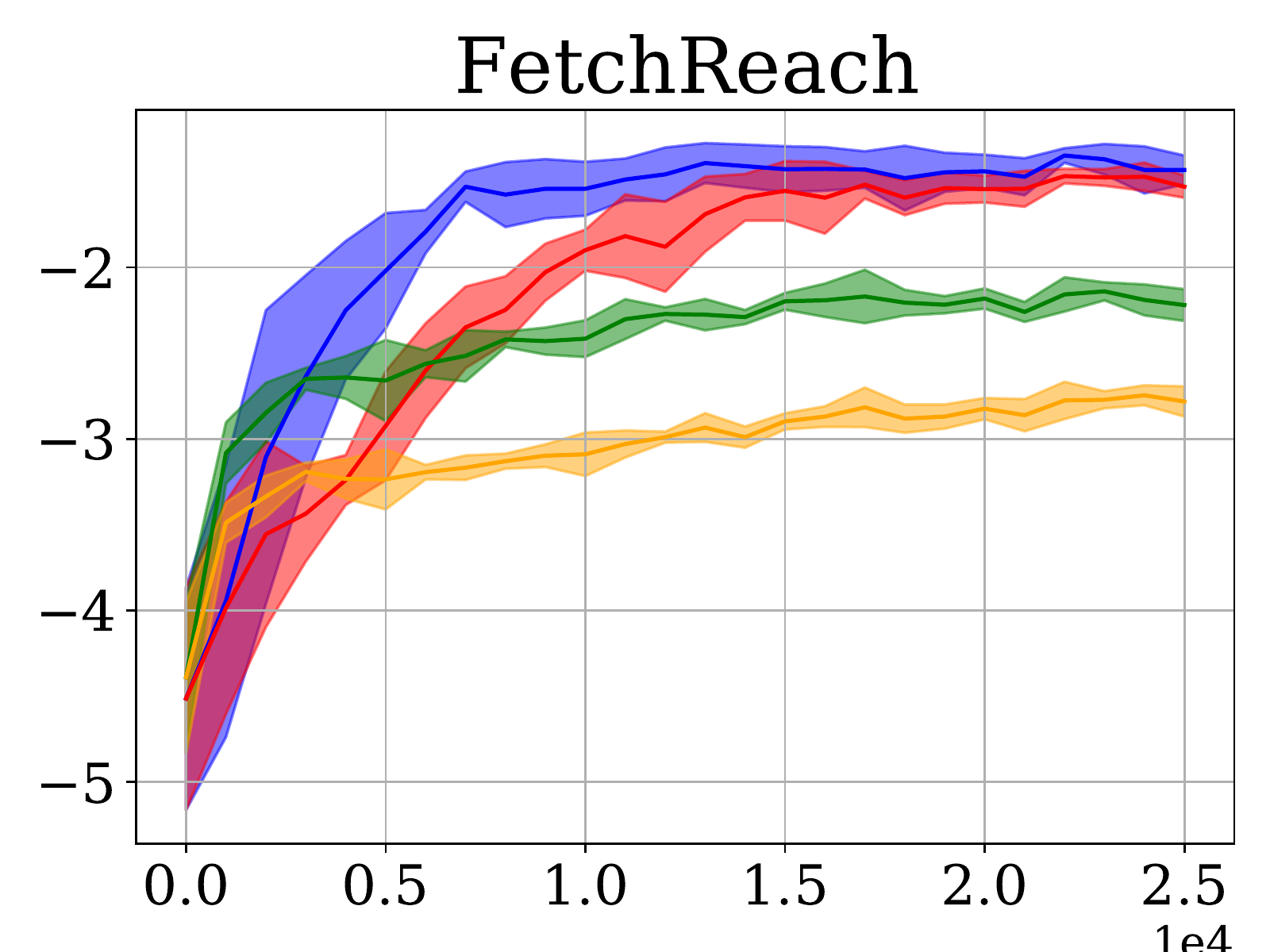}
         \centering
         \includegraphics[width=0.49\textwidth]{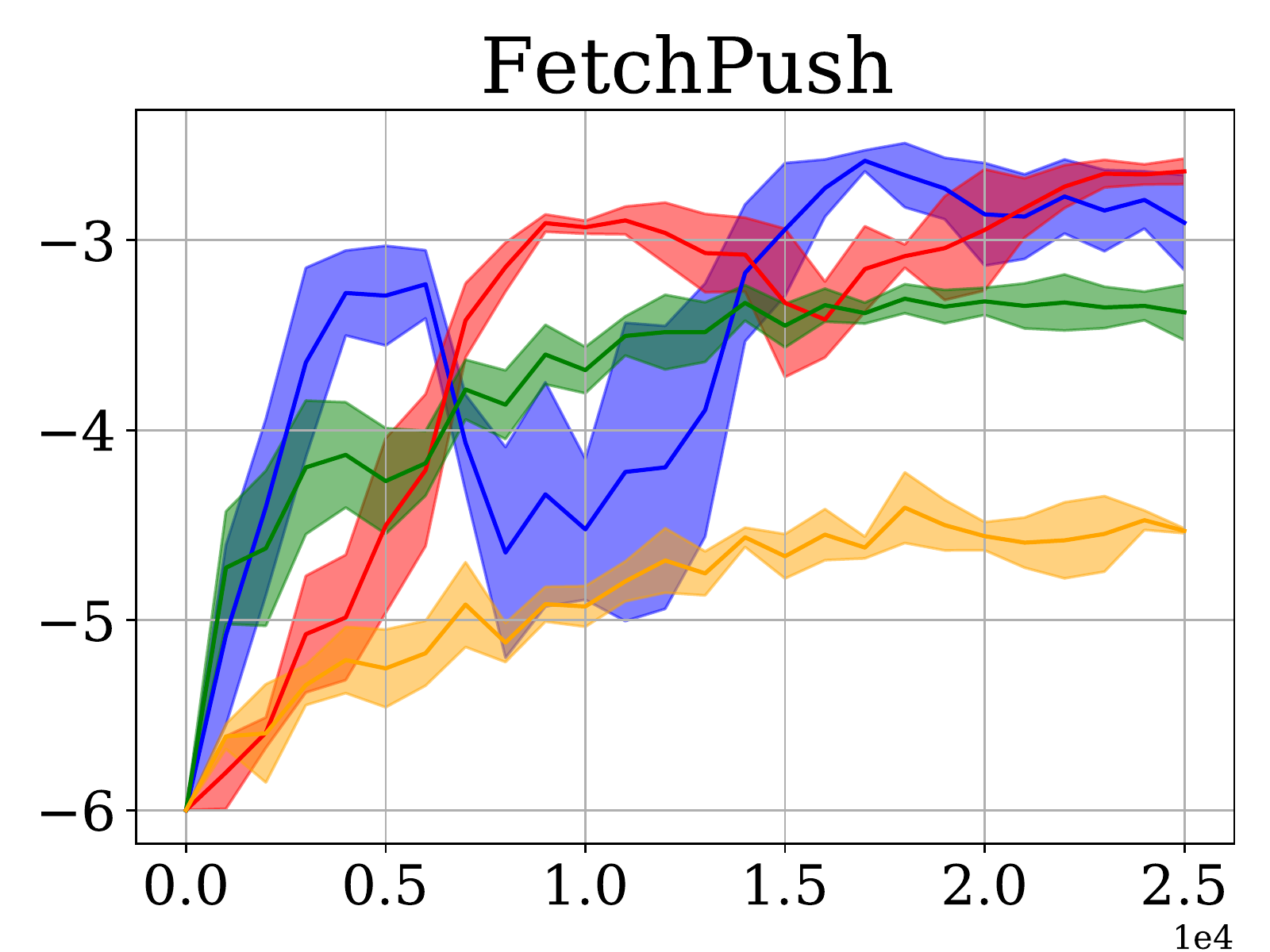}
         \vspace{-6mm}
         \subcaption{Episode return vs. interaction steps}
         \label{fig:sim_return_training_plot}
     \end{subfigure}
     \qquad
     \begin{subfigure}[b]{\doubleplotwidth}
         \centering
         \includegraphics[width=0.49\textwidth]{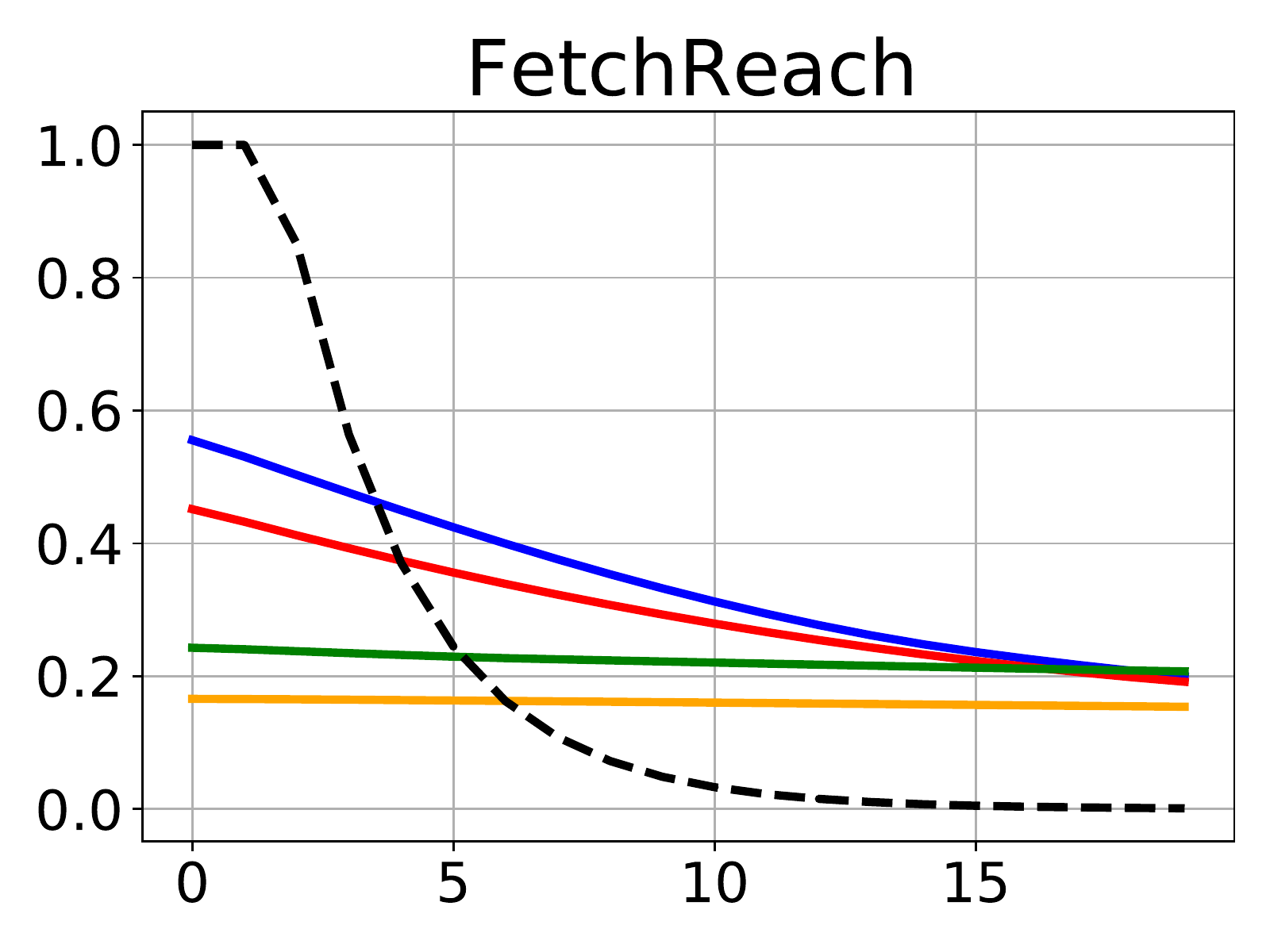}
         \centering
         \includegraphics[width=0.49\textwidth]{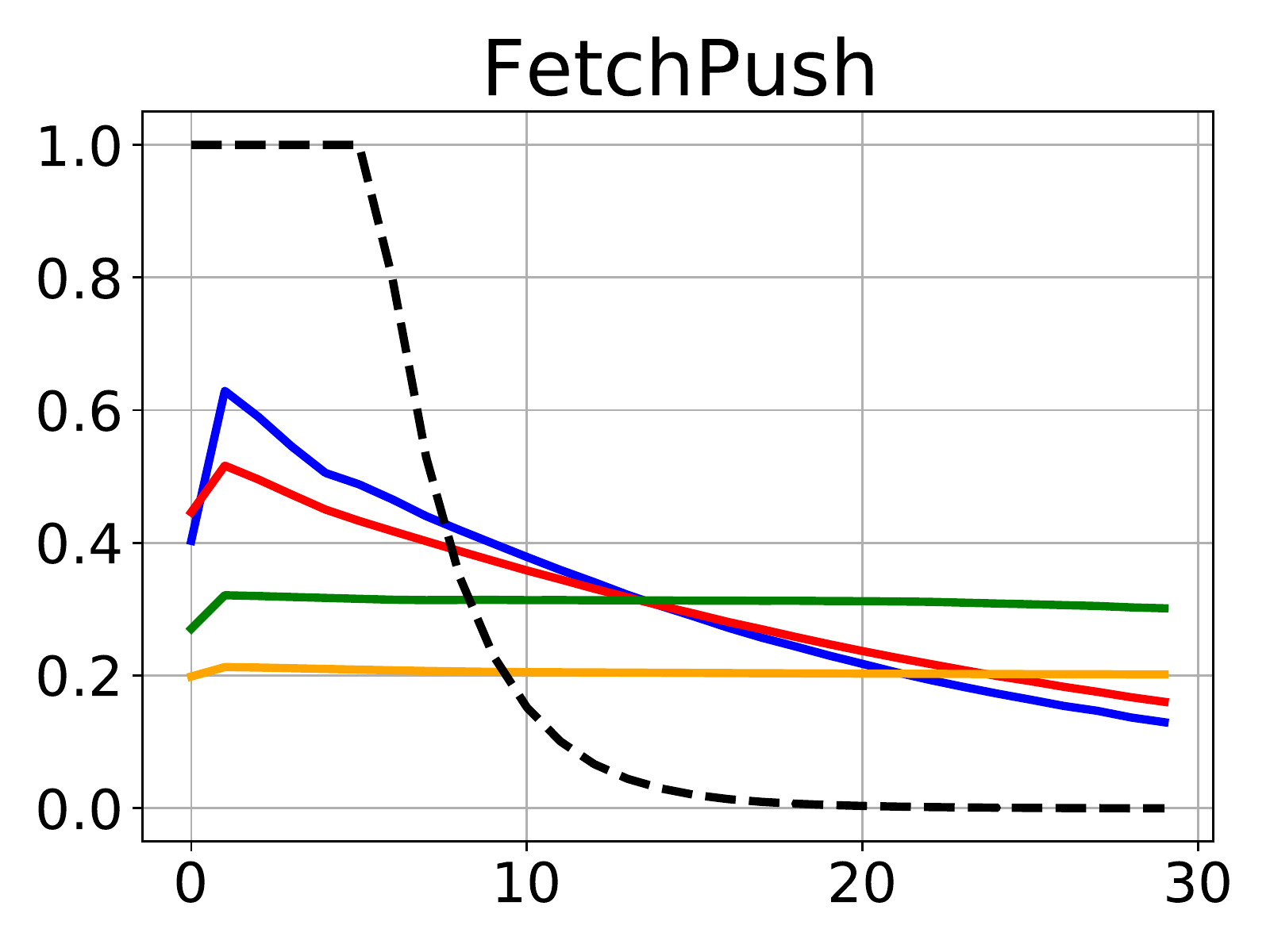}
         \vspace{-6mm}
         \subcaption{Action vs. time step}
         \label{fig:action_vs_timestep}
     \end{subfigure}
    %  \vspace{-1mm}
     \caption{\subref{fig:sim_return_training_plot} Episode return vs. number of environment interaction steps for different Adversarial Imitation Learning algorithms on FetchPush and FetchReach tasks. \subref{fig:action_vs_timestep} Magnitude of the $2^{nd}$ action dimension versus time step in a single episode for different algorithms.
    %  ARC aided AIL algorithms executed actions that resembled the expert actions more closely than standard AIL algorithms did.
     }
     \label{fig:robot_training_plots}
     \vspace{-3mm}
\end{figure}

% \vspace{-1mm}
\begin{table}[h!]
    % \vspace{2mm}
    \vspace{-2mm}
    \centering
    \scriptsize
    \begin{tabular}{c|c|c|c|c}
        \toprule
        &
         \multicolumn{2}{c|}{\textbf{Simulation}} &
        %  \multicolumn{1}{c|}{FetchPush} &
        %  \multicolumn{1}{c|}{JacoReach} &
         \multicolumn{2}{c}{\textbf{Real Robot}}\\
         Method &
         \multicolumn{1}{c}{FetchReach} & \multicolumn{1}{c|}{FetchPush} &
         \multicolumn{1}{c}{JacoReach} &
         \multicolumn{1}{c}{JacoPush}\\
        \midrule
        Expert return &
        \multicolumn{1}{c|}{-0.58 $\pm$ 0} & \multicolumn{1}{c|}{-1.18 $\pm$ 0.04}&
        -0.14 $\pm$ 0.01& 
        -0.77 $\pm$ 0.01\\
        \midrule
    ARC-$f$-Max-RKL (Our) & \textbf{-1.43} $\pm$ 0.08 & \textbf{-2.91} $\pm$ 0.25 & \textbf{-0.38} $\pm$ 0.02 & \textbf{-1.25} $\pm$ 0.06\\ 
    $f$-Max-RKL & -2.22 $\pm$ 0.09 & -3.38 $\pm$ 0.15 & -0.8 $\pm$ 0.05 & -2.03 $\pm$ 0.06 \\ 
    \midrule
    ARC-GAIL (Our) & \textbf{-1.53} $\pm$ 0.06 & \textbf{-2.64} $\pm$ 0.07 & \textbf{-0.46} $\pm$ 0.01 & \textbf{-1.56} $\pm$ 0.08\\ 
    GAIL & -2.78 $\pm$ 0.09 & -4.53 $\pm$ 0.01 & -1.05 $\pm$ 0.06 &-2.35 $\pm$ 0.06 \\ 
        \bottomrule
    \end{tabular}
    % \vspace{18mm}
    \caption{Policy return on simulated (FetchReach, FetchPush) and real (JacoReach, JacoPush) robotic manipulation tasks using different AIL algorithms. The reward at each time step is negative distance between end-effector \& goal for reach tasks and block \& goal for push tasks. The reward in the real and simulated tasks are on different scales due to implementation details described in Appendix \ref{appendix:sim_to_real_environment}.}
    \label{tab:robotic_task_performance}
    \vspace{-3mm}
\end{table}

% In each of the reach tasks, FetchReach \subref{fig:fetch_reach} \& JacoReach \subref{fig:jaco_reach}, a robot's end-effector needs to reach a goal location shown as a virtual red sphere as quicky as possible. In each of the push tasks, FetchPush \subref{fig:fetch_push} \& JacoPush \subref{fig:jaco_push}, a robot needs to push a block to a goal location as quickly as possible.

    \vspace{-5mm}
\section{Limitations}
\vspace{-3mm}
Three main limitations in our work are: (1) While many AIL algorithms can be trained using expert `states' only, ARC-AIL can only be trained with `state-action' $(s,a)$ pairs. There are several scenarios where obtaining $(s,a)$ pairs is challenging (e.g. kinesthetic teaching). In such scenarios, ARC is not directly applicable. People often use tricks to mitigate this issue and using $(s,a)$ pairs to train a policy is a popular choice \cite{young2020visual, peng2020learning,lu2021aw,scheel2021urban,hoque2022thriftydagger}. (2) ARC-AIL can only work with continuous action space. Most real world robotic tasks have or can be modified to have a continuous action space. (3) We haven't explored how the agent-adversary interaction in AIL affects the accuracy of the reward gradient and leave that for future work.
    % \section{Discussion}
% We believe that the main reason ARC aided AIL outperforms standard AIL is because of the fact that ARC leverages the exact gradient of the reward instead of approximating it within the $Q$ function.
\vspace{-3mm}
\section{Conclusion}
\vspace{-4mm}
We highlighted that the reward in popular Adversarial Imitation Learning (AIL) algorithms are differentiable but this property has not been leveraged by existing model-free RL algorithms to train a policy. Further, they are usually shaped.
We also showed that naively differentiating the policy through this reward function does not perform well.
To solve this issue, we proposed a class of Actor Residual Critic (ARC) RL algorithms that use a $C$ function as an alternative to standard Actor Critic (AC) algorithms which use a $Q$ function. An ARC algorithm can replace the AC algorithm in any existing AIL algorithm. We formally proved that Policy Iteration using $C$ function converges to an optimum policy in tabular environments.
For continuous-control tasks, using ARC can compute the exact gradient of the policy through the reward function which helps improve the performance of the AIL algorithms in simulated continuous-control and simulated \& real robotic manipulation tasks. Future work can explore the applicability of ARC algorithm to other scenarios which have a differentiable reward function.

%===============================================================================

% The maximum paper length is 8 pages excluding references and acknowledgements, and 10 pages including references and acknowledgements

\clearpage
% The acknowledgments are automatically included only in the final and preprint versions of the paper.
\acknowledgments{We are thankful to Swaminathan Gurumurthy and Tejus Gupta for several insightful discussions on the idea. We are also thankful to Rohit Jena, Advait Gadhikar for their feedback on the manuscript and Dana Hughes, Sushmita Das for their support with some logistics of the project.

Finally, we are thankful to the reviewers for their constructive feedback through the rebuttal period, which we believe helped strengthen the paper.

This work has been supported by the following grants: Darpa HR001120C0036, AFRL/AFOSR FA9550-18-1-0251 and ARL W911NF-19-2-0146.
}

%===============================================================================

\bibliography{references}  % .bib

\clearpage
\appendix
\section{Accuracy of gradient}

\subsection{Error in gradient of an approximate function}
\label{appendix:abritrary_large_error_in_gradient}
\setcounter{theorem}{1}
\begin{theorem}
\label{theorem:gradient_arbitrary_large_error}
The error in gradient of an approximation of a differentiable function can be arbitrarily large even if the function approximation is accurate (but not exact). Formally, for any differentiable function $f(x):A \rightarrow B$, any small value of $\epsilon>0$ and any large value of $D>0$, we can have an approximation $\hat{f}(x)$ s.t. the following conditions are satisfied:

\begin{align}
    & \abs{\hat{f}(x)-f(x)} \leq \epsilon \quad \forall x \in A \quad \quad \text{(Accurate approximation)} \label{eq:function_accurate_approximation}\\
    & \abs{\nabla_x \hat{f}(x) - \nabla_x f(x)} \geq D \quad \text{for some } x \in A \quad \quad \text{(Arbitrarily large error in gradient)} \label{eq:gradient_arbitrary_large_error}
\end{align}

\begin{figure}[H]
    \centering
    \includegraphics[width=0.6\textwidth]{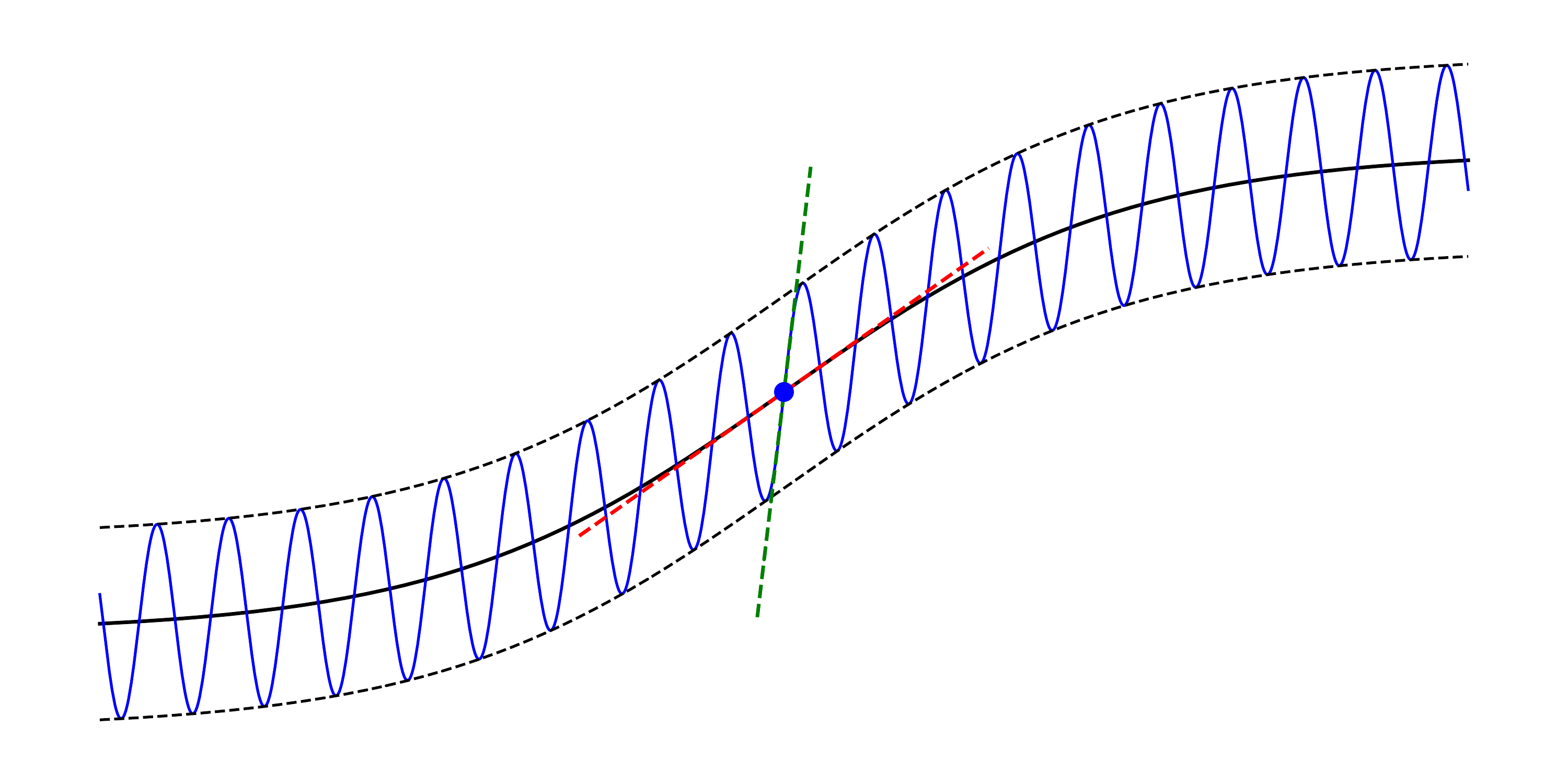}
    \includegraphics[width=\textwidth]{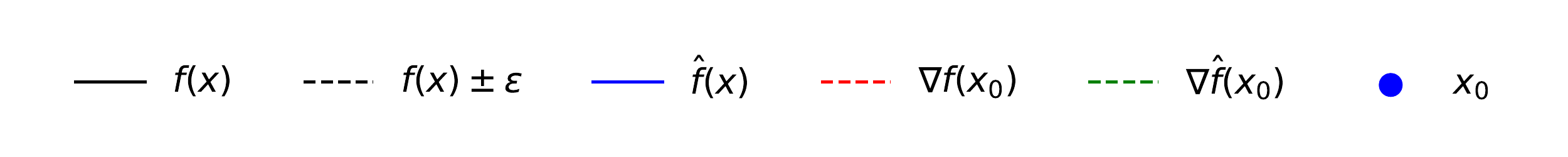}
    \vspace{-10mm}
    \caption{Illustrative example of a function approximation with accurate approximation, E.q.\ref{eq:function_accurate_approximation} but large error in gradient, E.q.\ref{eq:gradient_arbitrary_large_error}.}
    \label{fig:gradient_arbitrary_large_error}
\end{figure}

\begin{proof}

For any differentiable $f(x)$, $\epsilon>0$ and $D>0$, we can construct many examples of $\hat{f}(x)$ that satisfy the conditions in Eq. \ref{eq:function_accurate_approximation} and \ref{eq:gradient_arbitrary_large_error}. Here we show just one example that satisfies the 2 conditions. Let $x_0$ be any point $x_0 \in A$. We can choose $\hat{f}(x) = f(x) + \epsilon \sin{(b(x-x_0))}$, where $b=\frac{2D}{\epsilon}$. This is shown pictorially in Fig. \ref{fig:gradient_arbitrary_large_error}.

The error in function approximation is:
\begin{align*}
    \abs{\hat{f}(x)-f(x)} &= \abs{\epsilon \sin{b(x-x_0)}} = \epsilon \abs{\sin{b(x-x_0)}}\\
    &\leq \epsilon \quad \because \sin(x) \in [-1,1], \forall x \in \mathbb{R}
\end{align*}
Thus, $\hat{f}(x)$ satisfies Eq. \ref{eq:function_accurate_approximation} and approximates $f(x)$ accurately.

The error in gradient at $x_0$ is:
\begin{align*}
    \abs{\nabla_x \hat{f}(x) - \nabla_x f(x)} \Bigg|_{x=x_0} &= \abs{\nabla_xf(x) + \epsilon b \cos{(b(x-x_0))} - \nabla_x f(x)} \Bigg|_{x=x_0} \\
    &= \epsilon b \abs{\cos{(b(x_0-x_0))}}\\
    &= \epsilon \frac{2D}{\epsilon} \abs{\cos{(0)}} \quad \because b = \frac{2D}{\epsilon} \text{ and } \cos(0) = 1\\
    &= 2D > D
\end{align*}
Thus, $\hat{f}(x)$ satisfies Eq.\ref{eq:gradient_arbitrary_large_error}, i.e. the error in gradient can be arbitrarily large even if function approximation is accurate. We can also see visually from Fig. \ref{fig:gradient_arbitrary_large_error} that although $\abs{\hat{f}(x)-f(x)} < \epsilon$, there is a large difference between $\nabla \hat{f}(x_0)$ and $\nabla f(x_0)$.
\end{proof}
\end{theorem}

\subsection{Decomposition in ARC leads to more accurate gradient for AIL}
\label{appendix:arc_snr}

\begin{figure}[h!]
    \centering
    \includegraphics[width=0.7\textwidth]{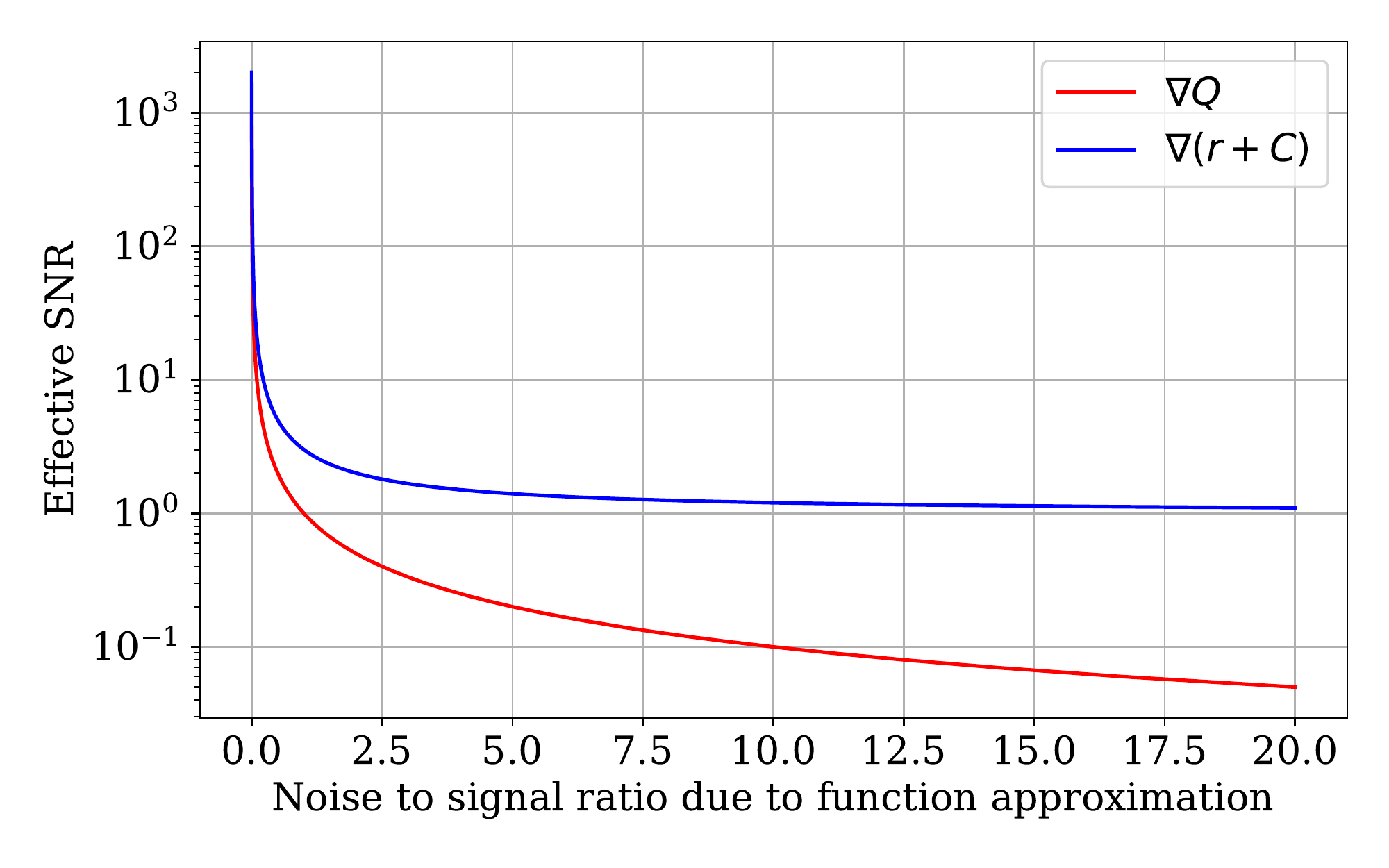}
    \caption{Signal to Noise Ratio (SNR) in Q gradient approximation as noise to signal ratio due to function approximation increases. Higher SNR is better. Using our proposed decomposition, $Q=r+C$, the effective SNR is higher than that without decomposition, when there is large noise due to function approximation.}
    \label{fig:gradient_snr}
\end{figure}

From Theorem \ref{theorem:gradient_arbitrary_large_error}, there is no bound on the error in gradient of an approximate function. Let $\hat{Q}$ and $\hat{C}$ denote the approximated $Q$ and $C$ values respectively. In the worst case, the gradients $\nabla_a \hat{Q}(s,a)$ and $\nabla_a \hat{C}(s,a)$ can both be completely wrong and act like random noise. Even in that case, the gradient obtained using our proposed decomposition ($Q=r+C$) would be useful because $\nabla_a r(s,a)$ is exact and hence $\nabla_a (r(s,a)+\hat{C}(s,a))$ would have useful information.

It is possible that the immediate “environment reward” is misleading which might hurt ARC. However, the “adversary reward” is a measure of closeness between agent and expert actions. It naturally is never misleading as long as we have a reasonably trained adversary. If we have an initial bad action that the expert takes to obtain a high reward later on, then the initial bad action will have a corresponding high adversary reward.

\newcommand{\snr}{\texttt{snr}}

In practice, we can expect both $\nabla_a \hat{Q}(s,a)$ and $\nabla_a \hat{C}(s,a)$ to have some finite noise. Signal to Noise (SNR) of a noisy signal is defined as the ratio of the magnitudes of true (signal strength) and noisy components in the noisy signal. If a signal $\hat{f} = f + \epsilon$ has a true signal component $f$ and a noisy component $\epsilon$, then the SNR is $\frac{\mathbb{E} f^2}{\mathbb{E} \epsilon^2}$. Higher SNR is better. SNR has been used in the past to analyze policy gradient algorithms \cite{robertssignal}.

Let us consider the case of a 1D environment which makes $\nabla_a r(s,a)$, $\nabla_a \hat{C}(s,a)$ and $\nabla_a \hat{Q}(s,a)$ scalars. 

\begin{enumerate}
    \item ``Signal strength of $\nabla_a r(s,a)" = \mathbb{E} [(\nabla_a r(s,a))^2] = S_r$ (say). Noise strength = $0$
    \item $\nabla_a \hat{C}(s,a) = \nabla_a C(s,a) + \epsilon_c \quad$   (i.e. True signal + noise)
    \item ``Signal strength of $\nabla_a \hat{C}(s,a)" = \mathbb{E} [(\nabla_a C(s,a))^2] = S_c$ (say)
    \item Let SNR of $\nabla_a \hat{C}(s,a) = \texttt{snr}_c$ (say).
    \item Noise strength = $\mathbb{E} [\epsilon_c^2] = S_n$ (say)
    \item  By definition of SNR, $\texttt{snr}_c = \frac{S_c}{S_n} \implies S_n = \frac{S_c}{\texttt{snr}_c}$
    \item 
    \begin{align}
    \text{Final signal}    &= \nabla_a r(s,a) + \nabla_a \hat{C}(s,a) \\
        &= \nabla_a r(s,a) + \nabla_a C(s,a) + \epsilon_c \\
        &= (\nabla_a r(s,a) + \nabla_a C(s,a)) + \epsilon_c \\
        &= \text{net true signal + net noise} 
    \end{align}
    \item 
    \begin{align}
    \text{Net signal strength}    &= \mathbb{E} [(\nabla_a r(s,a) + \nabla_a C(s,a))^2] \\
        &= \mathbb{E} [(\nabla_a r(s,a))^2] + \mathbb{E}[(\nabla_a C(s,a))^2] + 2\mathbb{E} [ \nabla_a r(s,a) \nabla_a C(s,a)] \\
        &= S_r + S_c + 2S_{r,C}
    \end{align}
    \item
    \begin{align}
     \text{Net SNR}   &= \frac{\text{Net signal strength}}{\text{Net noise strength}} \\
        &= \frac{S_r+S_c+2S_{r,C}}{S_n} \\
        &= \frac{S_r+S_c+2S_{r,C}}{ \frac{S_c}{\texttt{snr}_c} } \\
        &= \texttt{snr}_c \left(\frac{S_r}{S_c} + 1 + \frac{2S_{r,c}}{S_c} \right) \\ 
    \end{align}
\end{enumerate}

Let the SNR in $\nabla_a \hat{Q} (s,a)$ be $\texttt{snr}_Q$. Now, let's find when the net SNR in $\nabla_a \hat{C} (s,a)$ is higher than $\texttt{snr}_Q$, i.e. when does the decomposition lead to higher SNR.

\begin{align}
    \text{Net SNR in } \nabla_a \hat{C} (s,a) &\geq \texttt{snr}_Q\\
    \implies \texttt{snr}_c \left(\frac{S_r}{S_c} + 1 + \frac{2S_{r,c}}{S_c} \right) &\geq  \texttt{snr}_Q\\
    \implies \texttt{snr}_c \geq \frac{1}{\left(\frac{S_r}{S_c} + 1 + \frac{2S_{r,c}}{S_c} \right)} \texttt{snr}_Q \label{eq:snr_condition}
\end{align}

\textbf{Thus, Net SNR in $\nabla_a \hat{C} (s,a)$ is higher than $\texttt{snr}_Q$ if Eq. \ref{eq:snr_condition} holds true.}

Consider 3 cases:

\begin{itemize}
    \item Case 1: $S_{r,c} \geq 0$\\
    
    \textbf{What does this mean?}:\\
    $S_{r,c} = \mathbb{E} [ \nabla_a r(s,a) \nabla_a C(s,a)]$, this means $\nabla_a r(s,a)$ and $\nabla_a C(s,a)$ are positively correlated.\\
    
    \textbf{Implication}:\\
    $ \frac{S_r}{S_c} \geq 0$ since it is a ratio of signal strengths. Thus $\frac{S_r}{S_c} + 1 + \frac{2S_{r,c}}{S_c} \geq 1$ since we are adding non-negative terms to 1. Thus, $\frac{1}{ \frac{S_r}{S_c} + 1 + \frac{2S_{r,c}}{S_c}} \leq 1$. Let's call $\frac{1}{ \frac{S_r}{S_c} + 1 + \frac{2S_{r,c}}{S_c}} = $\texttt{fraction}. Thus, Eq. \ref{eq:snr_condition} reduces to $\texttt{snr}_c \geq \texttt{fraction} \times \texttt{snr}_Q$. 
    
    In other words, even if $\texttt{snr}_C$ is a certain fraction of $\texttt{snr}_Q$, the net SNR due to decomposition is higher than that without decomposition.
    
    \item Case 2: $-\frac{S_r}{2} \leq S_{r,c} < 0$\\
    
    \textbf{What does this mean?}:\\
    This means $\nabla_a r(s,a)$ and $\nabla_a C(s,a)$ are slightly negatively correlated.\\
    
    \textbf{Implication}:
    In this case,
    \begin{align}
        \frac{S_r}{S_c} + 1 + \frac{2S_{r,c}}{S_c}
        &\geq \frac{S_r}{S_c} + 1 + (\frac{2}{S_c})(\frac{-S_r}{2})\\
        \implies \frac{S_r}{S_c} + 1 + \frac{2S_{r,c}}{S_c}
        &\geq \frac{S_r}{S_c} + 1 - \frac{S_r}{S_c}\\
        \implies \frac{S_r}{S_c} + 1 + \frac{2S_{r,c}}{S_c}
        &\geq 1
    \end{align}
    Just like in case 1, we get the denominator in in Eq. \ref{eq:snr_condition} is a fraction. This in turn leads to the same conclusion that even if $\texttt{snr}_C$ is a certain fraction of $\texttt{snr}_Q$, the net SNR due to decomposition is higher than that without decomposition.
    
    \item Case 3: $S_{r,c} < -\frac{S_r}{2}$\\
    
    \textbf{What does this mean?}:\\
    This means $\nabla_a r(s,a)$ and $\nabla_a C(s,a)$ are highly negatively correlated.\\
    
    \textbf{Implication}:
    In this case, we get the denominator in Eq. \ref{eq:snr_condition} is $>1$. Decomposition would only help if $\texttt{snr}_c > \texttt{snr}_Q$ by the same factor.
\end{itemize}

\textbf{What determines relative values of $\texttt{snr}_c$  and $\texttt{snr}_Q$ in AIL?}

$\texttt{snr}_c$ and $\texttt{snr}_Q$ arise from noise in gradient due to function approximation. In other words, if Q and C both are similarly difficult to approximate, then we can expect $\texttt{snr}_c$ and $\texttt{snr}_Q$ to have similar values. In AIL, the adversary reward is dense/shaped which is why $\texttt{snr}_c$ is likely to be greater or at least similar to $\texttt{snr}_Q$.

\textbf{When is the decomposition likely to help in AIL?}

As long as $\texttt{snr}_c$ is similar to higher than $\texttt{snr}_Q$ and the gradients of the reward and $C$ are not highly negatively correlated (in expectation), the decomposition is likely to help.

In Fig. \ref{fig:gradient_snr} show how this looks visually for the special case where $\texttt{snr}_c = \texttt{snr}_c$ and that signal strength of $\nabla_a r(s,a)$ is equal to the signal strength of $\nabla_a C(s,a)$.

\textbf{When would the decomposition hurt?}
Two factors that can hurt ARC are:
\begin{enumerate}
    \item If $\texttt{snr}_c$ is significantly lower than $\texttt{snr}_Q$
    \item If $S_{r,c}$ is highly negative
\end{enumerate}
% Both of these factors are unlikely in AIL since the reward is dense/shaped and not misleading.

 % end blue
% Let the SNR in gradient due to function approximation be $\texttt{snr}$ for each of $\nabla_a \hat{Q}(s,a)$ and $\nabla_a \hat{C}(s,a)$. Using our proposed decomposition, the net signal strength in approximating $\nabla_a Q(s,a)$ would have components from $\nabla_a r(s,a)$ and $\nabla_a \hat{C}(s,a)$. The net signal strength would be $1 + \frac{\snr}{1+\snr}$ (assuming equal weightage of gradient from $r$ and $C$). The net noise strength would be $\frac{1}{1+\snr}$. Hence, the effective SNR using our proposed decomposition would be:

% \begin{align*}
%     SNR_{\nabla_a (\hat r(s,a)+\hat{C}(s,a))} &= \frac{1 + \frac{\snr}{1+\snr}}{\frac{1}{1+\snr}}\\
%     &= 2\snr + 1\\
%     &> \snr
% \end{align*}

% Thus, the effective SNR using our proposed decomposition $\nabla_a (\hatr(s,a)+\hat{C}(s,a))$ is larger than that obtained using the standard Actor Critic approach ($\nabla_a \hat{Q}(s,a)$) for the same $\snr$ due to function approximation. Fig. \ref{fig:gradient_snr} shows how the effective SNR in gradient changes as the noise to signal ratio due to function approximation increases. As expected, for small noise, both approaches result in high SNR (accurate gradient). However, as the noise increases, our proposed decomposition in ARC results in higher SNR.

Appendix \ref{appendix:toy_example_gradient_accuracy} experimentally verifies that the decomposition in ARC produces more accurate gradient than AC using a simple 1D driving environment.

\clearpage
\section{Properties of $C$ function}
\label{appendix:c_properties}
We show some useful properties of the $C$ function. We define the optimal $C$ function, $C^*$ as $C^*(s,a) = \max_\pi C^\pi(s,a)$. There exists a unique optimal $C$ function for any MDP as described in Appendix \ref{appendix:unique_optimal_c} Lemma \ref{lemma:unique_optimal_c}. We can derive the Bellman equation for $C^\pi$ (Appendix \ref{appendix:c_bellman} Lemma \ref{lemma:c_bellman}), similar to the Bellman equations for traditional action value function $Q^\pi$ \cite{sutton2018reinforcement}. Using the recursive Bellman equation, we can define a Bellman backup operation for policy evaluation which converges to the true $C^\pi$ function (Theorem \ref{theorem:c_policy_evaluation}). Using the convergence of policy evaluation, we can arrive at the Policy Iteration algorithm using $C$ function as shown in Algorithm \ref{algo:c_policy_iteration}, which is guaranteed to converge to an optimal policy (Theorem \ref{theorem:c_policy_iteration}), similar to the case of Policy Iteration with Q function or V function \cite{sutton2018reinforcement}. For comparison, the standard Policy Iteration with $Q$ function algorithm is described in Appendix \ref{appendix:q_policy_iteration} Algorithm \ref{algo:q_policy_iteration}.

% ------------------------------------------------ %
\subsection{Unique optimality of $C$ function}
\label{appendix:unique_optimal_c}
\begin{lemma}
There exists a unique optimum $C^*$ for any MDP.
\label{lemma:unique_optimal_c}
\begin{proof}
 The unique optimality of C function can be derived from the optimality ofthe $Q$ function \cite{sutton2018reinforcement}. The optimum $Q$ function, $Q^*$ is defined as:
\begin{align}
    Q^*(s,a) &= \max_\pi Q^\pi(s,a) \nonumber\\
             &= \max_\pi [r(s,a) + C^\pi(s,a)] \nonumber\\
             &= r(s,a) + \max_\pi C^\pi(s,a) \nonumber\\
             &= r(s,a) + C^*(s,a) \label{eq:qstar_cstar_relation}\\
    \therefore C^*(s,a) &= Q^*(s,a) - r(s,a) \label{eq:c_star}
\end{align}
% Here, $r(s,a) = \sum_{s'} P(s'|s,a) r(s,a,s')$ is the expected value of reward by taking action $a$ in state $s$ and is a property of the MDP.
Since $Q^*$ is unique \cite{sutton2018reinforcement}, \eqref{eq:c_star} implies $C^*$ must be unique.
\end{proof}
\end{lemma}

% ------------------------------------------------ %
\clearpage
\subsection{Bellman backup for $C$ function}
\label{appendix:c_bellman}
\begin{lemma}
\label{lemma:c_bellman}
The recursive Bellman equation for $C^\pi$ is as follows
\begin{align*}
    C^\pi(s,a) &= \gamma \sum_{s'} P(s'|s,a) \sum_{a'} \pi(a'|s') \left(r(s',a') + C^\pi(s',a') \right)  
\end{align*}
\begin{proof}
The derivation is similar to that of state value function $V^\pi$ presented in \cite{sutton2018reinforcement}. We start deriving the Bellman backup equation for $C^\pi$ function by expressing current $C(s_t, a_t)$ in terms of future $C(s_{t+1}, a_{t+1})$. In the following, the expectation is over the policy $\pi$ and the transition dynamics $\mathcal{P}$ and is omitted for ease of notation.

\begin{align}
    C^\pi(s_t,a_t) &= \Ebb \sum_{k\geq1} \gamma^{k} r_{t+k}\\
                   &= \Ebb \left(\gamma r_{t+1} + \sum_{k\geq2} \gamma^{k} r_{t+k} \right)\\
                   &= \gamma \left(\Ebb [r_{t+1}]
                   +\Ebb \sum_{k\geq1} \gamma^{k} r_{t+1} \right)\\
                   &= \gamma \left(\Ebb [r_{t+1}]
                   +\Ebb \sum_{k\geq1} \gamma^{k} r_{t+1+k} \right)\\
                   &= \gamma \Ebb \left(r_{t+1}
                   + C(s_{t+1}, a_{t+1}) \right) \label{eq:c_cnext}\\
\end{align}
Using Eq. \ref{eq:c_cnext}, we can write the recursive Bellman equation of C.
\begin{align}
    C(s,a) &= \gamma \sum_{s'} P(s'|s,a) \sum_{a'} \pi(a'|s')\left(r(s',a')+C(s',a') \right)  \label{eq:c_bellman}
\end{align}
\end{proof}
\end{lemma}

% ------------------------------------------------ %
\clearpage
\subsection{Convergence of policy evaluation using C function}
\label{appendix:c_policy_evaluation}
\begin{theorem}
\label{theorem:c_policy_evaluation}
The following Bellman backup operation for policy evaluation using $C$ function converges to the true C function, $C^\pi$ $$C^{n+1}(s,a) \leftarrow \gamma \sum_{s'} P(s'|s,a) \sum_{a'} \pi(a'|s')\left(r(s',a')+C^{n}(s',a')\right)$$
Here, $C^n$ is the estimated value of $C$ at iteration n.
\begin{proof}
Let us define $F_{(.)}$ as the Bellman backup operation over the current estimates of $C$ values:
\begin{align}
    F_C(s,a) &= \gamma \sum_{s'} P(s'|s,a) \sum_{a'} \pi(a'|s')\left(r(s',a')+C(s',a')\right)
\end{align}
We prove that $F$ is a contraction mapping w.r.t $\infty$ norm and hence is a fixed point iteration. Let, $C_1$ and $C_2$ be any 2 sets of estimated $C$ values.
\begin{align}
\infnorm{F_{C_1} - F_{C_2}} &= \max_{s,a} \abs{F_{C_1} - F_{C_2}}\\
                            &= \gamma \max_{s,a} \left|
                            \sum_{s'} P(s'|s,a) \sum_{a'} \pi(a'|s')
                            [r(s',a')+C_1(s',a')] \right.\nonumber\\
                            & \left. - \sum_{s'} P(s'|s,a) \sum_{a'} \pi(a'|s')[r(s',a')+C_2(s',a')]\right|\\
                            &= \gamma  \abs{\max_{s,a}  \sum_{s'} P(s'|s,a) \sum_{a'} \pi(a'|s') (C_1(s',a')-C_2(s',a'))}\\
                            &\leq \gamma \max_{s,a}  \sum_{s'} P(s'|s,a) \sum_{a'} \pi(a'|s') \abs{C_1(s',a')-C_2(s',a'))}\\
                            &\leq \gamma \max_{s,a}  \sum_{s'} P(s'|s,a) \max_{a'} \abs{C_1(s',a')-C_2(s',a'))}\\
                            &\leq \gamma \max_{s'} \max_{a'} \abs{C_1(s',a')-C_2(s',a'))}\\
                            &= \gamma \max_{s,a} \abs{C_1(s,a)-C_2(s,a))}\\
                            &= \gamma \infnorm{C_1-C_2}\\
\therefore \infnorm{F_{C_1} - F_{C_2}} &\leq \gamma \infnorm{C_1-C_2} \label{eq:c_eval_contraction}
\end{align}

Eq. \ref{eq:c_eval_contraction} implies that iterative operation of $F_{(.)}$ converges to a fixed point. The true $C^\pi$ function satisfies the Bellman equation Eq. \ref{eq:c_bellman}. These two properties imply the policy evaluation converges to the true $C^\pi$ function.
\end{proof}
\end{theorem}

% ------------------------------------------------ %
\subsection{Convergence of policy iteration using C function}
\label{appendix:c_policy_iteration}
\setcounter{theorem}{0}
\begin{theorem}
\label{theorem:c_policy_iteration}
The policy iteration algorithm defined by Algorithm \ref{algo:c_policy_iteration} converges to the optimal $C^*$ function and an optimal policy $\pi^*$.
% \vspace{-8mm}
\begin{proof}
From Theorem \ref{theorem:c_policy_evaluation}, the policy evaluation step converges to true $C^\pi$ function. The policy improvement step is exactly the same as in the case with Q function since $Q^\pi(s,a) = r(s,a) + C^\pi(s,a)$, which is known to converge to an optimum policy \cite{sutton2018reinforcement}. These directly imply that Policy Iteration with $C$ function converges to the optimal $C^*$ function and an optimal policy $\pi^*$.
\end{proof}
\end{theorem}

\section{Popular Algorithms}
\subsection{Policy Iteration using Q function}
\label{appendix:q_policy_iteration}

We restate the popular Policy Iteration using $Q$ function algorithm in Algorithm \ref{algo:q_policy_iteration}.
\begin{algorithm}[ht]
\SetAlgoLined
 Initialize $Q^0(s,a) \forall s,a$\;
 \While{$\pi$ not converged}{
 // Policy evaluation\\
  \For{n=1,2,\dots until $Q^n$ converges}
  {
      $Q^{n+1}(s,a) \leftarrow r(s,a) + \gamma \sum_{s'} P(s'|s,a) \sum_{a'} \pi(a'|s') Q^n(s',a') \quad \forall s,a $
  }
  // Policy improvement\\
  $\pi(s,a) \leftarrow$
  $\begin{cases}
  1, \text{ if } a = \argmax_{a'} Q(s,a')\\
  0, \text{ otherwise}
  \end{cases}
  \forall s,a$\\
 }
 \caption{Policy Iteration using Q function}
 \label{algo:q_policy_iteration}
\end{algorithm}

\section{Additional Results}
\label{appendix:additional_results}

\subsection{Visualization of Real Robot Policy Execution}
\begin{figure}[h!]
    \centering
    \begin{subfigure}[b]{0.24\textwidth}
         \centering
         \includegraphics[width=\textwidth, height=0.75\textwidth]{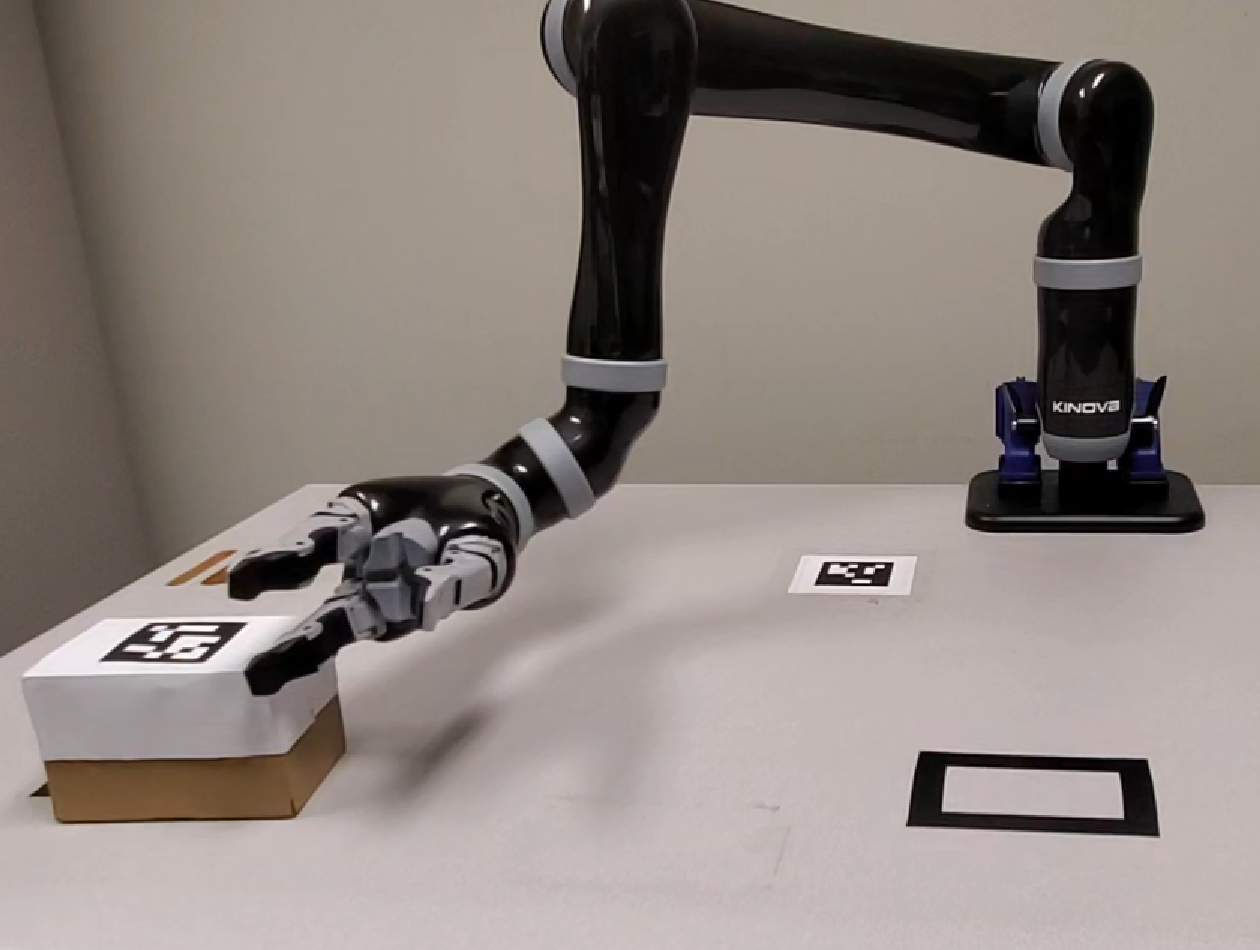}
         \caption{ARC-$f$-Max-RKL\\ (Our) 1.94 $\pm$ 0.27 cm}
     \end{subfigure}
     \begin{subfigure}[b]{0.24\textwidth}
         \centering
         \includegraphics[width=\textwidth, height=0.75\textwidth]{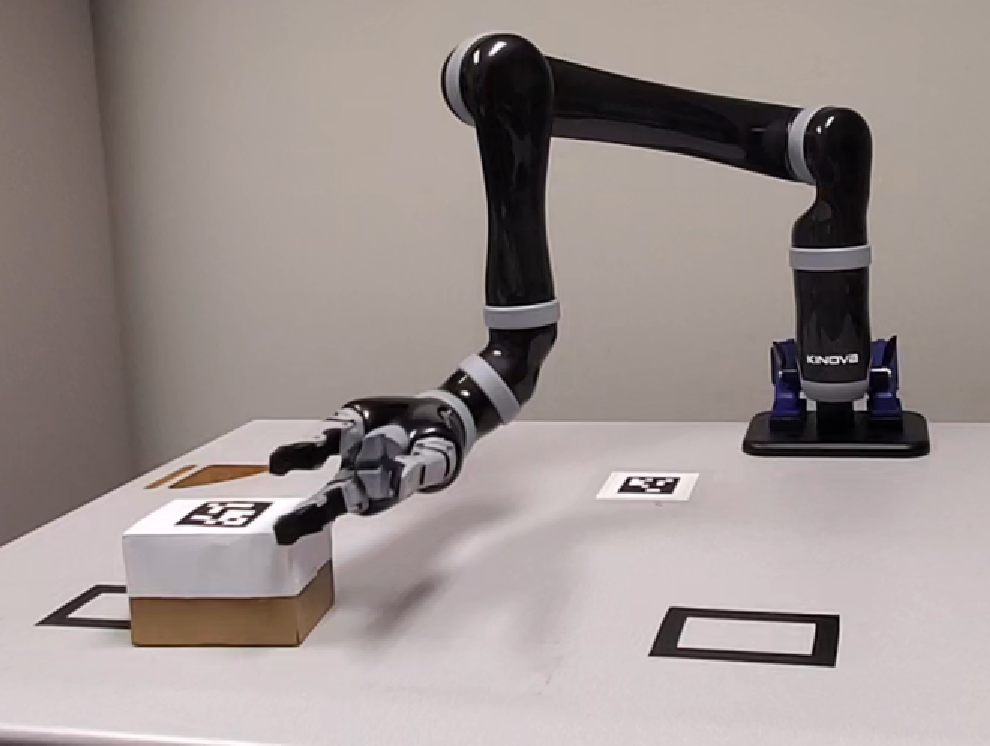}
         \caption{ARC-GAIL\\ (Our) 5.95 $\pm$ 0.99 cm}
     \end{subfigure}
     \begin{subfigure}[b]{0.24\textwidth}
         \centering
         \includegraphics[width=\textwidth, height=0.75\textwidth]{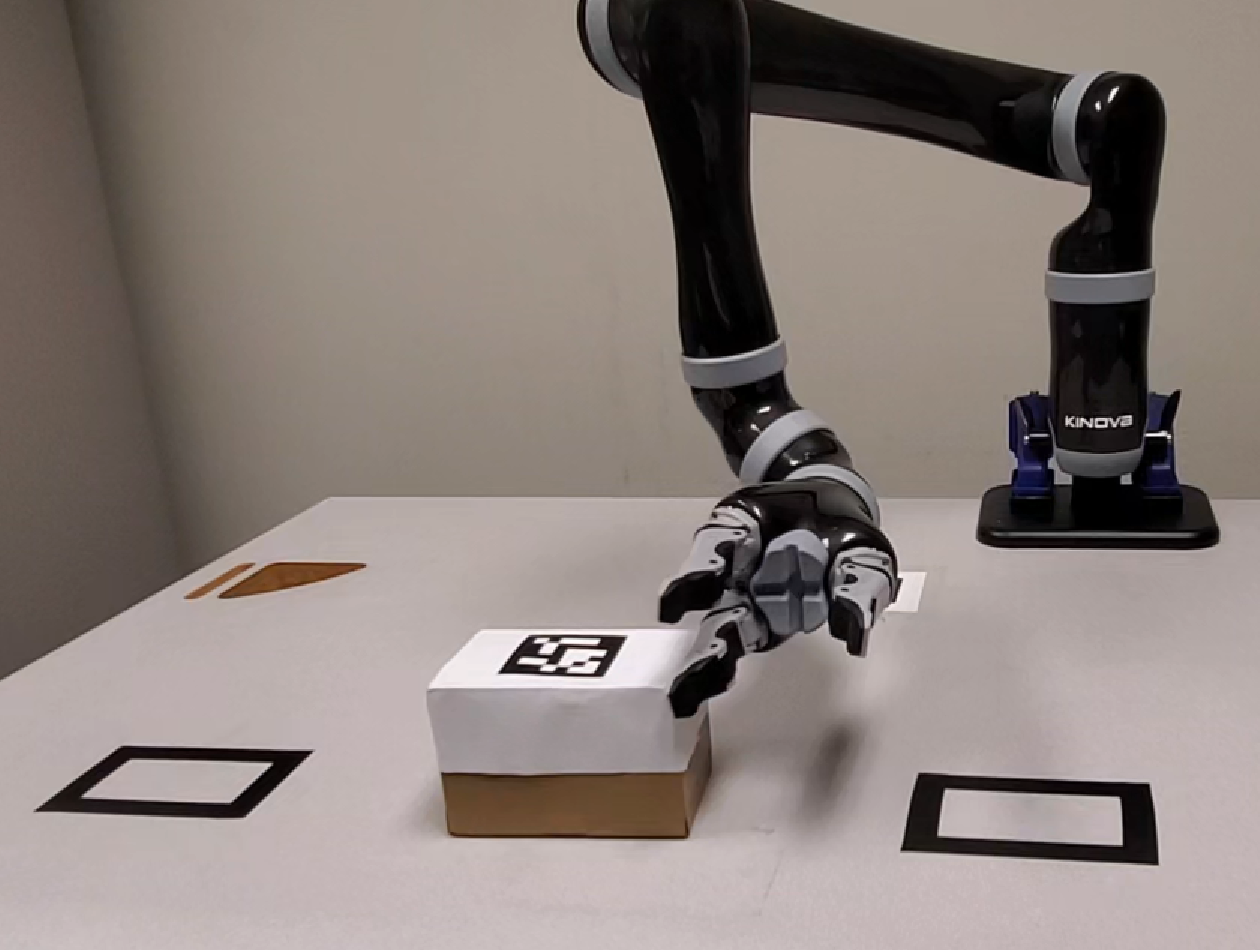}
         \caption{$f$-Max-RKL\\ 11.48 $\pm$ 0.84 cm}
     \end{subfigure}
     \begin{subfigure}[b]{0.24\textwidth}
         \centering
         \includegraphics[width=\textwidth, height=0.75\textwidth]{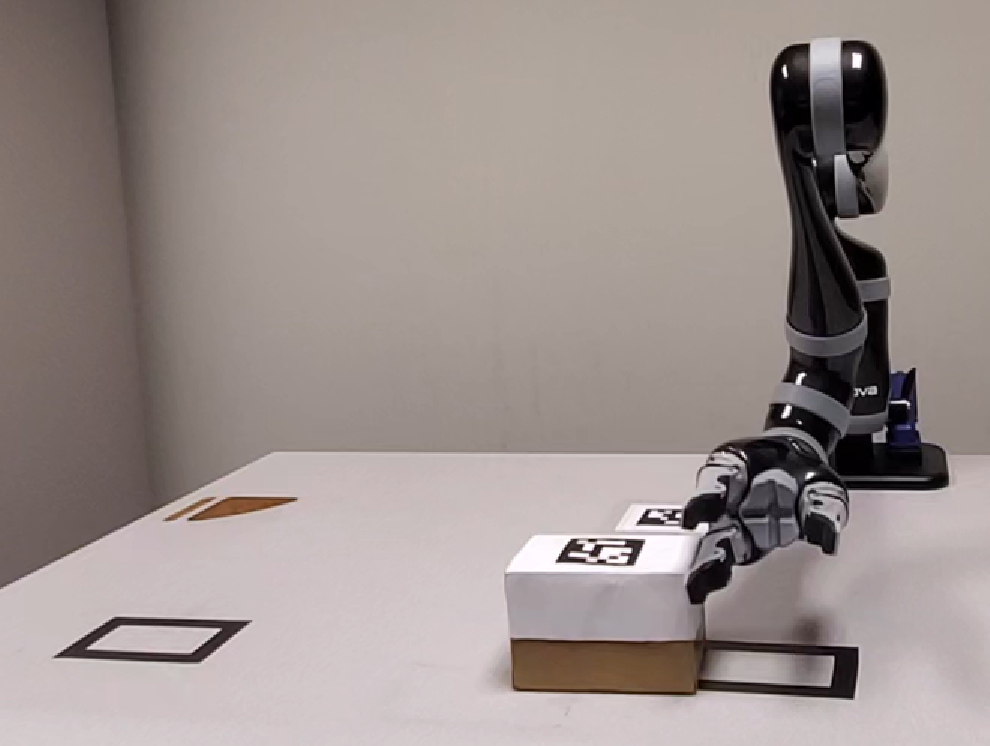}
         \caption{GAIL\\ 17.48 $\pm$ 1.03 cm}
     \end{subfigure}
    \caption{Example snapshots of final position of block in the JacoPush task using different Adversarial Imitation Learning algorithms and the average final block to goal distance (lower is better) in each case.}
    \label{fig:jaco_final_position}
\end{figure}

Fig. \ref{fig:jaco_final_position} shows example snapshots of the final block position in the JacoPush task using different AIL algorithms. ARC aided AIL algorithms were able to push the block closer the goal thereby achieving a lower final block to goal distance, as compared to the standard AIL algorithms.
\captionsetup[subfigure]{justification=centering}

\subsection{Accuracy of gradient of the proposed approach}
\label{appendix:toy_example_gradient_accuracy}
\begin{figure}[h!]
    \centering
    \includegraphics[width=0.6\textwidth]{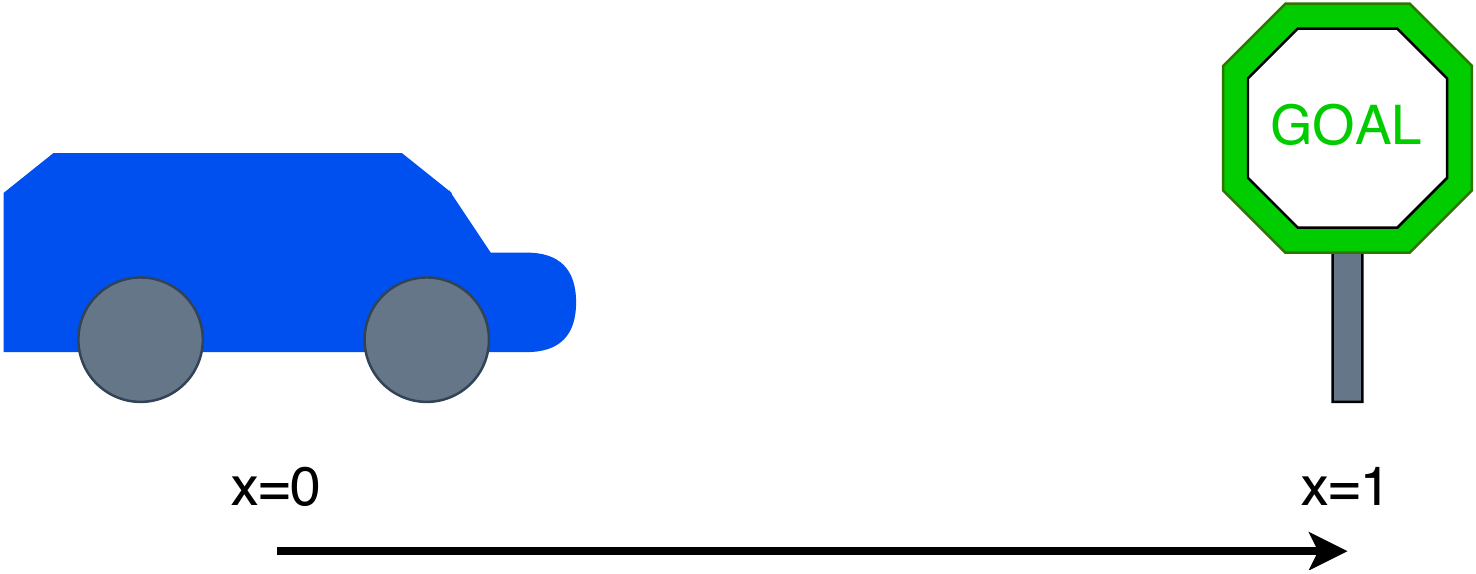}
    \caption{A 1D driving environment where a policy needs to imitate an expert driver which initially drives fast but slows down as the car approaches the goal.}
    \label{fig:car_environment}
\end{figure}

We use a simple toy environment to empirically show that our proposed approach of fitting $C$ function results in a better estimate of the policy gradient that the standard approach of fitting $Q$ function. Fig. \ref{fig:car_environment} shows the environment. An agent needs to imitate an expert policy that drives the car from the start location $x=0$ to the goal $x=1$. The expert policy initially drives the car fast but slows down as the car approaches the goal.

The expert policy, agent policy and reward functions are described by the following python functions:

\begin{lstlisting}[language=Python, caption=Python code defining the policies (expert and agent) and reward function.]
def expert_policy(obs):
    gain = 0.1
    goal = 1
    return gain*(goal-obs)+0.1
        
def agent_policy(obs):
    return 0.15

def reward_fn(obs, a):
    expert_a = expert_policy(obs)
    return -100*(a-expert_a)**2
\end{lstlisting}

We uniformly sampled states and actions in this environment. We then fit a neural network to the Q function and to the C function by running updates for different numbers of epochs and repeating the experiment 5 times.

After that, we compare the learnt Q function and (r + learnt C) function to the true Q function. (The true Q function is obtained by rolling out trajectories in the environment).

The following 2 figures show the results. On the left, we show the error in estimating the true Q function and on the right we show the error in estimating the true gradient of Q. (The True gradient of Q is calculated by a finite difference method).

\begin{figure}[h!]
    \centering
    \begin{subfigure}[b]{0.48\textwidth}
         \centering
         \includegraphics[width=\textwidth]{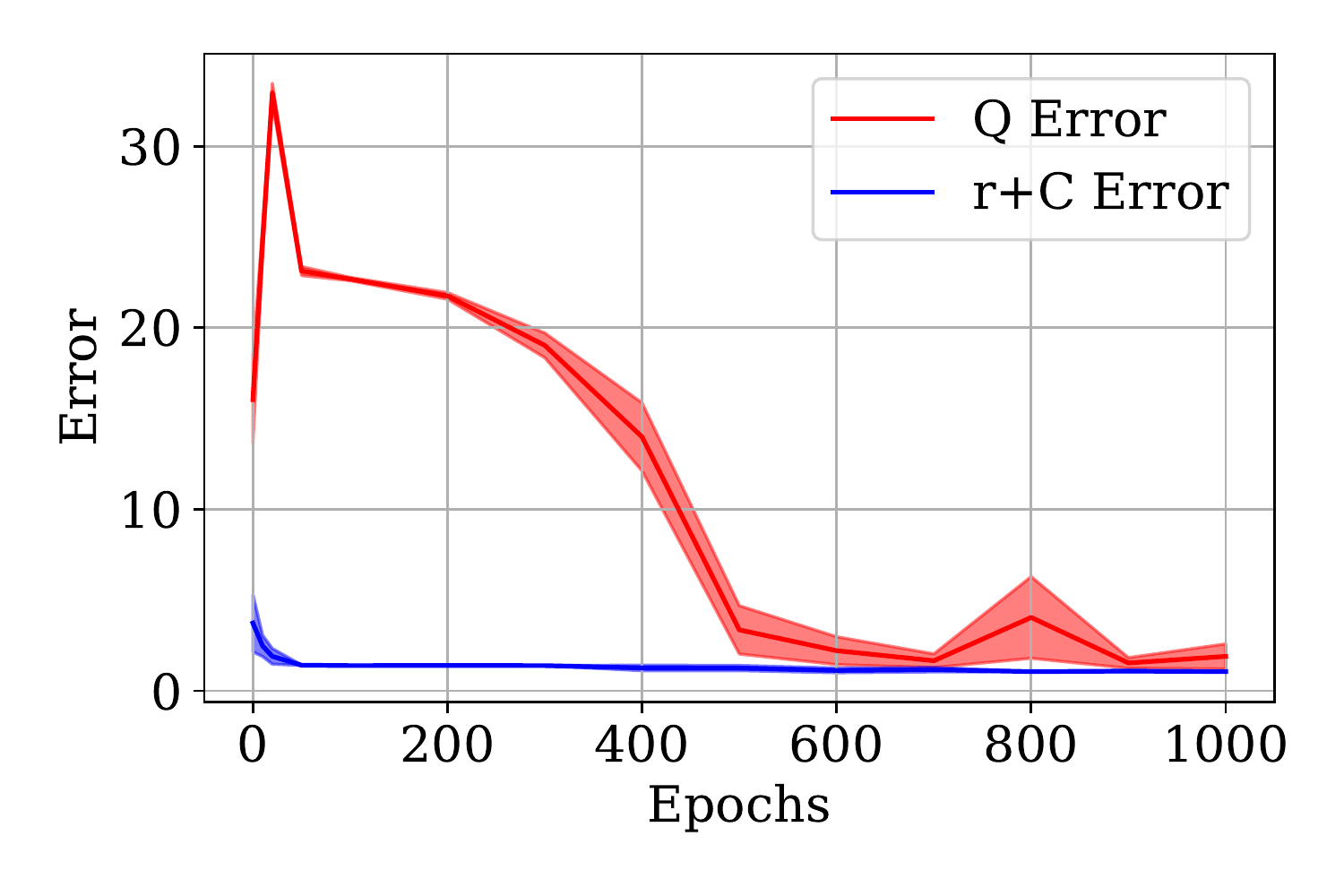}
         \vspace{-5mm}
         \caption{Error in estimating $Q$}
         \label{fig:q_eror}
     \end{subfigure}
     \begin{subfigure}[b]{0.48\textwidth}
         \centering
         \includegraphics[width=\textwidth]{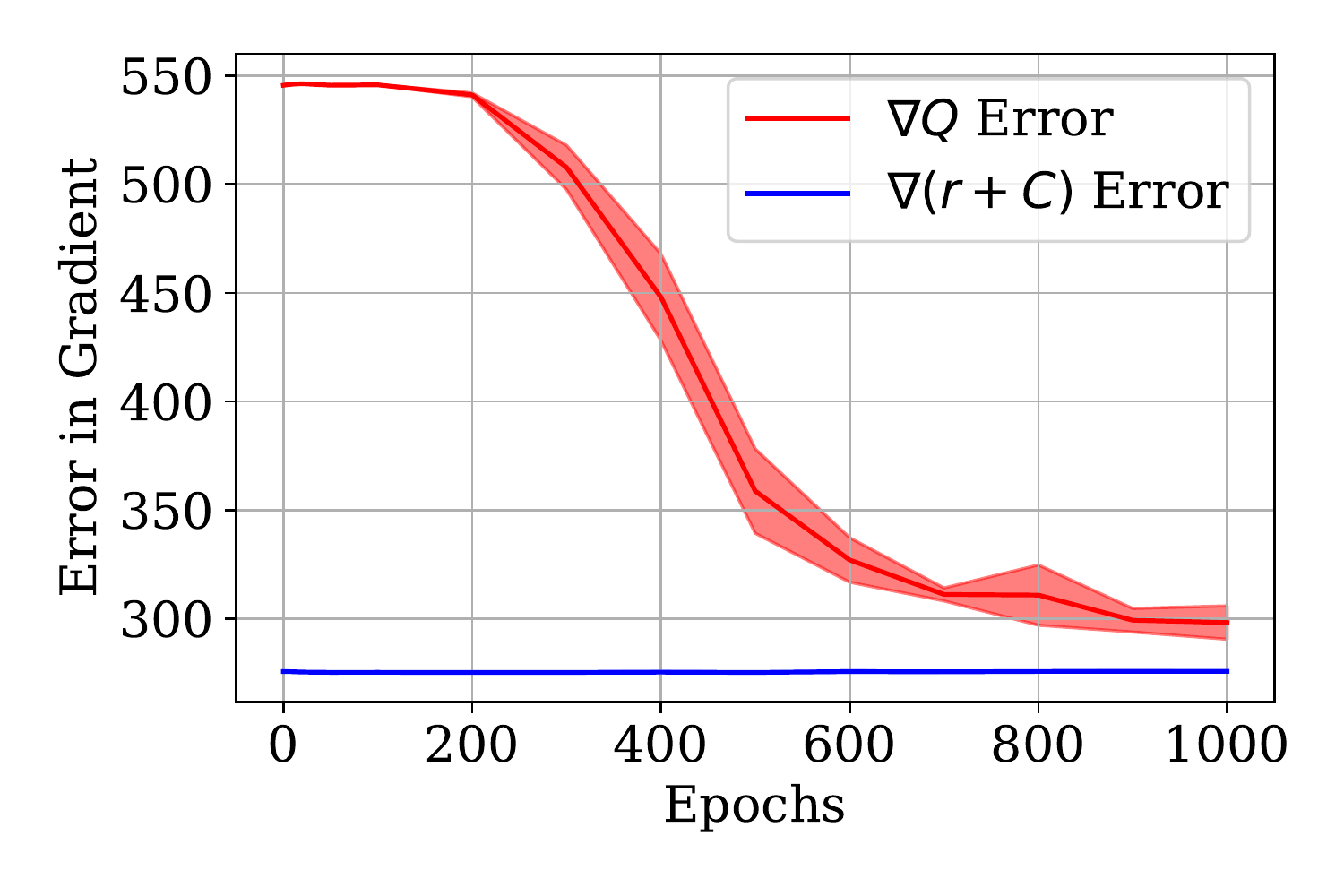}
         \vspace{-5mm}
         \caption{Error in estimating $\nabla_a Q(s,a)$}
         \label{fig:grad_q_error}
     \end{subfigure}
\end{figure}

Clearly, the decomposition leads to lower error and variance in estimating both the true Q function and its gradient. Even with slight error with r+C initially, the corresponding error in gradient is much lower for r+C than for Q. Moreover, towards the tail of the plots (after 600 epochs), both Q and r+C estimate the true Q function quite accurately but the error in the gradient of r+C is lower than that for directly estimating the Q function. 

\begin{figure}[ht]
    \centering
    \begin{subfigure}[b]{0.32\textwidth}
         \centering
         \includegraphics[width=\textwidth]{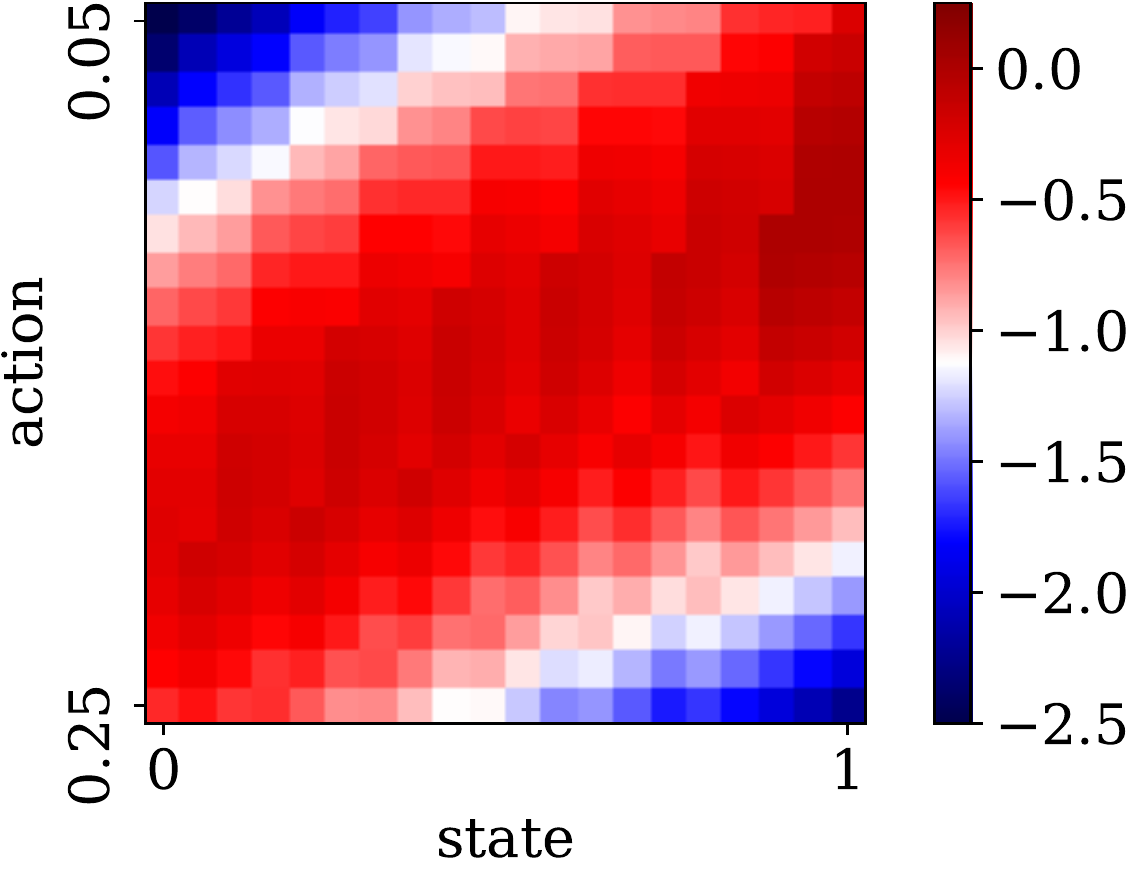}
         \vspace{-5mm}
         \caption{True $Q$}
         \label{fig:toy_true_q}
     \end{subfigure}
     \begin{subfigure}[b]{0.32\textwidth}
         \centering
         \includegraphics[width=\textwidth]{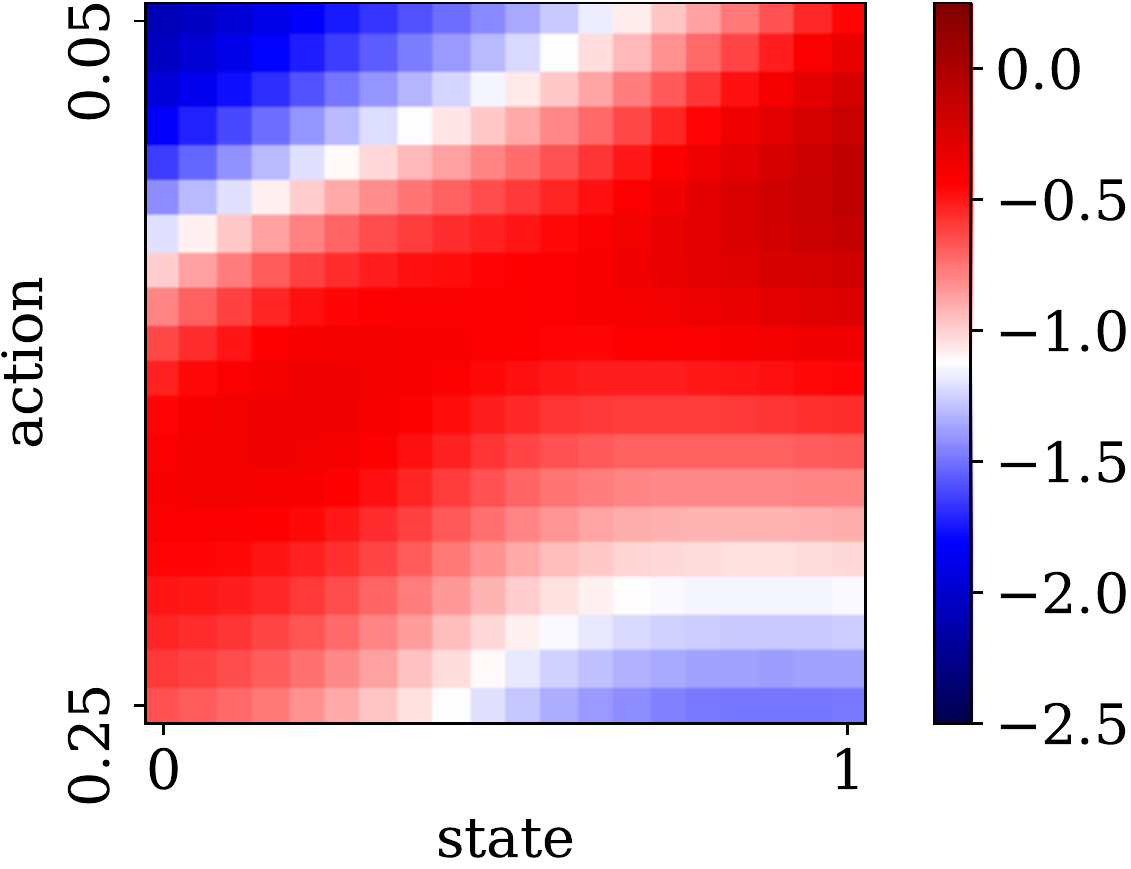}
         \vspace{-5mm}
         \caption{$\hat{Q}$}
         \label{fig:toy_q}
     \end{subfigure}
     \begin{subfigure}[b]{0.32\textwidth}
         \centering
         \includegraphics[width=\textwidth]{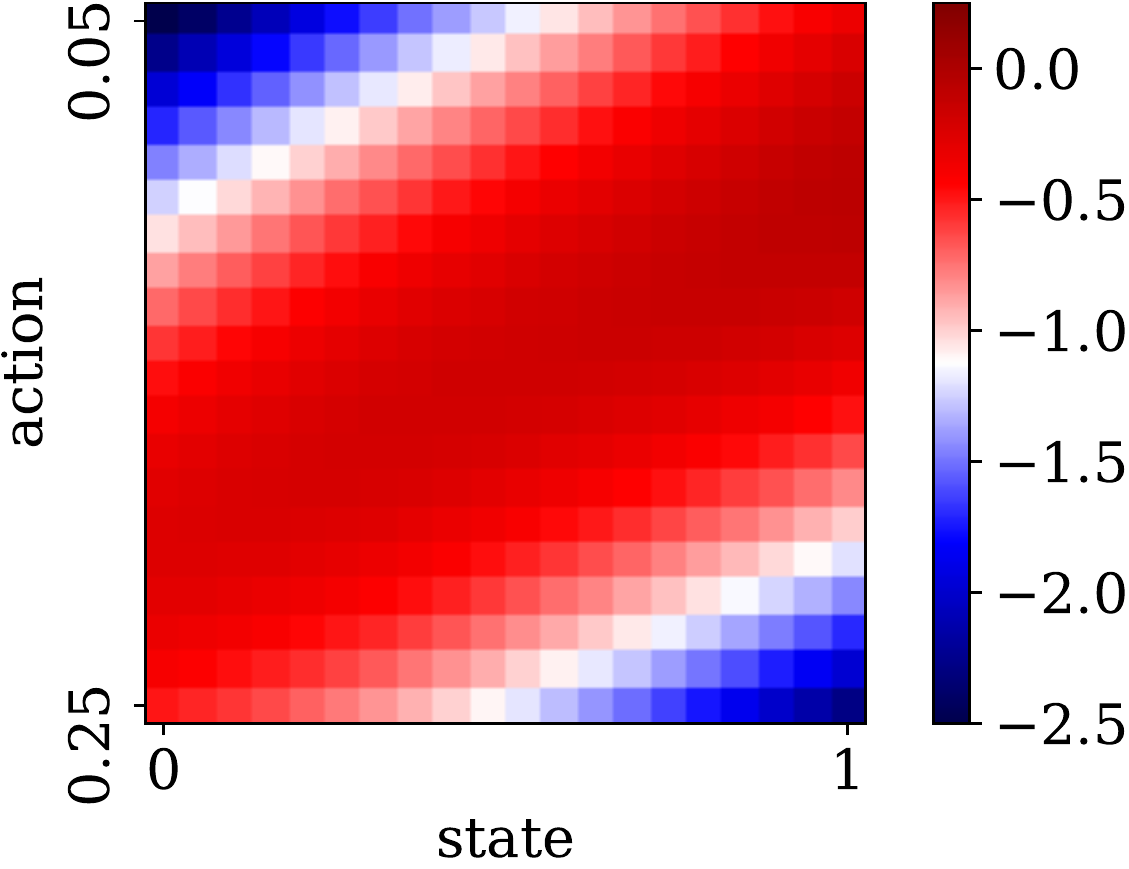}
         \vspace{-5mm}
         \caption{$r+\hat{C}$}
         \label{fig:toy_r_plus_c}
     \end{subfigure}
     \vspace{-2mm}
    \caption{True value of $Q$, Fig. \ref{fig:toy_true_q} along with estimated values of $Q$ by directly fitting a $Q$ network, Fig. \ref{fig:toy_q} and by fitting a $C$ network, Fig. \ref{fig:toy_r_plus_c}.}
    \label{fig:toy_q_estimates}
\end{figure}

\begin{figure}[ht]
    \centering
    \begin{subfigure}[b]{0.32\textwidth}
         \centering
         \includegraphics[width=\textwidth]{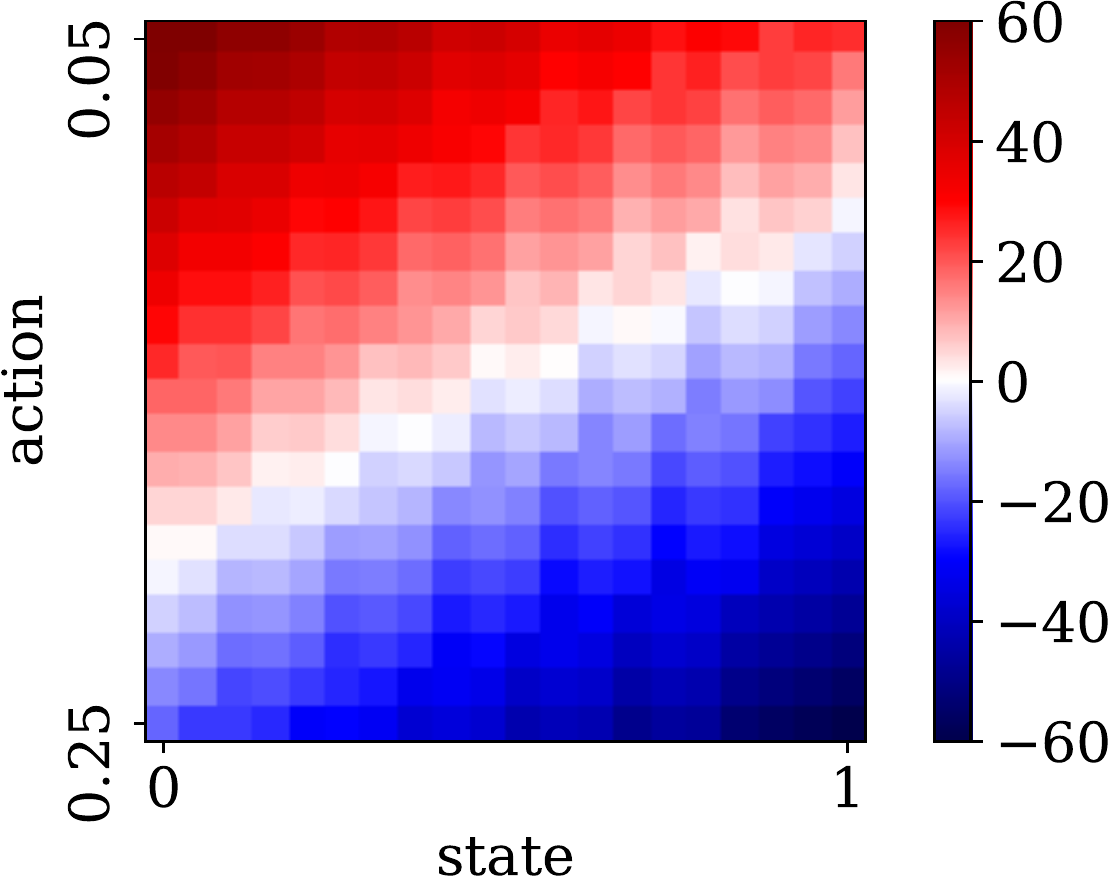}
         \vspace{-5mm}
         \caption{True $\nabla_a{Q(s,a)}$}
         \label{fig:toy_true_q}
     \end{subfigure}
     \begin{subfigure}[b]{0.32\textwidth}
         \centering
         \includegraphics[width=\textwidth]{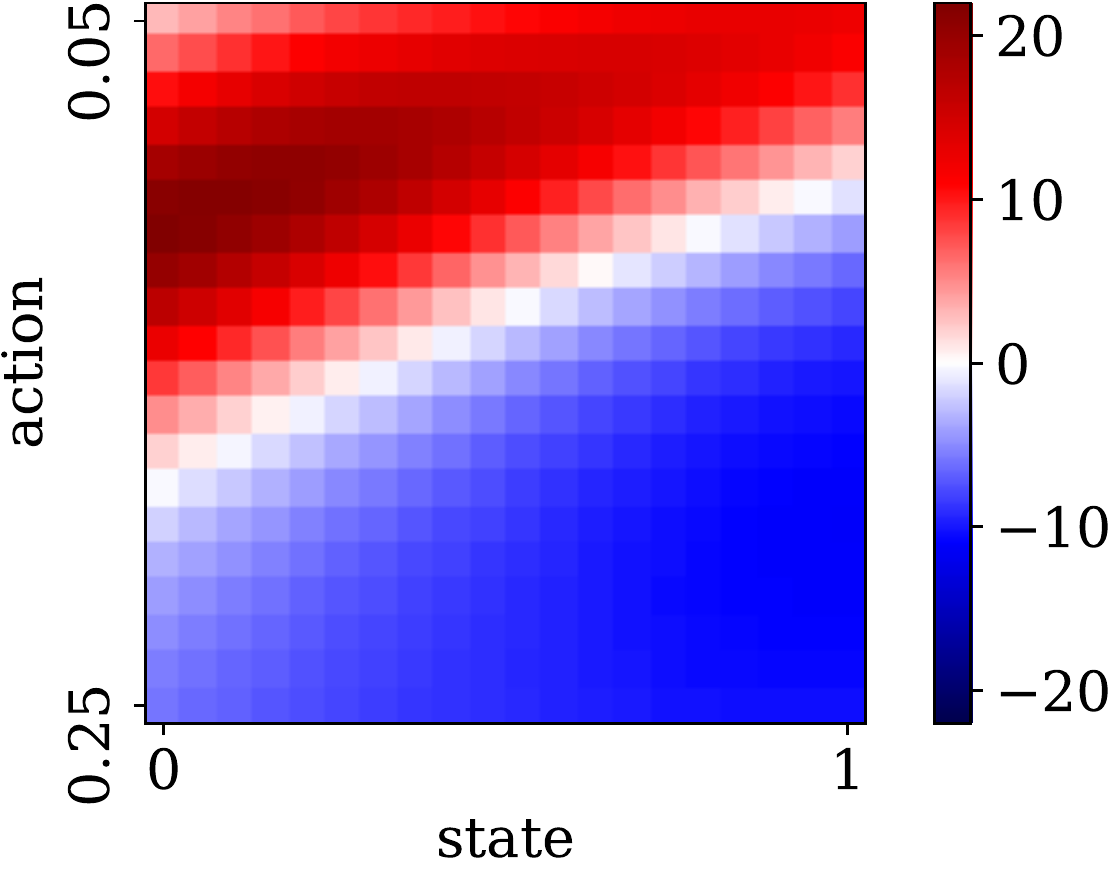}
         \vspace{-5mm}
         \caption{$\nabla_a{\hat{Q}(s,a)}$}
         \label{fig:toy_q}
     \end{subfigure}
     \begin{subfigure}[b]{0.32\textwidth}
         \centering
         \includegraphics[width=\textwidth]{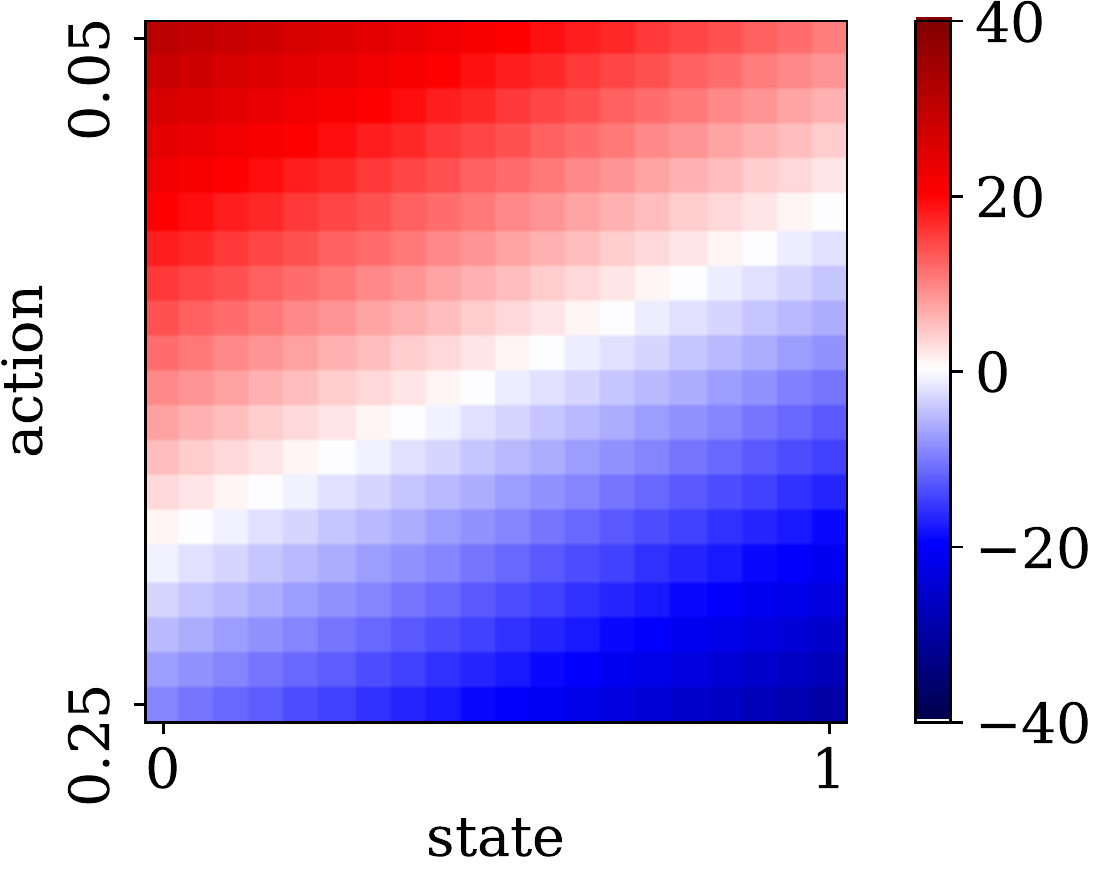}
         \vspace{-5mm}
         \caption{$\nabla_a(r(s,a)+\hat{C}(s,a))$}
         \label{fig:toy_r_plus_c}
     \end{subfigure}
     \vspace{-2mm}
    \caption{True value of $\nabla_a{Q(s,a)}$, Fig. \ref{fig:toy_true_q} along with estimated values of $\nabla_a{Q(s,a)}$ function by directly fitting a $Q$ network, Fig. \ref{fig:toy_q} and by fitting a $C$ network, Fig. \ref{fig:toy_r_plus_c}.}
    \label{fig:toy_grad_q_estimates}
\end{figure}

We visualize the estimated values of Q in Fig. \ref{fig:toy_q_estimates} and the estimated gradients of Q in Fig. \ref{fig:toy_grad_q_estimates} by the two methods after 500 epochs of training. Using r+C estimates the gradients much better than Q.

\clearpage
\section{Experimental Details}
\label{appendix:experimental_details}

\subsection{Policy iteration on a Grid World}
\label{appendix:grid_world}
Our objective is to experimentally validate if Policy Iteration (PI) with $C$ function converges to an optimal policy (Theorem \ref{theorem:c_policy_iteration}). We choose a simple Grid World environment as shown in Fig. \ref{fig:policy_iteration_grid} to illustrate this. At every time step, the agent can move in one of 4 directions - left, right, up or down. The reward is $1$ for reaching the goal (G) and $0$ otherwise. The discount factor $\gamma=0.9$.

On this environment, we run two PI algorithms - PI with $C$ function (Algorithm \ref{algo:c_policy_iteration}) and the standard PI with d function (Appendix \ref{appendix:q_policy_iteration} Algorithm \ref{algo:q_policy_iteration}).
Fig. \ref{fig:policy_iteration_grid} shows the results of this experiment. Both the algorithms converge to the same optimal policy $\pi^*$ shown in Fig. \ref{fig:policy_iteration_grid_policy}. This optimal policy receives the immediate reward shown in Fig. \ref{fig:policy_iteration_grid_r}. Note that the immediate reward is $1$ for states adjacent to the goal G as the agent receives $1$ reward for taking an action that takes it to the goal. 
Fig. \ref{fig:policy_iteration_grid_c} and Fig. \ref{fig:policy_iteration_grid_q} show the values of $C^*$, $Q^*$ that PI with $C$ function and PI with $Q$ function respectively converge to. In Fig. \ref{fig:policy_iteration_grid_q}, $Q^*=r^*+C^*$, which is consistent with the relation between $Q$ function and $C$ function \eqref{eq:q_c_relation}. In Fig. \ref{fig:policy_iteration_grid_q}, the $Q^*$ values in the states adjacent to the goal are $1$ since $Q$ function includes the immediate reward \eqref{eq:q_definition_repeat}. $C$ function doesn't include the immediate reward \eqref{eq:c_definition} and hence the $C^*$ values in these states are $0$ (Fig. \ref{fig:policy_iteration_grid_c}). This experiment validates that PI with $C$ function converges to an optimal policy as already proved in Theorem \ref{theorem:c_policy_iteration}.

\subsection{Imitation Learning in Mujoco continuous-control tasks}
\label{section:mujoco_appendix}
\paragraph{Environment}
We use Ant-v2, Walker-v2, HalfCheetah-v2 and Hopper-v2 Mujoco continuous-control environments from OpenAI Gym \cite{brockman2016openai}. All $4$ environments use Mujoco, a realistic physics-engine, to model the environment dynamics. The maximum time steps, $T$ is set to $1000$ in each environment.

\paragraph{Code}
We implemented our algorithm on top of the AIL code of \cite{ni2021f}. The pre-implemented standard AIL algorithms ($f$-MAX-RKL, GAIL) used SAC \cite{haarnoja2018soft} as the RL algorithm and the ARC aided AIL algorithms are SARC-AIL (Algorithm \ref{algo:sarc_ail}) algorithms.

\paragraph{Expert trajectories}
We used the expert trajectories provided by \cite{ni2021f}. They used SAC \cite{haarnoja2018soft} to train an expert in each environments. The policy network, $\pi_\theta$ was a tanh squashed Gaussian which parameterized the mean and standard deviation with two output heads. Each of the policy network, $\pi_\theta$ and the 2 critic networks, $Q_{\phi_1},Q_{\phi_1}$ was a $(64,64)$ ReLU MLP. Each of them was optimized by Adam optimizer with a learning rate of $0.003$. The entropy regularization coefficient, $\alpha$ was set to $1$, the batch size was set to $256$, the discount factor $\gamma$ was set to $0.99$ and the polyak averaging coefficient $\zeta$ for target networks was set to $0.995$. The expert was trained for 1 million time steps on Hopper and 3 million time steps on the other environments. For each environment, we used 1 trajectory from the expert stochastic policies to train the imitation learning algorithms.

\paragraph{Standard AIL}
For the standard AIL algorithms ($f$-MAX-RKL \cite{ghasemipour2020divergence} and GAIL \cite{ho2016generative}) we used the code provide by \cite{ni2021f}. The standard AIL algorithms used SAC \cite{haarnoja2018soft} as the RL algorithm. SAC used the same network and hyper-parameters that were used for training the expert policy except the learning rate and the entropy regularization coefficient, $\alpha$. The learning rate was set to 0.001. $\alpha$ was set to 0.05 for HalfCheetah and to 0.2 for the other environments. The reward scale scale and gradient penalty coefficient were set to 0.2 and 4.0 respectively. In each environment, the observations were normalized in each dimension of the state using the mean and standard deviation of the expert trajectory.

Baseline GAIL in Hopper was slightly unstable and we had to tune GAIL separately for the Hopper environment. We adjusted the policy optimizer's learning rate schedule to decay by a factor of 0.98 at every SAC update step after 5 epochs of GAIL training.

For the discriminator, we used the same network architecture and hyper-parameters suggested by \cite{ni2021f}. The discriminator was a (128,128) tanh MLP network with the output clipped within [-10,10]. The discriminator was optimized with Adam optimizer with a learning rate of 0.0003 and a batch size of 128. Once every 1000 environment steps, the discriminator and the policy were alternately trained for 100 iterations each.

Each AIL algorithm was trained for 1 million environment steps on Hopper, 3 million environment steps on Ant, HalfCheetah and 5 million environment steps on Walker2d.

\paragraph{ARC aided AIL}
For ARC aided AIL algorithms, we modified the SAC implementation of \cite{ni2021f} to SARC - Soft Actor Residual Critic. This was relatively straight forward, we used the same networks to parameterize $C_{\phi_1}, C_{\phi_2}$ instead of $Q_{\phi_1}, Q_{\phi_2}$ based on the steps of SARC-AIL (Algorithm \ref{algo:sarc_ail}). For SARC, we used the same network and hyper-parameters as SAC except the following changes. Learning rate was set to 0.0001. Entropy regularization coefficient, $\alpha$ was set to 0.05 for HalfCheetah and 1 for the other environments. No reward scaling was used (reward scale was set to 1). The C networks were updated 10 times for every update of the policy network. We did so because we noticed that otherwise the C networks ($C_{\phi_1}, C_{\phi_2}$) were slower to update as compared to the policy network.

The discriminator was the same as with standard AIL algorithms except it had 2 Resnet blocks of 128 dimension each, with batch normalization and leaky ReLU activation. These changes were motivated by common tricks to train stable GANs \cite{goodfellowgenerative}. In GANs, the generator is differentiated through the discriminator and the use of leaky ReLU and Resnet helps in gradient flow through the discriminator. In ARC aided AIL we have a similar scenario, the policy is differentiated through the reward function. We briefly tried to make the same changes with standard AIL algorithms as well but didn't see an improvement in performance.

\paragraph{Naive-Diff}
For the Naive-Diff aided AIL algorithms (Naive-Diff-$f$-MAX-RKL and Naive-Diff-GAIL), we used the same network architectures and hyper-parameters as with ARC aided AIL.

\paragraph{Behavior Cloning}
For Behavior Cloning, we trained the agent to regress on expert actions by minimizing the mean squared error for 10000 epochs using Adam optimizer with learning rate of 0.001 and batch size of 256.

\paragraph{Evaluation}
We evaluated all the imitation learning algorithms based on the true environment return achieved by the deterministic version of their policies. Each algorithm was run on 5 different seeds and each run was evaluated for 20 episodes. The final mean reward was used for comparing the algorithms. The results are presented in Table \ref{tab:mujoco_performance}. 

% ----------------------------------------------- %
\subsection{Imitation Learning in robotic manipulation tasks}
\label{appendix:experimental_details_robotic_sim}
\paragraph{Environment}
We simplified 2D versions of the FetchReach-v1 and FetchPush-v1 environments from OpenAI gym, \cite{brockman2016openai}. In the FetchReach task, the observation is a 2D vector containing the 2D position of the block with respect to the end-effector and needs to take it's end-effector to the goal as quickly as possible. In the FetchPush task, the robot's ob can observe a block and the goal location and needs to push the block to the goal as quickly as possible. Actions are 2D vectors controlling the relative displacement of the end-effector from the current position by a maximum amount of $\pm \Delta_{\max} = 3.3$cm. In the FetchReach task, the goal is initially located at (15cm,-15cm) + $\epsilon$ w.r.t the end-effector. Where $\epsilon$ is sampled from a 2D diagonal Normal distribution with 0 mean and 0.01cm standard deviation in each direction. In the FetchPush task, initially, the block is located at (0cm,-10cm)+$\epsilon_{\text{block}}$ and the goal is located at (0cm,-30cm)+$\epsilon_{\text{goal}}$ w.r.t the end-effector. $\epsilon_\text{block}, \epsilon_\text{goal}$ are sampled from 2D diagonal Normal distributions with 0 mean and 0.01cm standard deviation in each direction. The reward at each time step is -$d$, where $d$ is the distance between end-effector and goal (in case of FetchReach) or the distance between the block and the goal (in case of FetchPush). $d$ is expressed in meters. FetchReach task has 20 time steps and FetchPush task has 30 time steps.

\paragraph{Expert trajectories}
We used hand-coded proportional controller to generate expert trajectories for these tasks. For each task, we used 64 expert trajectories.

\paragraph{Hyper-parameters}
For each AIL algorithm, once every 20 environment steps, the discriminator and the policy were alternately trained for 10 iterations each. Each AIL algorithm was trained for 25,000 environment steps. All the other hyper-parameters were the same as those used with the Ant, Walker and Hopper Mujoco continuous-control environments (Section \ref{section:mujoco_appendix}). We didn't perform any hyper-parameter tuning (for both our methods and the baselines) in these experiments and the results might improve with some hyper-parameter tuning.

\paragraph{Evaluation}
For the simulated tasks, each algorithms is run with 5 random seeds and each seed is evaluated for 20 episodes.

\subsection{Sim-to-real transfer of robotic manipulation policies}
\begin{figure}[h!]
    \centering
    \includegraphics[width=0.6\textwidth]{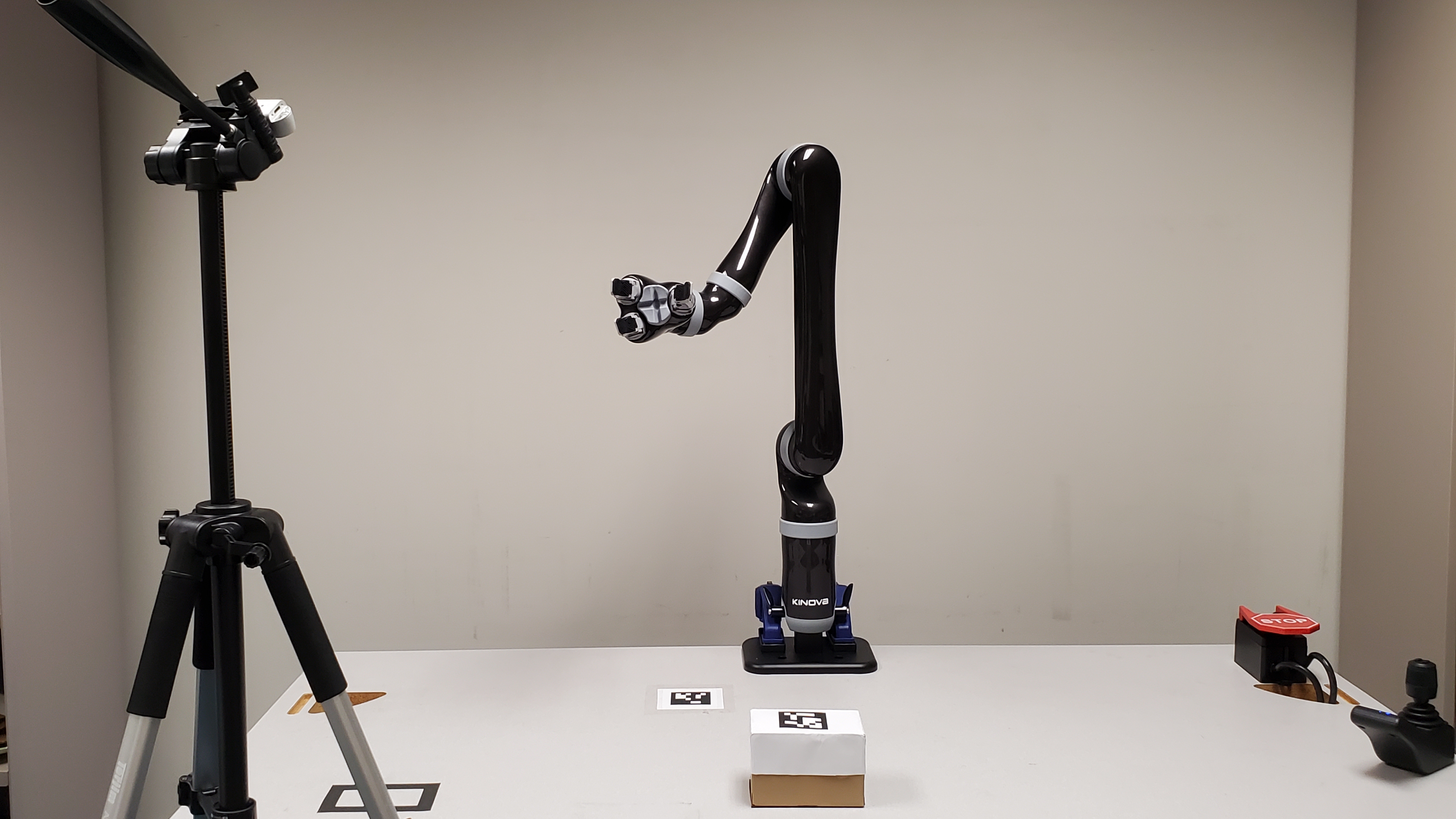}
    \caption{Experimental setup for the real robot experiments with a Kinova Jaco Gen 2 arm, \cite{campeau2019kinova}. An overhead Intel RealSense camera tracks an Aruco marker on the table to calibrate its 3D position w.r.t the world. It also tracks an Aruco marker on the block to extract its position.}
    \label{fig:jaco_robot_setup}
\end{figure}

\paragraph{Environment}
\label{appendix:sim_to_real_environment}
We setup a Kinova Jaco gen 2 arm as shown in Fig. \ref{fig:jaco_robot_setup}. Aruco marker are used to get the position of the block and forward kinematics of the robot is used to get the position of the robot end-effector.

The JacoReach and JacoPush tasks with the real robot have the same objective as the FetchReach and FetchPush tasks as described in the previous section. The observations and actions in the real robot were transformed (translated, rotated and scaled) to map to those in the simulated tasks.  20cm in the FetchReach task corresponds to 48cm in the JacoReach task. Thus, observations in the real robot were scaled by (20/48) before using as input to the trained policies. Similarly 20cm in the FetchPush task corresponds to 48cm in the JacoPush task and thus observations were scaled by (20/48). The policy commands sent to the robot were in the form of cartesian dispplacements which were finally executed by Moveit path planner. Due to inaccuracies in the real world, small actions couldn't be executed and this hurt the performance the algorithms (particularly the baseline algorithms which produced very small actions). To address this, the actions were scaled up by a factor of 7. Correspondingly, the timesteps were scaled down by a factor to 7 to adjust for action scaling. Thus the JacoReach task had (20/7 $\sim$ 3) timesteps and the JacoPush task had (30/7 $\sim$ 5) timesteps.

Due to the different scale in distance and length of episodes (timesteps), the rewards in the simulator and the real robot are in different scales.

\paragraph{Evaluation}
For the real robot tasks, the best seed from each algorithm is chosen and is evaluated over 5 episodes.

\subsection{Comparison to fully differentiable model based policy learning}
\label{appendix:comparison_differentiable_il}
If we have access to a differentiable model, we can directly obtain the gradient expected return (policy objective) w.r.t. the policy parameters $\theta$:

\begin{align}
    \Ebb_{\mathcal{T}\sim\mathcal{D}}
    \left[
    \nabla_\theta r(s_1,a_1) + \gamma \nabla_\theta r(s_2,a_2) + \gamma^2 \nabla_\theta r(s_3,a_3) + \dots
    \right] 
    \label{eq:naive_diff_appendix}
\end{align}

Since we can directly obtain the objective's gradient, we do not necessarily need to use either a critic ($Q$) as in standard Actor Critic (AC) algorithms or a residual critic ($C$) as in our proposed Actor Residual Critic (ARC) algorithms.

In ARC, we do not assume access to a differentiable dynamics model.

\clearpage
\section{Discussion on Results}
\label{apendix:result_discussion}

\subsection{Imitation Learning in Mujoco continuous-control tasks}
\begin{figure}[h!]
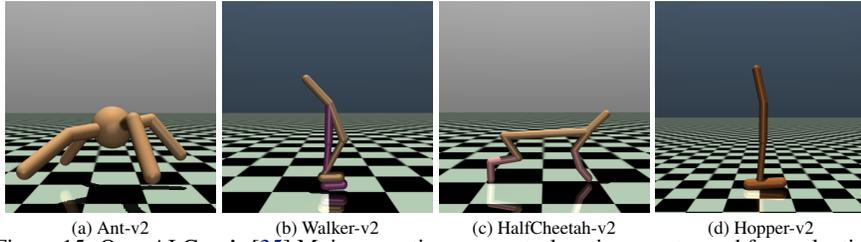

    \centering
    \begin{subfigure}[b]{\subfigwidth}
         \centering
         \includegraphics[width=\textwidth]{Figures/ant.png}
         \vspace{-5mm}
         \caption{Ant-v2}
     \end{subfigure}
     \begin{subfigure}[b]{\subfigwidth}
         \centering
         \includegraphics[width=\textwidth]{Figures/walker.png}
         \vspace{-5mm}
         \caption{Walker-v2}
     \end{subfigure}
     \begin{subfigure}[b]{\subfigwidth}
         \centering
         \includegraphics[width=\textwidth]{Figures/halfcheetah.png}
         \vspace{-5mm}
         \caption{HalfCheetah-v2}
     \end{subfigure}
     \begin{subfigure}[b]{\subfigwidth}
         \centering
         \includegraphics[width=\textwidth]{Figures/hopper.png}
         \vspace{-5mm}
         \caption{Hopper-v2}
     \end{subfigure}
     \vspace{-4mm}
    \caption{OpenAI Gym's \cite{brockman2016openai} Mujoco continuous-control environments used for evaluation.}
    % \label{fig:mujoco_envs}
\end{figure}
% \vspace{-5mm}
% \newcommand{\plotwidth}{0.2\textwidth}
\begin{figure}[h!]
     \vspace{-2mm}
     \centering
     \begin{subfigure}[b]{\textwidth}
         \centering
         \includegraphics[width=\textwidth]{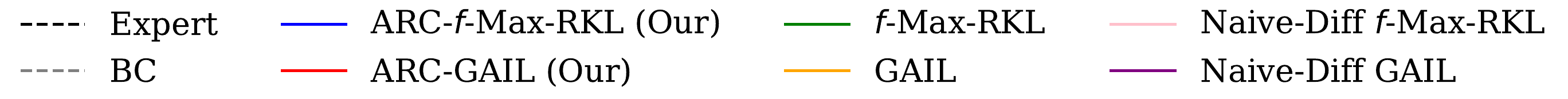}
         \vspace{-4mm}
     \end{subfigure}
     \begin{subfigure}[b]{\plotwidth}
         \centering
         \includegraphics[width=\textwidth]{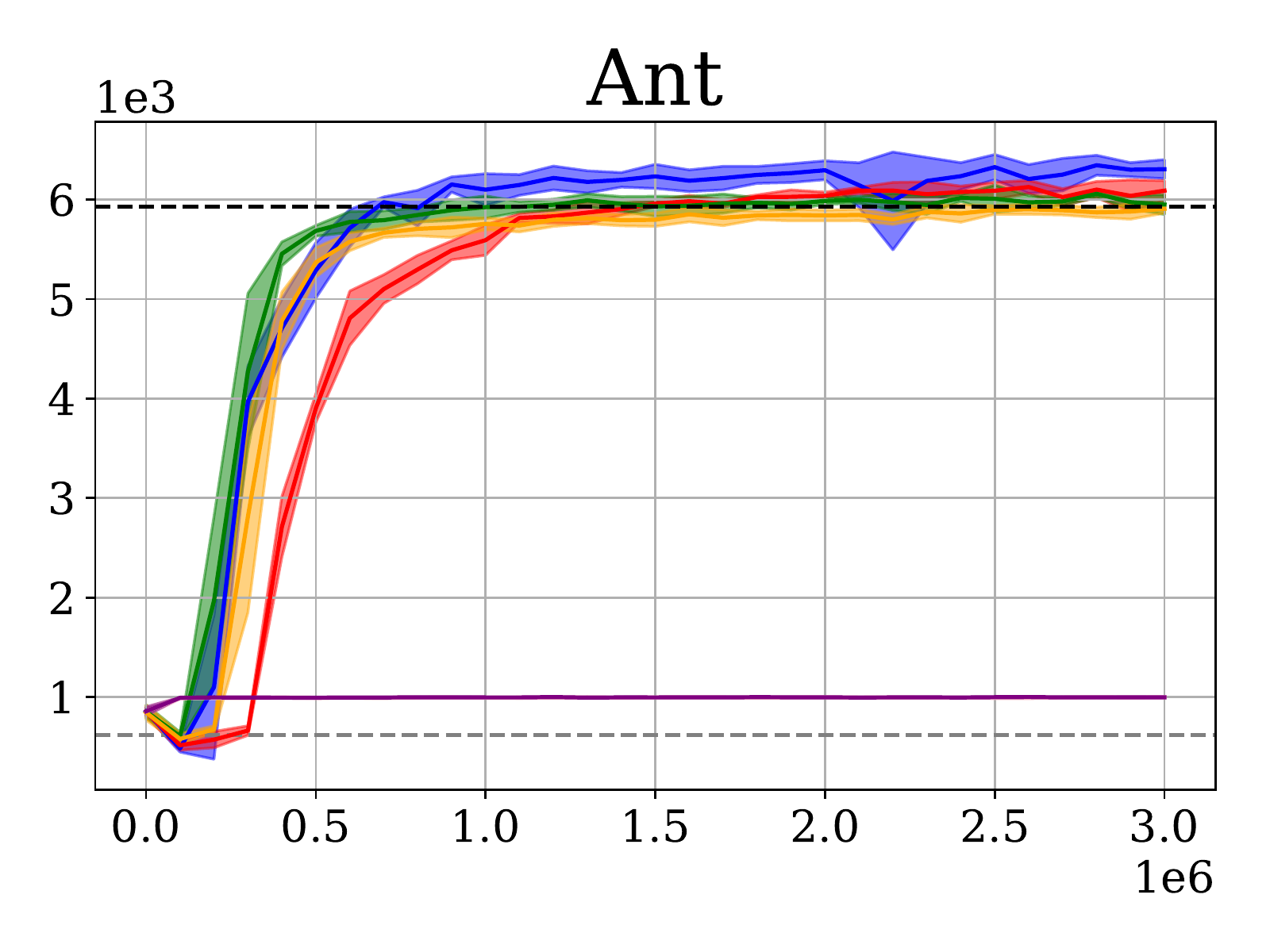}
     \end{subfigure}
     \hfill
     \begin{subfigure}[b]{\plotwidth}
         \centering
         \includegraphics[width=\textwidth]{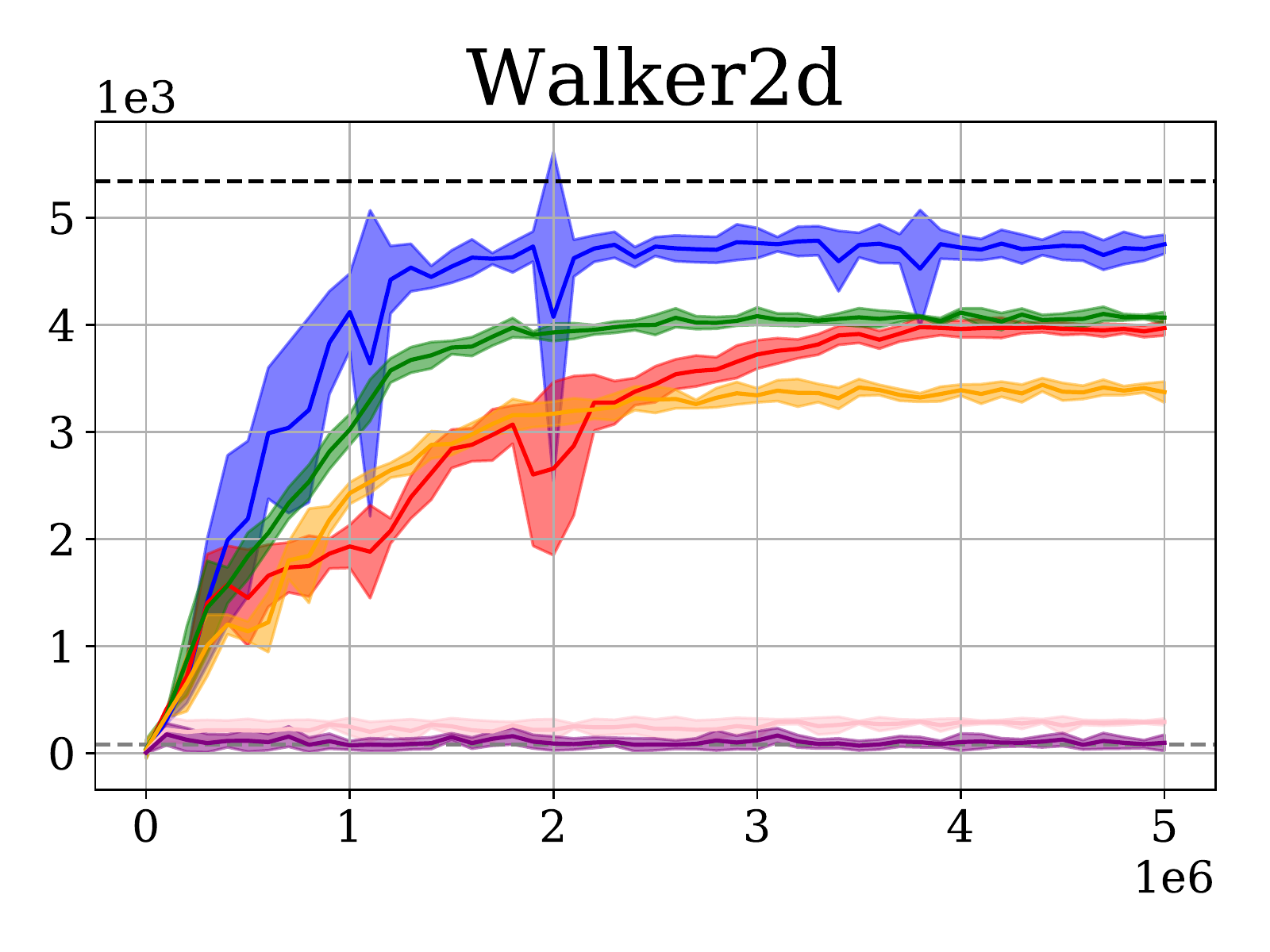}
     \end{subfigure}
     \hfill
     \begin{subfigure}[b]{\plotwidth}
         \centering
         \includegraphics[width=\textwidth]{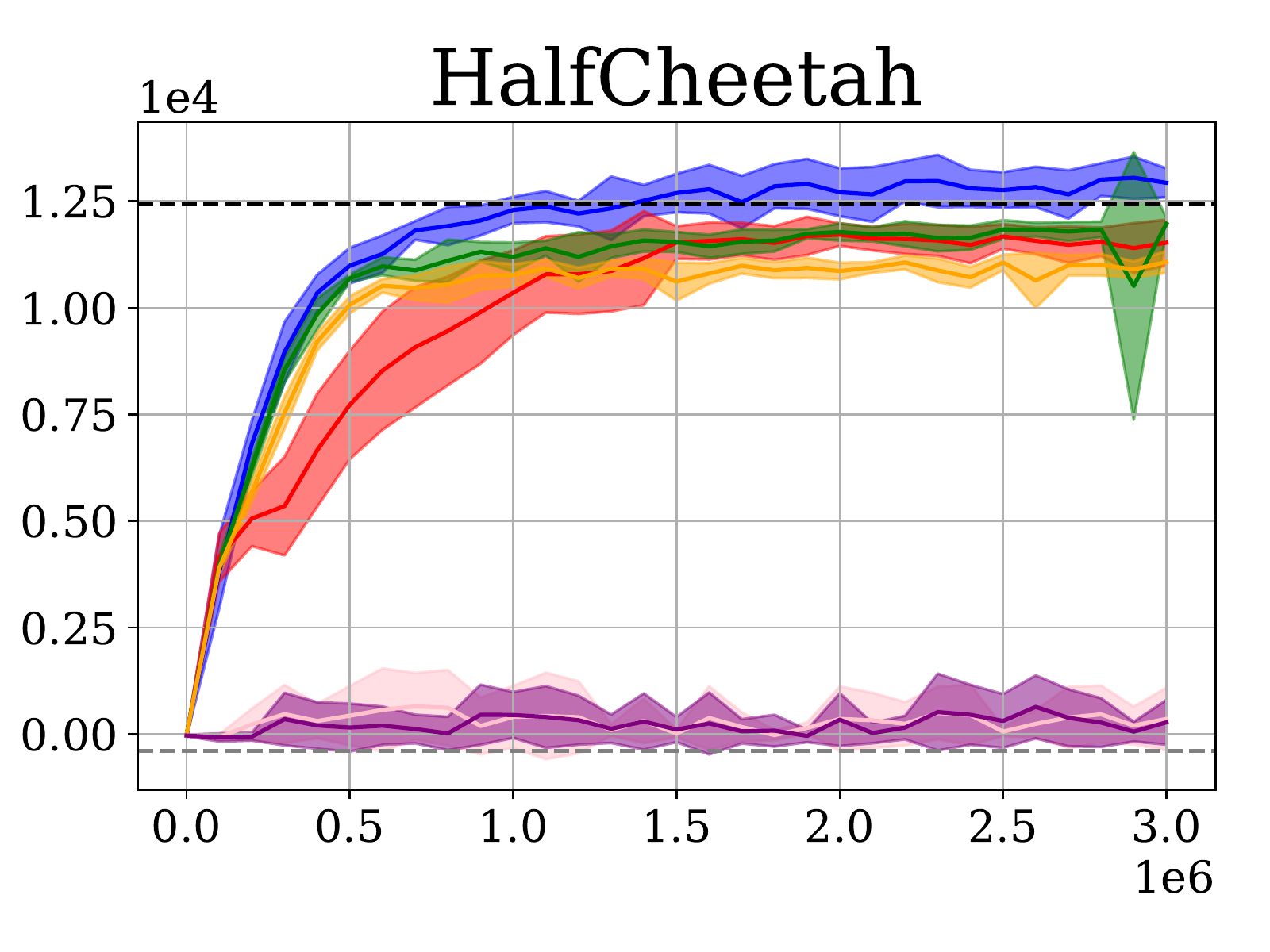}
     \end{subfigure}
     \hfill
     \begin{subfigure}[b]{\plotwidth}
         \centering
         \includegraphics[width=\textwidth]{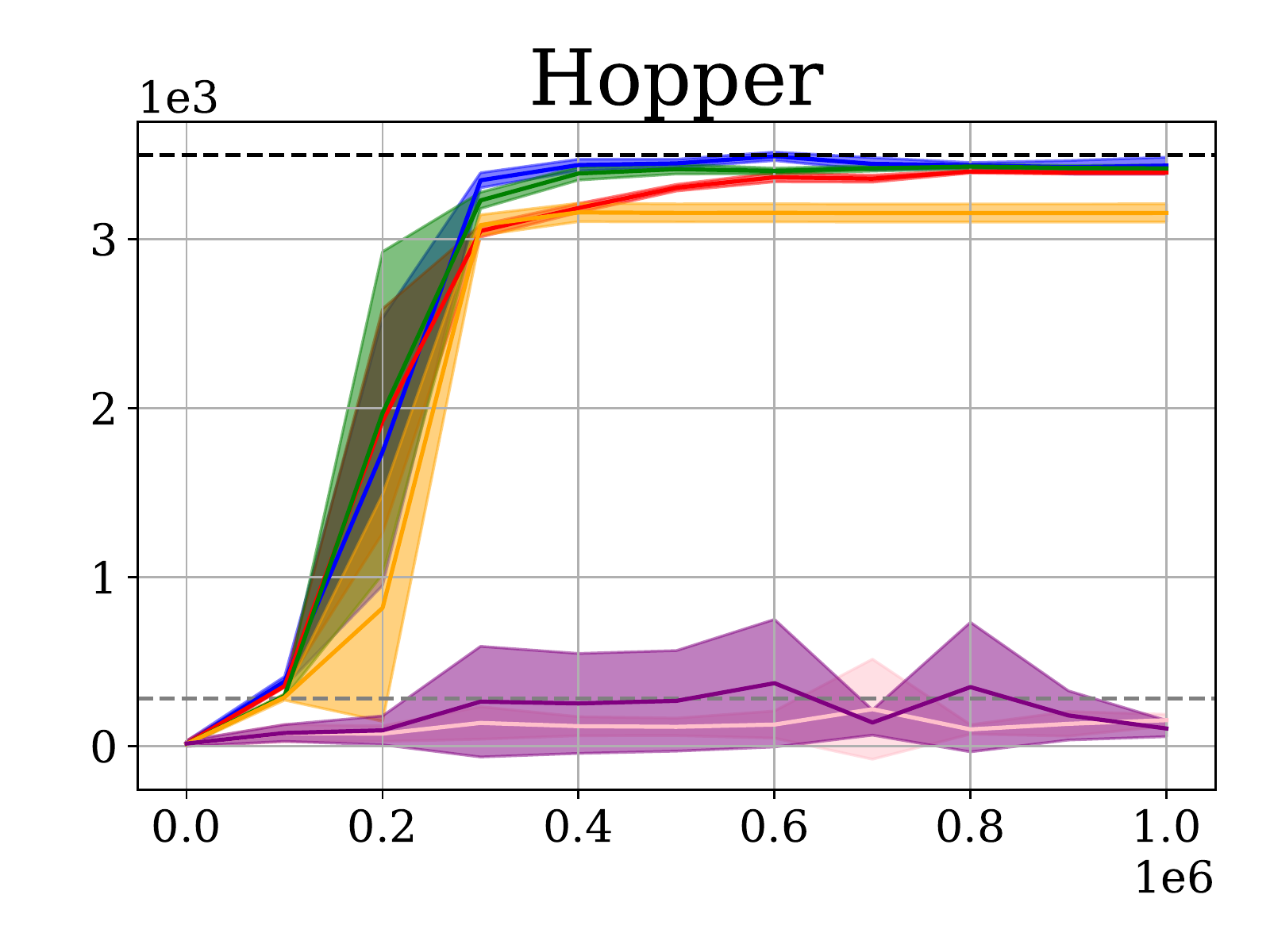}
        %  \caption{$Q^* = r^* + C^*$}
        %  \label{fig:policy_iteration_grid_q}
     \end{subfigure}
     \vspace{-5mm}
     \caption{Episode return versus number of environment interaction steps for different Imitation Learning algorithms on Mujoco continuous-control environments.}
     \label{fig:mujoco_training_plots_appendix}
    %  \vspace{-0.5cm}
\end{figure}
% \vspace{-3mm}
\begin{table}[h!]
    % \vspace{2mm}
    % \vspace{-2cm}
    \centering
    \scriptsize
    \begin{tabular}{c|c|c|c|c}
        \toprule
         Method &
         \multicolumn{1}{c|}{Ant} & \multicolumn{1}{c|}{Walker2d} & \multicolumn{1}{c|}{HalfCheetah} &
         \multicolumn{1}{c}{Hopper} \\
        \midrule
        Expert return &
  \multicolumn{1}{c|}{5926.18 $\pm$ 124.56} & \multicolumn{1}{c|}{5344.21 $\pm$ 84.45} & \multicolumn{1}{c|}{12427.49 $\pm$ 486.38} &        
  \multicolumn{1}{c}{3592.63 $\pm$ 19.21} \\
        %  \# Expert traj
        %   & 1 & 4 & 16
        %   & 1 & 4 & 16
        %   & 1 & 4 & 16
        %   & 1 & 4 & 16 \\
       \midrule
    ARC-$f$-Max-RKL (Our) & \textbf{6306.25} $\pm$ 95.91 & \textbf{4753.63} $\pm$ 88.89 & \textbf{12930.51} $\pm$ 340.02 & \textbf{3433.45} $\pm$ 49.48 \\  
    $f$-Max-RKL & 5949.81 $\pm$ 98.75 & 4069.14 $\pm$ 52.14 & 11970.47 $\pm$ 145.65 & 3417.29 $\pm$ 19.8 \\ 
    Naive-Diff $f$-Max-RKL & 998.27 $\pm$ 3.63 & 294.36 $\pm$ 31.38 & 357.05 $\pm$ 732.39 & 154.57 $\pm$ 34.7 \\ 
    \midrule
    ARC-GAIL (Our) & \textbf{6090.19} $\pm$ 99.72 & \textbf{3971.25} $\pm$ 70.11 & \textbf{11527.76} $\pm$ 537.13 & \textbf{3392.45} $\pm$ 10.32 \\
    GAIL & 5907.98 $\pm$ 44.12 & 3373.26 $\pm$ 98.18 & 11075.31 $\pm$ 255.69 & 3153.84 $\pm$ 53.61 \\ 
    Naive-Diff GAIL & 998.17 $\pm$ 2.22 & 99.26 $\pm$ 76.11 & 277.12 $\pm$ 523.77 & 105.3 $\pm$ 48.01 \\
    \midrule
    BC & 615.71 $\pm$ 109.9 & 81.04 $\pm$ 119.68 & -392.78 $\pm$ 74.12 & 282.44 $\pm$ 110.7 \\ 
        \bottomrule
    \end{tabular}
    \vspace{3mm}
    \caption{Policy return on Mujoco environments using different Adversarial Imitation Learning algorithms. Each algorithm is run with 10 random seeds and each seed is evaluated for 20 episodes.}
    \label{tab:mujoco_performance_appendix}
    \vspace{-5mm}
\end{table}
Fig. \ref{fig:mujoco_training_plots_appendix} shows the training plots and Table \ref{tab:mujoco_performance_appendix} shows the final performance of the various algoritms. Across all environments and across both the AIL algorithms, incorporating ARC shows consistent improvement over standard AIL algorithms. That is, ARC-$f$-Max-RKL outperformed $f$-Max-RKL and ARC-GAIL outperformed GAIL. Across all algorithms, ARC-$f$-Max-RKL showed the highest performance. BC suffers from distribution shift at test time \cite{ross2011reduction,ho2016generative} and performs very poorly.
As we predicted in Section \ref{subsection:naive_diff}, Naive-Diff algorithms don't perform well as naively using autodiff doesn't compute the gradients correctly. 

\textbf{Walker2d}
ARC algorithms show the highest performance gain in the Walker2d environment. ARC-$f$-Max-RKL shows the highest performance followed by $f$-Max-RKL, ARC-GAIL and GAIL respectively. Naive-Diff and BC algorithms perform poorly and the order is Naive-Diff $f$-Max-RKL, Naive-Diff GAIL and BC.

\textbf{Ant}, \textbf{HalfCheetah} and \textbf{Hopper}
ARC algorithms show consistent improvement over the standard AIL algorithms. However, there is only a modest improvement. This can be attributed to the fact that the baseline standard AIL algorithms already perform very well (almost matching expert performance). This leaves limited scope of improvement for ARC.

\textbf{Ranking the algorithms} Table \ref{tab:mujoco_ranking} ranks the various algorithms in each of these environments. Amongst the 4 AIL algorithms, ARC-$f$-Max-RKL consistently ranked 1 and GAIL consistently ranked 4. The relative performance of $f$-Max-RKL and ARC-GAIL varied across the environments, i.e. sometimes the former performed better and at other times the later performed better. The relative performance of Naive-Diff $f$-Max-RKL, Naive-Diff GAIL and BC also varied across the environments and they got ranks in the range 5 to 7.

\begin{table}[h!]
    % \vspace{2mm}
    % \vspace{-2cm}
    \centering
    \scriptsize
    \begin{tabular}{c|c|c|c|c}
        \toprule
         Method &
         \multicolumn{1}{c|}{Ant} & \multicolumn{1}{c|}{Walker2d} & \multicolumn{1}{c|}{HalfCheetah} &
         \multicolumn{1}{c}{Hopper} \\
   \midrule
    ARC-$f$-Max-RKL (Our) & 1 & 1 & 1 & 1 \\ 
    $f$-Max-RKL & 3 & 2 & 2 & 2\\ 
    Naive-Diff $f$-Max-RKL & 5 & 5 & 5 & 6 \\ 
    \midrule
    ARC-GAIL (Our) & 2 & 3 & 3 & 3 \\
    GAIL & 4 & 4 & 4 & 4\\ 
    Naive-Diff GAIL & 6 & 6 & 6 & 7 \\
    \midrule
    BC & 7 & 7 & 7 & 5 \\ 
        \bottomrule
    \end{tabular}
    \vspace{3mm}
    \caption{Ranking different Imitation Learning algorithms based on policy return in Mujoco environments. Each algorithms is run with 5 random seeds and each seed is evaluated for 20 episodes.}
    \label{tab:mujoco_ranking}
    \vspace{-5mm}
\end{table}

%%%%%%%%%%%%%%%%%%%%%%%%%%%%%%%%%%%%%

\subsection{Imitation Learning in robotic manipulation tasks}
\begin{figure}[h!]
    \centering
    \begin{subfigure}[b]{\subfigwidth}
         \centering
         \includegraphics[width=\textwidth]{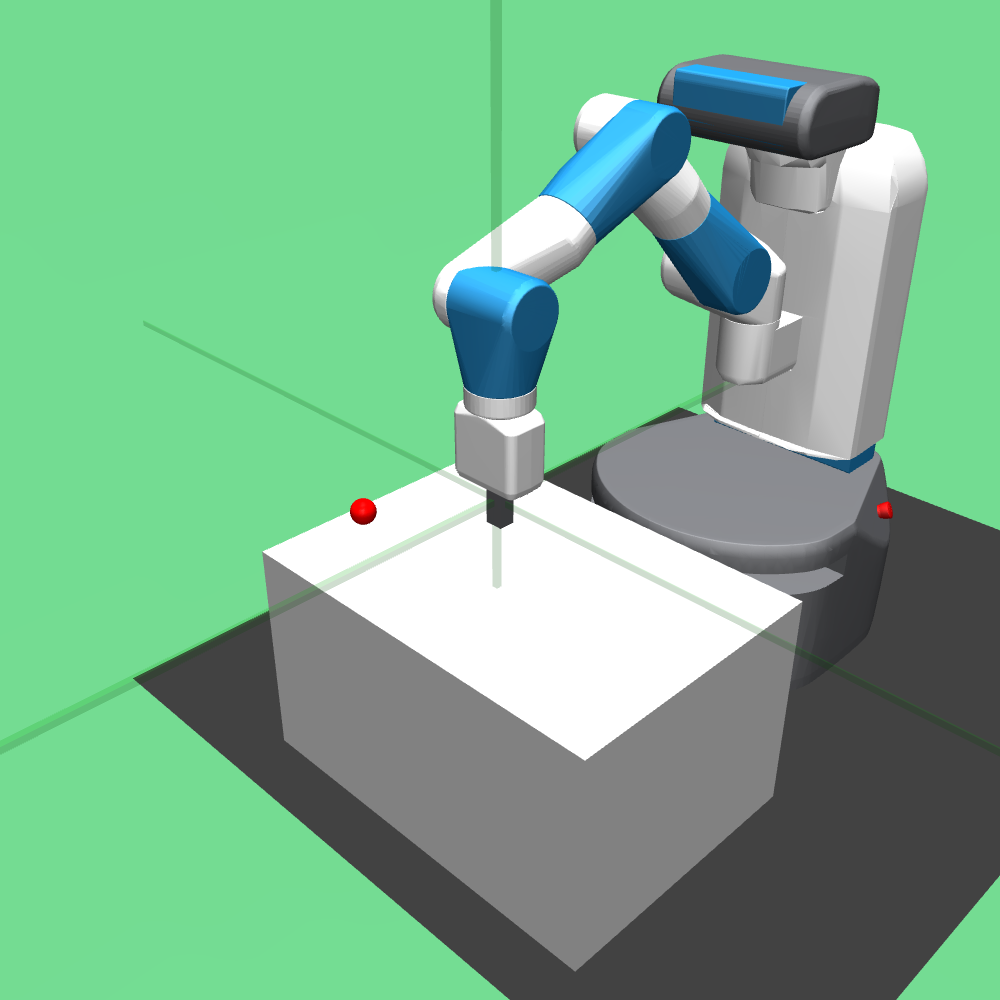}
         \vspace{-4mm}
         \caption{FetchReach}
         \label{fig:fetch_reach_appendix}
     \end{subfigure}
     \begin{subfigure}[b]{\subfigwidth}
         \centering
         \includegraphics[width=\textwidth]{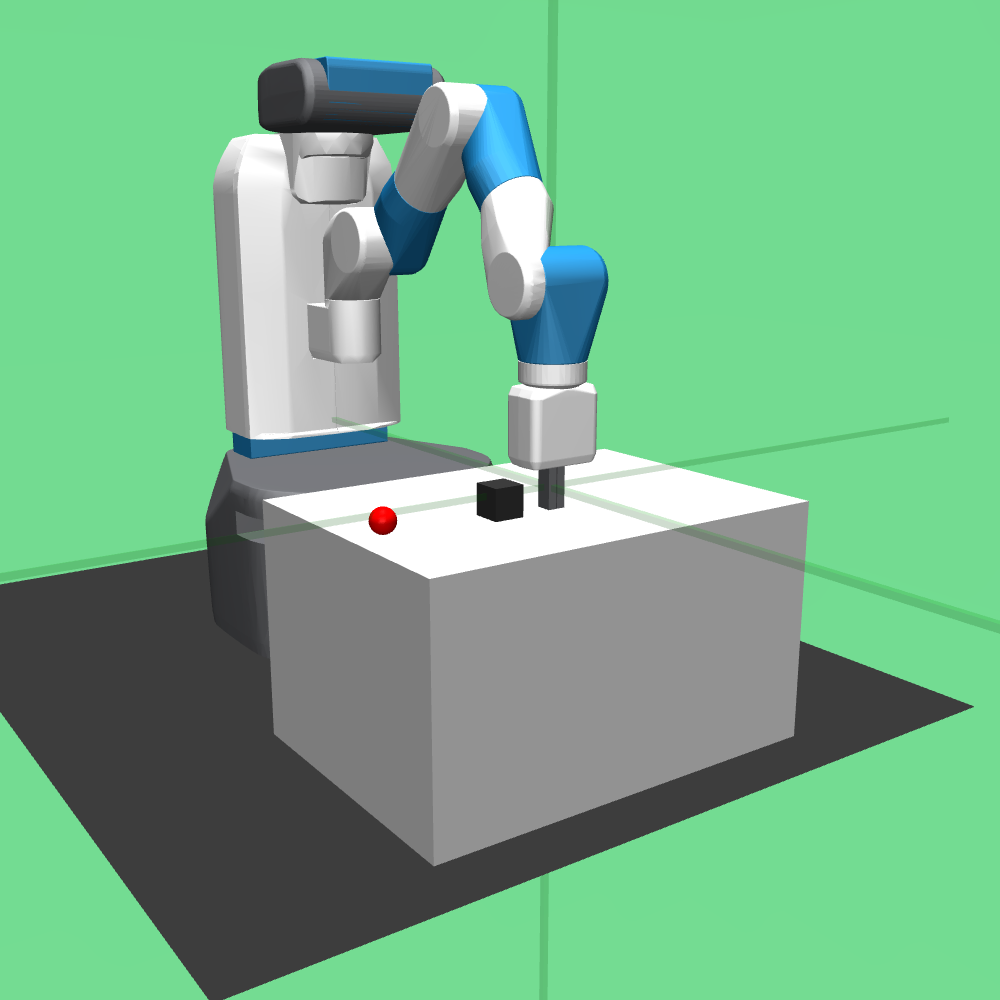}
         \vspace{-4mm}
         \caption{FetchPush}
         \label{fig:fetch_push_appendix}
     \end{subfigure}
     \begin{subfigure}[b]{\subfigwidth}
         \centering
         \includegraphics[width=\textwidth]{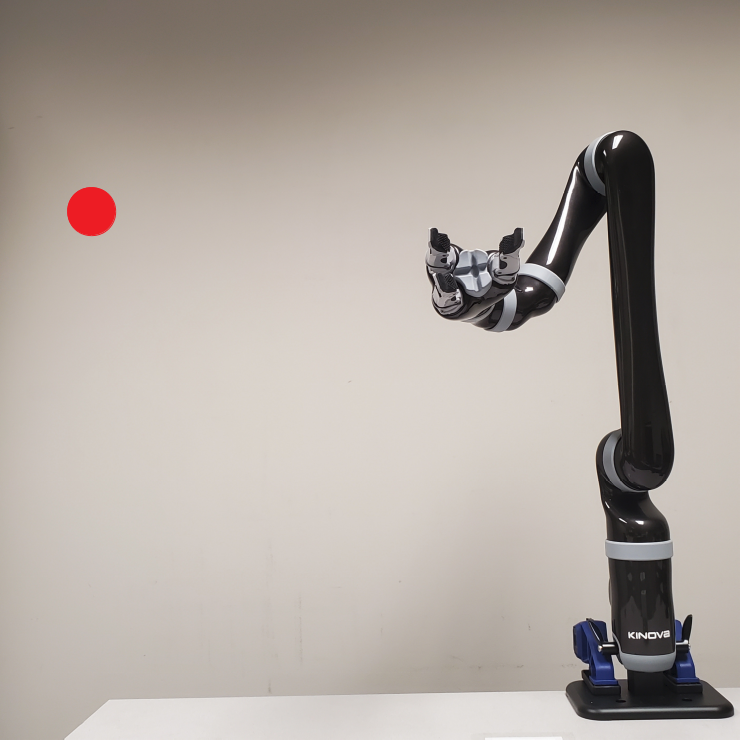}
         \vspace{-4mm}
         \caption{JacoReach}
         \label{fig:jaco_reach_appendix}
     \end{subfigure}
     \begin{subfigure}[b]{\subfigwidth}
         \centering
         \includegraphics[width=\textwidth]{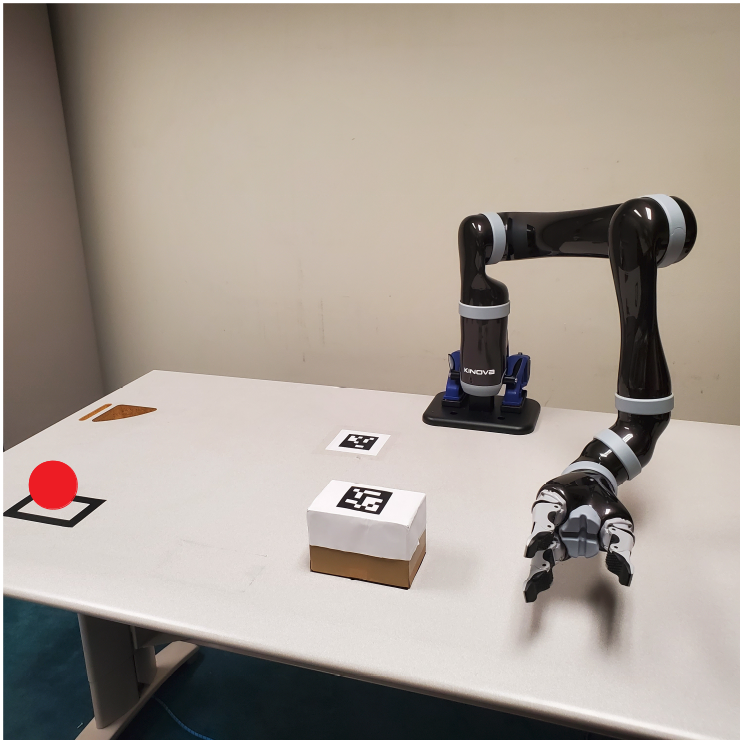}
         \vspace{-4mm}
         \caption{JacoPush}
         \label{fig:jaco_push_appendix}
     \end{subfigure}
     \vspace{-4mm}
    \caption{Simulated and real robotic manipulation tasks used for evaluation. Simplified 2D versions of the FetchReach \subref{fig:fetch_reach} and FetchPush \subref{fig:fetch_push} tasks from OpenAI Gym, \cite{brockman2016openai} with a Fetch robot, \cite{wise2016fetch}. Corresponding JacoReach \subref{fig:jaco_reach} and JacoPush \subref{fig:jaco_push} tasks with a real Kinova Jaco Gen 2 arm, \cite{campeau2019kinova}.} 
    \label{fig:robot_envs_appendix}
\end{figure}
\begin{figure}[h!]
     \vspace{-3mm}
     \centering
     \begin{subfigure}[b]{0.95\textwidth}
         \centering
         \includegraphics[width=\textwidth]{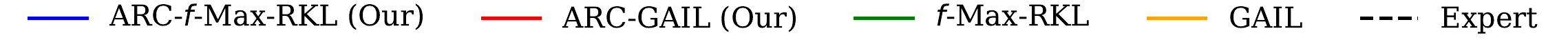}
     \end{subfigure}
     \begin{subfigure}[b]{\doubleplotwidth}
         \centering
         \includegraphics[width=0.49\textwidth]{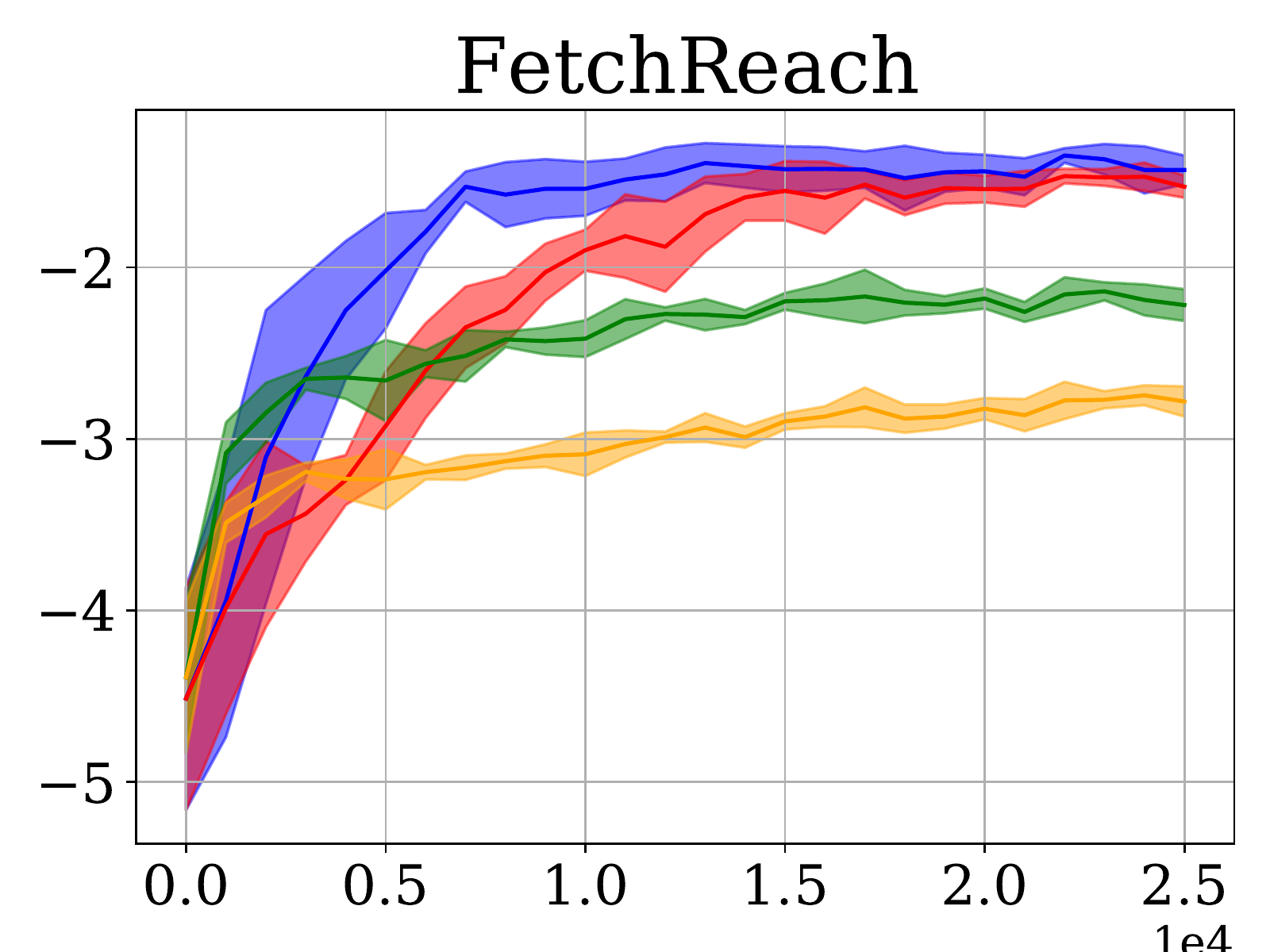}
         \centering
         \includegraphics[width=0.49\textwidth]{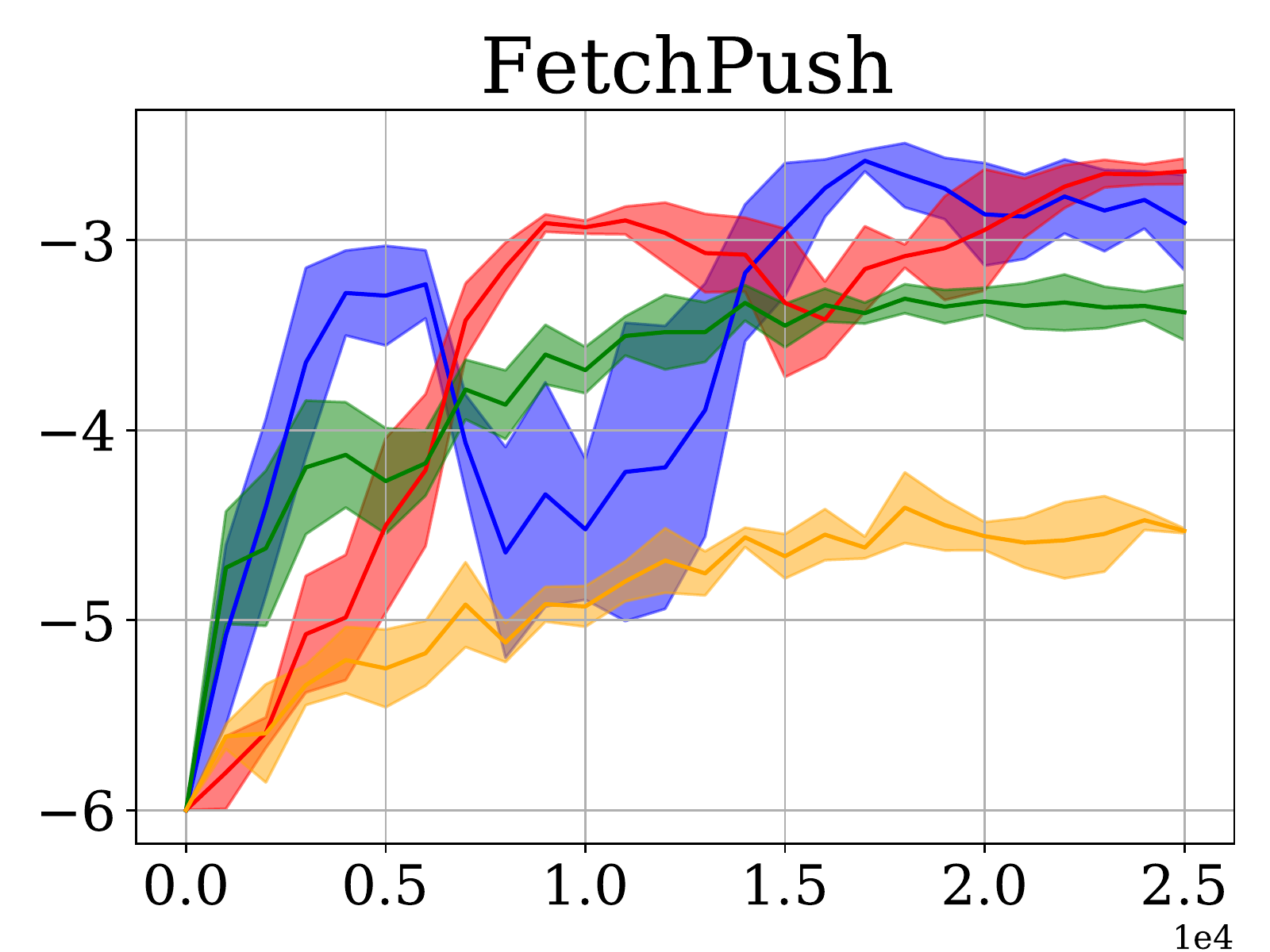}
         \vspace{-6mm}
         \subcaption{Episode return vs. interaction steps}
         \label{fig:sim_return_training_plot_appendix}
     \end{subfigure}
     \qquad
     \begin{subfigure}[b]{\doubleplotwidth}
         \centering
         \includegraphics[width=0.49\textwidth]{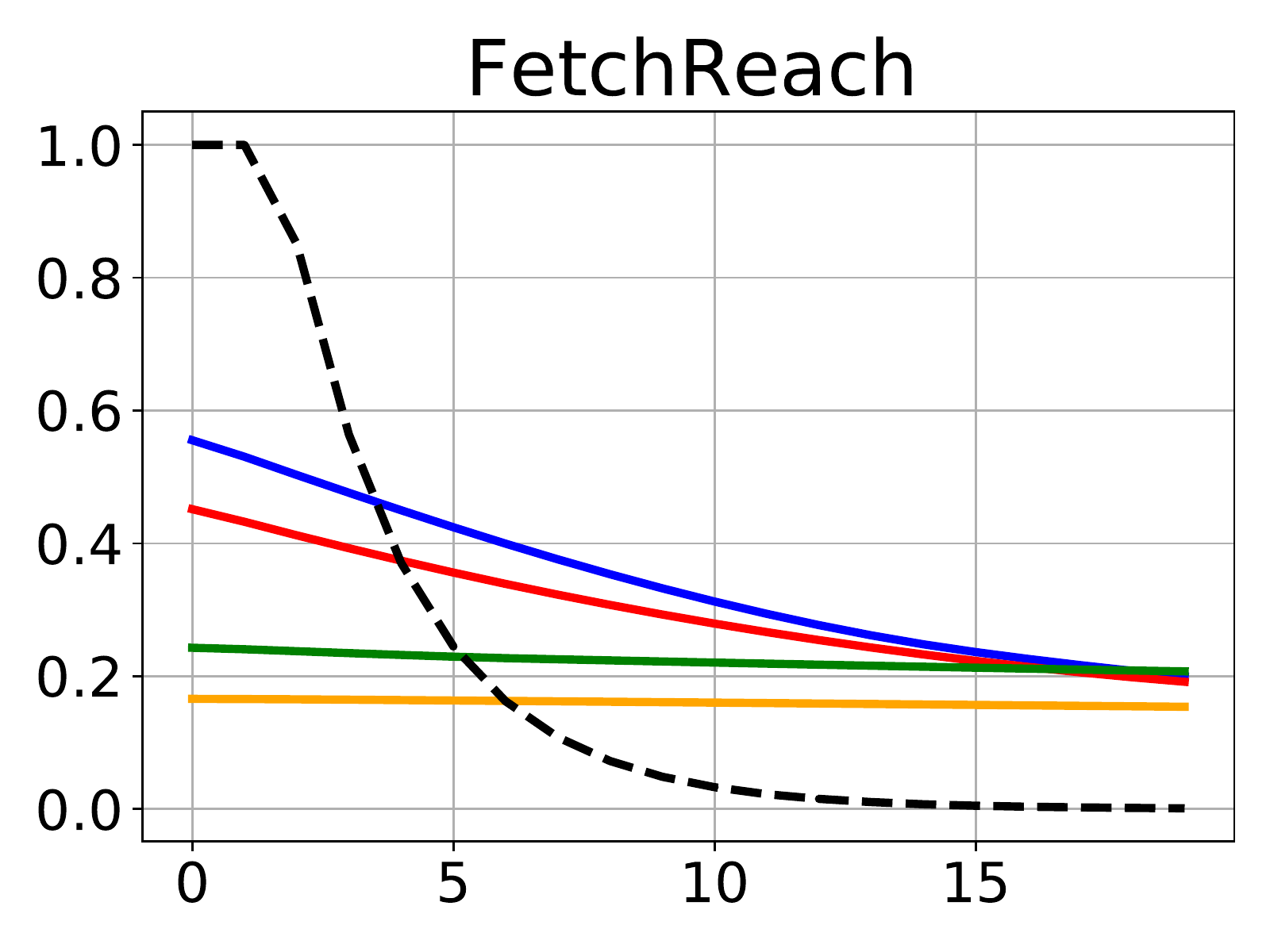}
         \centering
         \includegraphics[width=0.49\textwidth]{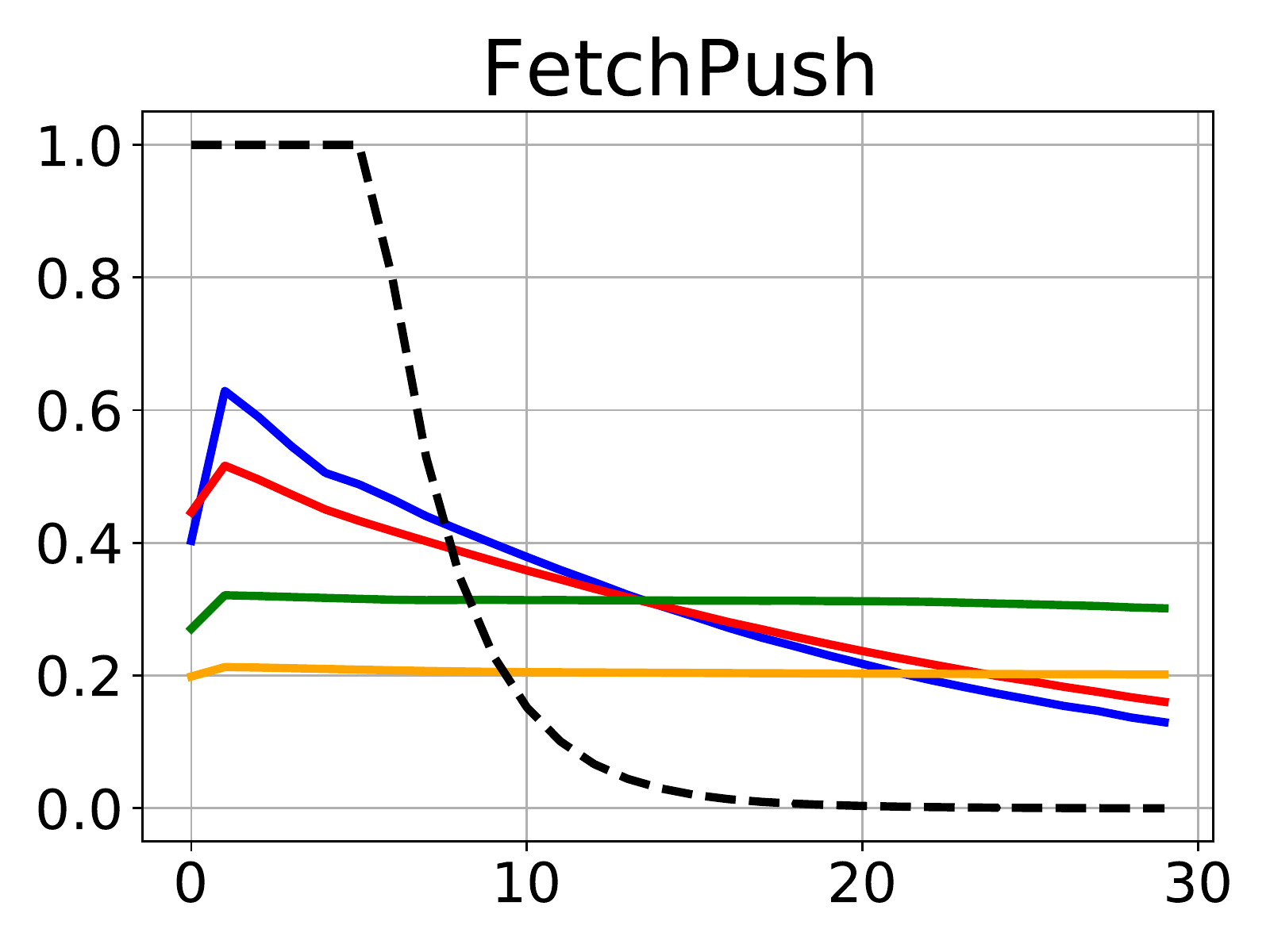}
         \vspace{-6mm}
         \subcaption{Action vs. time step}
         \label{fig:action_vs_timestep_appendix}
     \end{subfigure}
    %  \vspace{-1mm}
     \caption{\subref{fig:sim_return_training_plot} Episode return vs. number of environment interaction steps for different Adversarial Imitation Learning algorithms on FetchPush and FetchReach tasks. \subref{fig:action_vs_timestep} Magnitude of the $2^{nd}$ action dimension versus time step in a single episode for different algorithms.
    %  ARC aided AIL algorithms executed actions that resembled the expert actions more closely than standard AIL algorithms did.
     }
     \label{fig:robot_training_plots_appendix}
     \vspace{-3mm}
\end{figure}

% \vspace{-1mm}
\begin{table}[h!]
    % \vspace{2mm}
    \vspace{-2mm}
    \centering
    \scriptsize
    \begin{tabular}{c|c|c|c|c}
        \toprule
        &
         \multicolumn{2}{c|}{\textbf{Simulation}} &
        %  \multicolumn{1}{c|}{FetchPush} &
        %  \multicolumn{1}{c|}{JacoReach} &
         \multicolumn{2}{c}{\textbf{Real Robot}}\\
         Method &
         \multicolumn{1}{c}{FetchReach} & \multicolumn{1}{c|}{FetchPush} &
         \multicolumn{1}{c}{JacoReach} &
         \multicolumn{1}{c}{JacoPush}\\
        \midrule
        Expert return &
        \multicolumn{1}{c|}{-0.58 $\pm$ 0} & \multicolumn{1}{c|}{-1.18 $\pm$ 0.04}&
        -0.14 $\pm$ 0.01& 
        -0.77 $\pm$ 0.01\\
        \midrule
    ARC-$f$-Max-RKL (Our) & \textbf{-1.43} $\pm$ 0.08 & \textbf{-2.91} $\pm$ 0.25 & \textbf{-0.38} $\pm$ 0.02 & \textbf{-1.25} $\pm$ 0.06\\ 
    $f$-Max-RKL & -2.22 $\pm$ 0.09 & -3.38 $\pm$ 0.15 & -0.8 $\pm$ 0.05 & -2.03 $\pm$ 0.06 \\ 
    \midrule
    ARC-GAIL (Our) & \textbf{-1.53} $\pm$ 0.06 & \textbf{-2.64} $\pm$ 0.07 & \textbf{-0.46} $\pm$ 0.01 & \textbf{-1.56} $\pm$ 0.08\\ 
    GAIL & -2.78 $\pm$ 0.09 & -4.53 $\pm$ 0.01 & -1.05 $\pm$ 0.06 &-2.35 $\pm$ 0.06 \\ 
        \bottomrule
    \end{tabular}
    % \vspace{18mm}
    \caption{Policy return on simulated (FetchReach, FetchPush) and real (JacoReach, JacoPush) robotic manipulation tasks using different AIL algorithms. The reward at each time step is negative distance between end-effector \& goal for reach tasks and block \& goal for push tasks. The reward in the real and simulated tasks are on different scales due to implementation details described in Appendix \ref{appendix:sim_to_real_environment}.}
    \label{tab:robotic_task_performance_appendix}
    \vspace{-3mm}
\end{table}

Fig. \ref{fig:sim_return_training_plot_appendix} shows the training plots and Table \ref{tab:robotic_task_performance_appendix} under the heading `Simulation' shows the final performance of the different algorithms. In both the FetchReach and FetchPush tasks, ARC aided AIL algorithms consistently outperformed the standard AIL algorithms.
Amongst all the evaluated algorithms, ARC-$f$-Max-RKL performed the best in the FetchReach task and ARC-GAIL performed the best in the FetchPush task.

\textbf{Parameter robustness} In the robotic manipulation tasks, we didn't extensively tune the hyper-parameters for tuning (both ARC as well as the baselines). ARC algorithms performed significantly better than the standard AIL algorithms. This shows that ARC algorithms are parameter robust, which is a desirable property for real world robotics.

\textbf{Ranking the algorithms} Table \ref{tab:robotic_ranking} ranks the different algorithms based on the policy return. ARC-$f$-Max-RKL and ARC-GAIL rank either 1 or 2 in all the environments. $f$-Max-RKK and GAIL consistently rank 3 and 4 respectively.

\begin{table}[h!]
    % \vspace{2mm}
    \vspace{-2mm}
    \centering
    \scriptsize
    \begin{tabular}{c|c|c|c|c}
        \toprule
        &
         \multicolumn{2}{c|}{\textbf{Simulation}} &
        %  \multicolumn{1}{c|}{FetchPush} &
        %  \multicolumn{1}{c|}{JacoReach} &
         \multicolumn{2}{c}{\textbf{Real Robot}}\\
         Method &
         \multicolumn{1}{c}{FetchReach} & \multicolumn{1}{c|}{FetchPush} &
         \multicolumn{1}{c}{JacoReach} &
         \multicolumn{1}{c}{JacoPush}\\
        \midrule
    ARC-$f$-Max-RKL (Our) & 1 & 2 & 1 & 1 \\ 
    $f$-Max-RKL & 3 & 3 & 3 & 3 \\ 
    \midrule
    ARC-GAIL (Our) & 2 & 1 & 2 & 2 \\ 
    GAIL & 4 & 4 & 4 & 4\\ 
        \bottomrule
    \end{tabular}
    % \vspace{18mm}
    \caption{Ranking different Imitation Learning algorithms based on policy return in simulated and real robotic manipulation tasks. Each algorithms is run with 5 random seeds and each seed is evaluated for 20 episodes.}
    \label{tab:robotic_ranking}
    \vspace{-3mm}
\end{table}

Fig. \ref{fig:action_vs_timestep_appendix} shows the magnitude of the $2^{nd}$ action dimension vs. time-step in one episode for different algorithms. The expert initially executed large actions when the end-effector/block was far away from the goal. As the end-effector/block approached the goal, the expert executed small actions. ARC aided AIL algorithms (ARC-$f$-Max-RKL and ARC-GAIL) showed a similar trend while standard AIL algorithms ($f$-Max-RKL and GAIL) learnt a nearly constant action. Thus, ARC aided AIL algorithms were able to better imitate the expert than standard AIL algorithms.

\subsection{Sim-to-real transfer of robotic manipulation policies}
Table \ref{tab:robotic_task_performance_appendix} under the heading `Real Robot' shows the performance of the different AIL algorithms in the real robotic manipulation tasks. The real robot evaluations showed a similar trend as in the simulated tasks. ARC aided AIL consistently outperformed the standard AIL algorithms.

\textbf{Ranking the algorithms} Table \ref{tab:robotic_ranking} ranks the different algorithms based on the policy return. ARC-$f$-Max-RKL, ARC-GAIL, $f$-Max-RKL and GAIL consistently ranked 1, 2, 3 and 4 respectively.

\end{document}